\documentclass[onecolumn]{cohere}

\usepackage[dvipsnames]{xcolor} 
\usepackage[utf8]{inputenc}
\usepackage[T1]{fontenc}
\usepackage{lipsum}
\usepackage{placeins}
\usepackage{geometry}
\usepackage{colortbl}

\usepackage{stfloats} 
\usepackage[labelformat=empty]{subcaption}
\usepackage{hyperref}
\hypersetup{colorlinks,citecolor=blue}
\usepackage{url}
\usepackage{booktabs}
\usepackage{graphicx}
\usepackage{multirow}
\usepackage{adjustbox}
\usepackage{float}
\usepackage{caption}
\usepackage{subcaption}
\usepackage[colorinlistoftodos]{todonotes}
\usepackage{lscape}
\usepackage{rotating}
\usepackage{enumitem}
\usepackage{latexsym}
\usepackage[normalem]{ulem}
\usepackage{amssymb}

\newcommand\blfootnote[1]{%
  \begingroup
  \renewcommand\thefootnote{}\footnote{#1}%
  \addtocounter{footnote}{-1}%
  \endgroup
}

\newcommand{\earthmoji}{\raisebox{-1pt}{\includegraphics[width=0.9em]{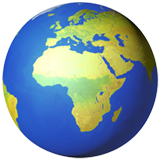}}}

\newcommand{\scalesemoji}{\raisebox{-1pt}{\includegraphics[width=0.9em]{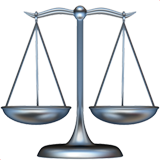}}}

\newcommand{\liteemoji}{\raisebox{-1pt}{\includegraphics[width=0.9em]{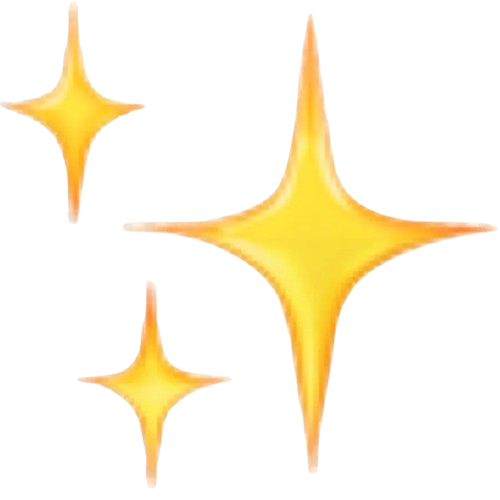}}}

\newcommand{\statuelibertyemoji}{\raisebox{-1pt}{\includegraphics[width=0.9em]{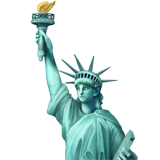}}}

\newcommand{\chopsticksmoji}{\raisebox{-1pt}{\includegraphics[width=0.9em]{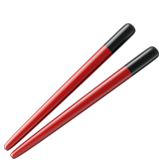}}}

\newcommand{\worldmapsmoji}{\raisebox{-1pt}{\includegraphics[width=0.9em]{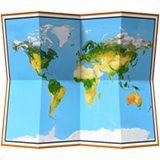}}}

\newcommand{\speakersmoji}{\raisebox{-1pt}{\includegraphics[width=0.9em]{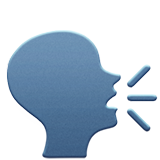}}}

\newcommand{\memoemoji}{\raisebox{-1pt}{\includegraphics[width=0.9em]{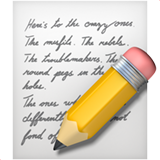}}}

\newcommand{\timeremoji}{\raisebox{-1pt}{\includegraphics[width=0.9em]{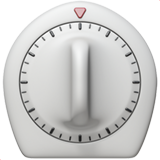}}}

\newcommand{\highemoji}{\raisebox{-1pt}{\includegraphics[width=0.9em]{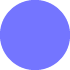}}}

\newcommand{\midemoji}{\raisebox{-1pt}{\includegraphics[width=0.9em]{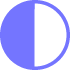}}}

\newcommand{\lowemoji}{\raisebox{-1pt}{\includegraphics[width=0.9em]{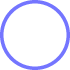}}}

\newcommand{\gmmlu}{\textbf{Global-MMLU} \earthmoji}
\newcommand{\cs}{\textbf{CS}\statuelibertyemoji}
\newcommand{\ca}{\textbf{CA}\scalesemoji}
\newcommand{\gmmlulite}{\textbf{Global-MMLU Lite} \liteemoji}

\usepackage[framemethod=TikZ]{mdframed}
\usepackage{tcolorbox}
\usepackage{multicol}
\usepackage{longtable}
\usepackage{enumitem}

\definecolor{pt_symbol}{RGB}{207, 159, 255} 

\definecolor{ct_symbol}{RGB}{173, 216, 230} 

\DeclareSymbolFont{extraup}{U}{zavm}{m}{n}
\DeclareMathSymbol{\vardiamond}{\mathalpha}{extraup}{87}
\DeclareMathSymbol{\varclub}{\mathalpha}{extraup}{85}

\newcommand{\dia}{%
  \ooalign{%
    \scalebox{1.09}{\textcolor{pt_symbol}{$\vardiamond$}}\cr
    \raisebox{.2ex}{\hspace{0.029em}\scalebox{1.1}{\textcolor{black}{$\diamondsuit$}}}\cr
  }%
}

\newcommand{\spade}{%
  \ooalign{%
    \textcolor{ct_symbol}{\raisebox{0.21ex}{\hspace{.085em}\resizebox{0.75em}{0.6em}{$\clubsuit$}}}\cr 
    \raisebox{-0.18ex}{\scalebox{1.1}{\textcolor{black}{$\varclub$}}}\cr
  }%
}

\DeclareRobustCommand*{\specialdia}{\dia}
\DeclareRobustCommand*{\specialspade}{\spade}

\colorlet{LightGreen}{green!20}
\colorlet{LightRed}{red!20}
\colorlet{LightGrey}{black!20}

\tcbset{on line,
        boxsep=1.5pt,left=0pt,right=0pt,top=0pt,bottom=0pt,
        colframe=white
        }

\newtcolorbox{mybox}[2][]{
  colback=white, 
  colframe=lightblue,
  fonttitle=\bfseries,
  coltitle=black,  
  title=#2, 
  #1
}

\title{Global MMLU \earthmoji: Understanding and Addressing Cultural and Linguistic Biases in Multilingual Evaluation}

\multiauthors
\author{name={Shivalika Singh\fa},affiliation={1}}
\author{name={Angelika Romanou},affiliation={2}}
\author{name={Clémentine Fourrier},affiliation={3}}
\author{name={David I. Adelani},affiliation={4}}
\author{name={Jian Gang Ngui},affiliation={5,6}}
\author{name={Daniel Vila-Suero},affiliation={3}}
\author{name={Peerat Limkonchotiwat},affiliation={5,6}}
\author{name={Kelly Marchisio},affiliation={7}}
\author{name={Wei Qi Leong},affiliation={5,6}}
\author{name={Yosephine Susanto},affiliation={5,6}}
\author{name={Raymond Ng},affiliation={5,6}}
\author{name={Shayne Longpre},affiliation={8}}
\author{name={Sebastian Ruder},affiliation={15}}
\author{name={Wei-Yin Ko},affiliation={7}}
\author{name={Madeline Smith},affiliation={1}}
\author{name={Antoine Bosselut},affiliation={2}}
\author{name={Alice Oh},affiliation={9}}
\author{name={Andr\'e F. T. Martins},affiliation={10,11}}
\author{name={Leshem Choshen},affiliation={12}}
\author{name={Daphne Ippolito},affiliation={13}}
\author{name={Enzo Ferrante},affiliation={14}}
\author{name={Marzieh Fadaee},affiliation={1}}
\author{name={Beyza Ermis\faa},affiliation={1}}
\author{name={Sara Hooker\faa},affiliation={1}}
\affiliations{
    \item[1] Cohere For AI
    \item[2] EPFL
    \item[3] Hugging Face
    \item[4] Mila, McGill University \& Canada CIFAR AI Chair
    \item[5] AI Singapore
    \item[6] National University of Singapore
    \item[7] Cohere
    \item[8] MIT
    \item[9] KAIST
    \item[10] Instituto de Telecomunica\c{c}\~{o}es
    \item[11] Instituto Superior T\'ecnico, Universidade de Lisboa
    \item[12] MIT, MIT-IBM Watson AI Lab
    \item[13] Carnegie Mellon University
    \item[14] CONICET \& Universidad de Buenos Aires
    \item[15] Meta AI Research

}

\date{\today}
\abstract{
Cultural biases in multilingual datasets pose significant challenges for their effectiveness as global benchmarks. These biases stem not only from differences in language but also from the cultural knowledge required to interpret questions, reducing the practical utility of translated datasets like MMLU. Furthermore, translation often introduces artifacts that can distort the meaning or clarity of questions in the target language. A common practice in multilingual evaluation is to rely on machine-translated evaluation sets, but simply translating a dataset is insufficient to address these challenges.
In this work, we trace the impact of both of these issues on multilingual evaluations and ensuing model performances. Our large-scale evaluation of state-of-the-art open and proprietary models illustrates that progress on MMLU depends heavily on learning Western-centric concepts, with 28\% of all questions requiring culturally sensitive knowledge. Moreover, for questions requiring geographic knowledge, an astounding 84.9\% focus on either North American or European regions. Rankings of model evaluations change depending on whether they are evaluated on the full portion or the subset of questions annotated as culturally sensitive, showing the distortion to model rankings when blindly relying on translated MMLU. We release \gmmlu, an improved MMLU with evaluation coverage across 42 languages -- with improved overall quality by engaging with compensated professional and community annotators to verify translation quality while also rigorously evaluating cultural biases present in the original dataset. This comprehensive \gmmlu{} set also includes designated subsets labeled as \textbf{culturally sensitive} \statuelibertyemoji{} and \textbf{culturally agnostic} \scalesemoji{} to allow for more holistic, complete evaluation.

\small
\gmmlu: \url{https://hf.co/datasets/CohereForAI/Global-MMLU} 

\gmmlulite: \url{https://huggingface.co/datasets/CohereForAI/Global-MMLU-Lite}
\normalsize
}

\begin{document}

\blfootnote{Corresponding authors: \{\url{shivalika}, \url{beyza}, \url{sarahooker}\}\url{@cohere.com}}

\section{Introduction}
\label{sec:intro}

\begin{quote}
\textit{I contain multitudes. -- Walt Whitman, 1855}
\end{quote} 
Language cannot be simply reduced to a utilitarian tool, otherwise there would be no reason to have so many diverse ways for saying the same thing or referring to similar concepts. Indeed, language is also a marker of belonging and a repository of cultural knowledge \citep{labov1963social,labov1986social,karlık-meryem-2023-culture-on-language-learning}.
Today, state-of-the-art generative AI is used around the world and yet evaluation of these systems is primarily conducted using English benchmarks \citep{zellers2019hellaswag,hendrycks2020measuring,suzgun2022challenging,zhang2023evaluating}. Where multilingual evaluations are relied upon, these are often simply machine translations of widely adopted English benchmarks~\citep{lai2023okapi,ustun2024ayamodelinstructionfinetuned}.

A pressing question arises: \textit{how can we develop large language models (LLMs) that perform effectively and fairly across the full spectrum of languages and cultures?} The lack of comprehensive evaluation benchmarks for many languages poses a significant obstacle for researchers and practitioners striving to create truly multilingual systems. Often, a common practice is to simply translate English benchmarks into other languages. In this work, we consider the implications of this given one of the most ubiquitous examples --  the Massive Multitask Language Understanding (MMLU) dataset~\citep{hendrycks2020measuring}. Originally compiled using sources in the English language across 57 diverse subject areas such as elementary mathematics, computer science, and law, the dataset is often machine-translated into resources for multilingual assessment, which we collectively term \emph{transMMLU} \citep{lai2023okapi,ustun2024ayamodelinstructionfinetuned,openai2024gpt4technicalreport,dubey2024llama,bendale2024sutrascalablemultilinguallanguage}.
However, the growing adoption of automatically translated \textit{``as-is''} \emph{transMMLU} as a barometer of global AI progress deserves closer inspection and reflection. 

While widely adopted for multilingual evaluations, the multilinguality achieved through the translation of English datasets does not guarantee multiculturality. Evaluating on blindly-translated datasets risks overemphasizing Western-centric concepts and knowledge. Cultural bias can reduce the dataset's practical effectiveness (and conceptual relevance) as a global benchmark when translated. For example, the original English MMLU dataset contains several subsets which are US-specific, such as examinations in \textit{US History}, \textit{US Accounting}, and \textit{US Law}. Such cultural bias reduces the dataset's practical effectiveness (and conceptual relevance) as a global benchmark when translated. Furthermore, as these translated datasets become adopted for multilingual evaluation and developers optimize models for performance on \emph{transMMLU} datasets, we risk overfitting to the datasets' cultural biases and incidentally setting multilingual evaluation standards to be aligned with certain culture paradigms. Second, while machine translation expands language coverage, it also introduces practical evaluation challenges. Translation artifacts known as \textit{translationese}~\citep{bizzoni-etal-2020-human,vanmassenhove-etal-2021-machine,koppel2011translationese} can be introduced, which causes a breakdown in evaluation quality. Automatic data curation is also known to often exacerbate common data quality issues~\citep{luccioni-viviano-2021-whats,kreutzer-etal-2022-quality,Ferrara_2023,caswell-etal-2020-language}. 

Our effort to address the above is twofold. We conduct an extensive evaluation to quantify the impact of cultural biases in MMLU on model evaluations to-date \emph{and} contribute improvements to the overall translation quality to solve linguistic qualms. We hire professional annotators to verify translation quality and include improvements from rigorous per-question post-edits as well as human translations. We release the comprehensive improved dataset \gmmlu{} for 42 languages: 

\colorbox{blue!20}{\parbox{16cm}{
   Amharic, Arabic, Bengali, Chinese, Czech, Dutch, English, Filipino, French, 
   German, Greek, Hausa, Hebrew, Hindi, Igbo, Indonesian, Italian, Japanese, Korean, Kyrgyz, Lithuanian, Malagasy, Malay, Nepali, Nyanja, Persian, Polish, Portuguese, Romanian, Russian, Serbian, Sinhala, Somali, Shona, Spanish, Swahili, Swedish, Telugu, Turkish, Ukrainian, Vietnamese, and Yoruba.
}}

To address regional and cultural biases, we systematically annotate a subset of the original English MMLU to identify questions where correctly answering requires cultural, geographical, or dialect-specific knowledge. We refer to such questions as being \textit{Culturally-Sensitive} (\cs), in contrast to questions which do not require this prior knowledge, referred to as being \textit{Culturally-Agnostic} (\ca). We evaluate 14 state-of-the-art open-weight and proprietary models from 9 model families, focusing on those known for their high multilingual performance. This enables rigorous evaluation of how such models serve diverse language users and isolates how ranking may be subverted by questions which require primarily Western-centric knowledge. Through extensive evaluations, we consistently find that \emph{cultural sensitivity} has a significant impact on model rankings. Our core contributions can be enumerated as follows:
\begin{itemize}
    \item \textbf{Analysis of MMLU for cultural biases}: We observe that progress on MMLU depends heavily on learning Western-centric concepts. Out of the annotated sample, we found that 28\% of questions require specific knowledge of Western cultures. Moreover, for questions requiring geographic knowledge, an astounding 84.9\% focus on either North American or European regions. 
    \item \textbf{Introducing \gmmlu }: We release a new multilingual MMLU test set spanning 42 languages, including English. This dataset combines professional translations with post-edits (14 languages), crowdsourced translations (11 languages), and machine translations (16 languages). By integrating this dataset with our cultural bias study, evaluations can now report on both the \cs{} and \ca{} subsets. Additionally, we introduce \gmmlulite{} that provides a compact but high-quality alternative for multilingual evaluation.
    \item \textbf{Re-evaluation of state-of-the-art models}: We evaluate the impact of the re-annotated dataset on the relative performance of multilingual models. Among the 14 models tested, rankings on \ca{} datasets exhibited an average of 3.4 rank changes and 3.7 position shifts compared to their performance on a uniform subsample of the MMLU dataset (\textit{MMLU Annotated}\memoemoji). However, \cs{} datasets showed significantly greater variability, with an average of 5.7 rank changes and 7.3 position shifts across all languages.
    \item \textbf{Role of data quality improvements}: Our analysis highlights notable performance differences between human-translated and machine-translated datasets for both high-resource and low-resource languages. Human-translated datasets are essential for accurately assessing model performance, especially on low-resource languages, as relying solely on machine-translated data may obscure the true capabilities of models in these contexts. Without access to high-quality human-translated or in-language datasets, the evaluation of low-resource language performance remains uncertain.

\end{itemize}

\begin{figure}[t]
     \centering\includegraphics[width=0.9\textwidth]{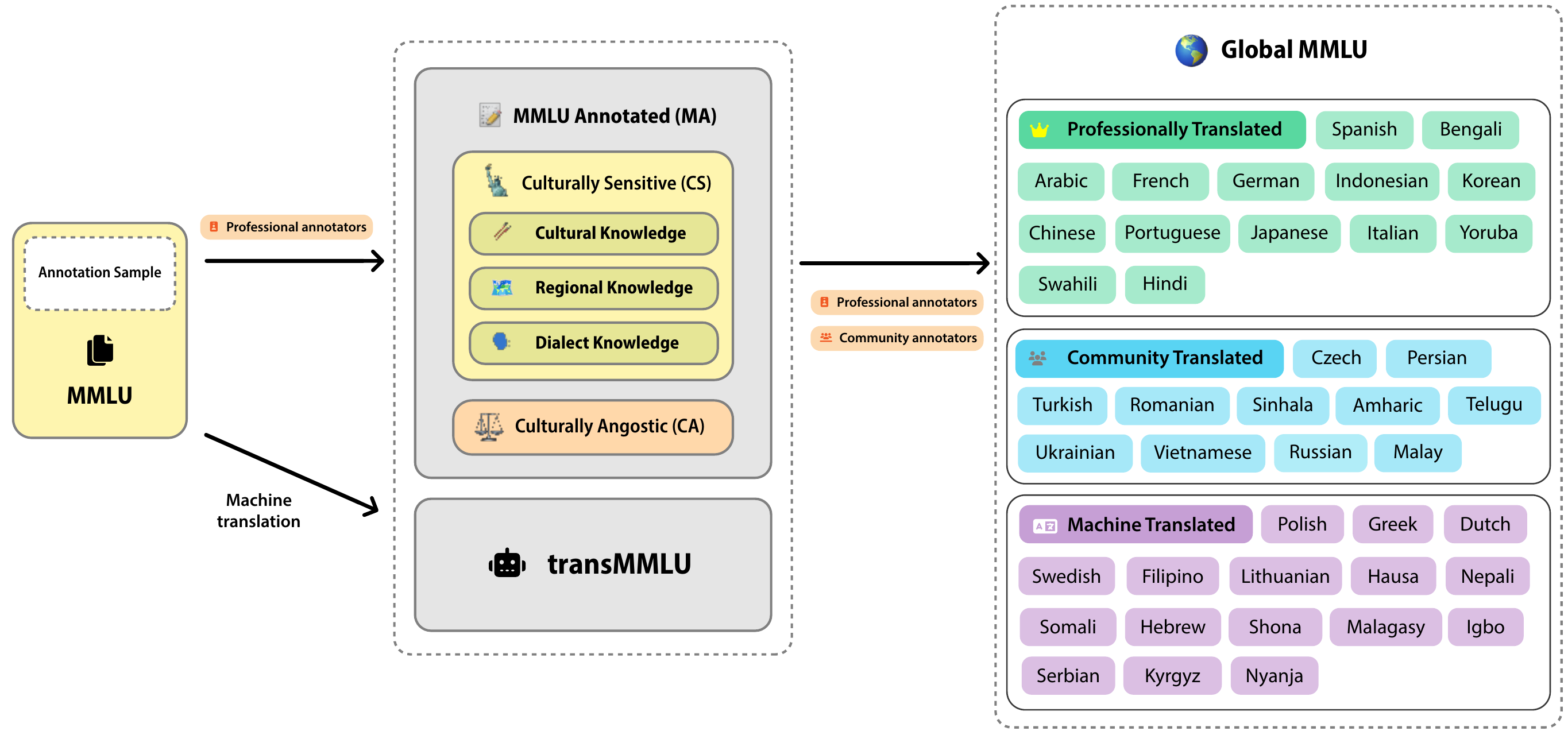}
     \caption{Overview of \textbf{Global-MMLU} \earthmoji{} preparation process. We engage with professional and community annotators to improve the quality of translated MMLU. Additionally, we engage in extensive annotation to provide rich meta-data for what questions in MMLU require \textit{Culturally-Sensitive} (\cs) knowledge such as  1) \textbf{Cultural Knowledge \chopsticksmoji}, 2) \textbf{Geographical Knowledge \worldmapsmoji} or 3) \textbf{Dialect Knowledge \speakersmoji} to answer correctly. We release this improved \textbf{Global-MMLU}\earthmoji{} alongside extensive metadata annotations.
     }
     \label{fig:mmlu_approach_overview}
\end{figure}

Stemming from our comprehensive results, we make the following recommendations for multilingual evaluation of generative models:

\begin{itemize}
    \item \textbf{Report on \gmmlu, instead of translated MMLU.}  We recommend prioritizing \gmmlu{} over translated versions of MMLU for multilingual evaluation. With its extensive language coverage and improvements based on professional annotations and post-edited translations, \gmmlu{} provides a more reliable and accurate benchmark for assessing model performance across diverse languages.
    \item \textbf{Report performance on culturally-sensitive and culturally-agnostic subsets separately.} Our analysis demonstrates significant variability in model rankings between \ca{} and \cs{} datasets, with \cs{} subsets showing greater variability. This variability, especially pronounced for low-resource languages and smaller models, highlights the importance of evaluating these subsets independently. We recommend reporting performance on \ca{} and \cs{} subsets separately to provide a clearer understanding of model capabilities and better address the unique challenges posed by cultural and linguistic nuances in \cs{} tasks.

\end{itemize}

\section{Evaluating cultural bias in MMLU}
\label{sec:analysis}

\subsection{Data Annotation Process}
\label{sec:annotation}

\begin{figure}[ht!]
    \centering\includegraphics[width=0.99\textwidth]{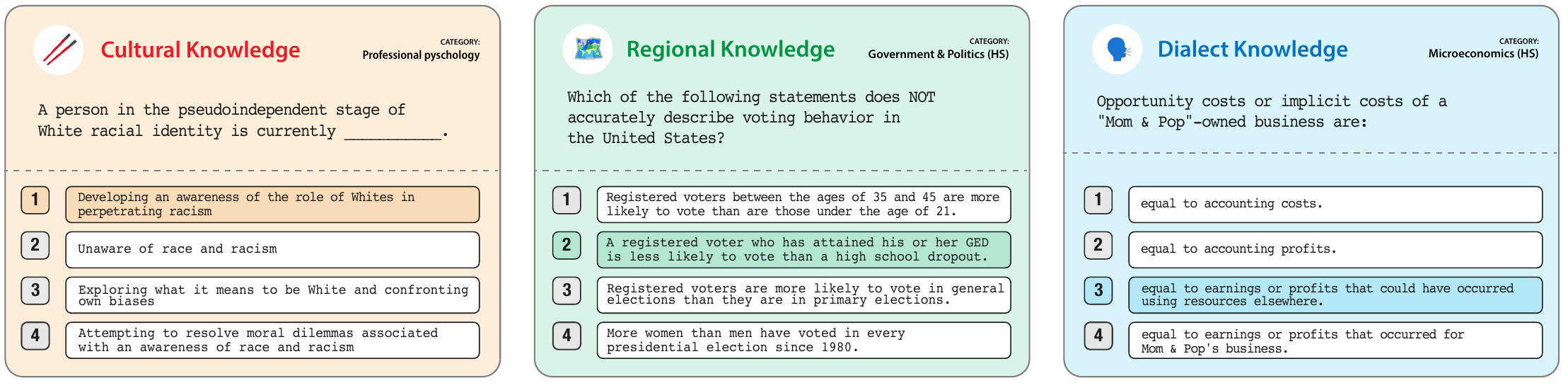}
    \caption{Examples of questions from MMLU dataset labelled as requiring cultural, regional or dialectal knowledge.}
    \label{fig:mmlu_CS_CA_examples}
\end{figure}

The goal of this work is to study how cultural biases in translated datasets influence the performance of widely-used multilingual models. To achieve this, we worked with 200 professional compensated and community annotators to review MMLU questions \textbf{from the original English MMLU dataset} to assess its cultural sensitivity. Annotators were presented with a representative random sample from each of the 57 exam subjects that compose MMLU (50 per subject), totaling 2,850 samples. This annotated set is referred to as \emph{MMLU Annotated} (MA) \memoemoji{} throughout the paper. Annotators were asked to identify questions where correctly answering depended upon 1) cultural knowledge \chopsticksmoji, 2) geographic knowledge \worldmapsmoji{} 
 or 3) dialect knowledge \speakersmoji. We provide more context about each of these categories below: 

\begin{itemize}
\item \textbf{Cultural Knowledge \chopsticksmoji.}
Annotators evaluated whether answering a question required culture-specific knowledge. If so, they selected the relevant culture from a drop-down menu with options: Western Culture, Eastern Asian Culture, Middle Eastern Culture, South Asian Culture, African Culture, Latin American Culture, or Other. Cultural knowledge encompasses recognizing and appreciating the beliefs, values, customs, and artistic expressions of a particular group, shaped by shared traditions and heritage~\citep{kipuri2009chapter,liu2024culturally,mukherjee2024cultural}.
\item \textbf{Geographical or Regional Knowledge \worldmapsmoji.} Geographical knowledge refers to understanding characteristics tied to specific regions, such as natural landmarks or environmental features. Annotators determined whether answering correctly required region-specific knowledge. If applicable, they identified the relevant region from a drop-down menu with the following options: North America, South America, Europe, Asia, Africa, Australia and Oceania, and Antarctica.
\item \textbf{Dialect Knowledge \speakersmoji.} This category involves recognizing distinctive language variations or speech patterns used by people from specific regions or communities in English. It includes slang terms, idiomatic expressions, and pronunciation differences that distinguish regional speech from standardized forms of language. Notably, this assessment was conducted on the original English sentences. Therefore, it specifically addresses variations in English dialects or regional vocabulary, rather than any nuances that might arise during the translation process.
\end{itemize}

\begin{figure}[ht!]
    \centering\includegraphics[width=0.99\textwidth]{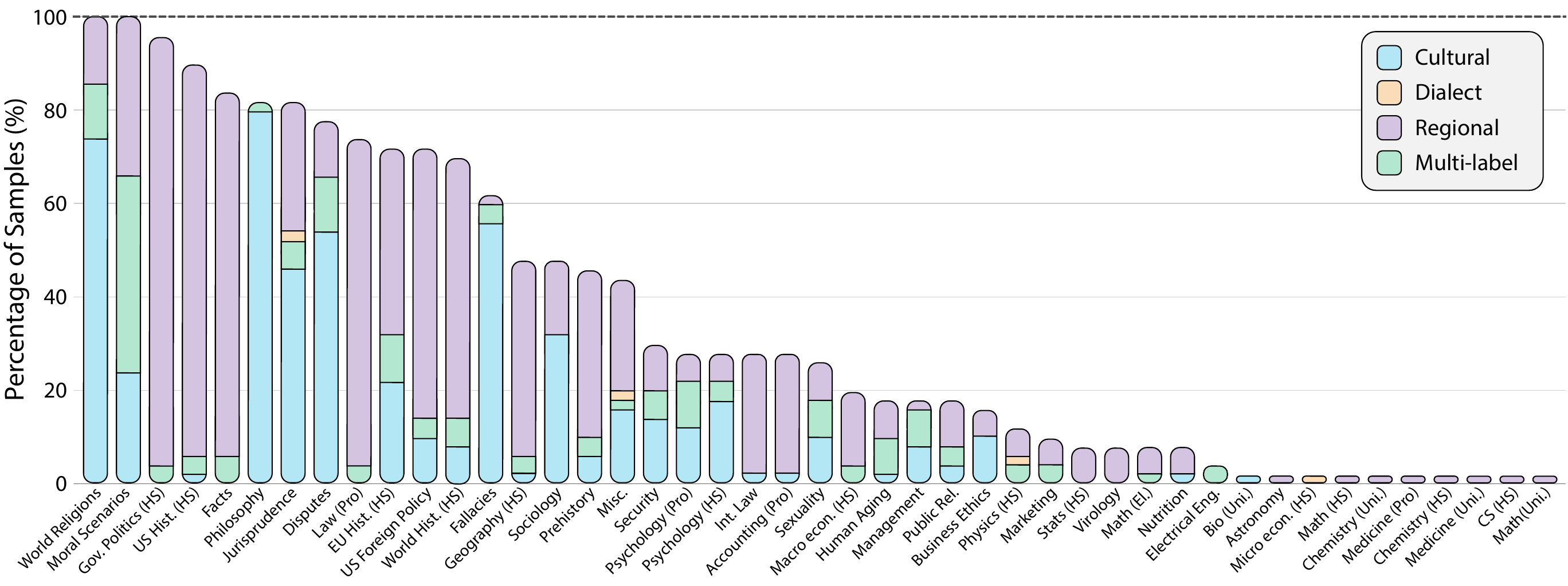}
    \caption{Proportion of samples containing cultural, regional, or dialect-specific references per subject in the MMLU dataset. Notably, all samples in the \emph{World Religions} and \emph{Moral Scenarios} subjects include at least one such reference. Note that 12 subjects did not contain any Culturally-Sensitive \cs{} samples and have been excluded from the figure.
    }
    \label{fig:mmlu_cul_reg_dia_ref}
\end{figure}

Figure~\ref{fig:argilla_annotation_ui_phase1} in Appendix~\ref{app:annotation_process} illustrates the annotation interface used during this process. Annotators were presented with questions one at a time from each of the 57 MMLU subjects and had to analyze and label them for the presence of cultural, geographic, dialect knowledge. Each data point was reviewed by at least three annotators, and some data-points had a maximum of 10 annotators. 96.4\% of all data points were reviewed by more than 3 human annotators. 
We classify each question as presenting cultural, geographic and dialect sensitivity according to majority vote among annotators who reviewed each data point~\citep{Feldman1980}. If half or more of the annotators apply the same tag to a question, it is categorized under that tag. Detailed information about the annotators and the annotation process is available in Appendix~\ref{app:annotation_process}.

We also asked annotators to annotate for temporal knowledge to determine if answers for questions change with time. We find that only 2.4\% of annotated samples depend on temporal knowledge. We provide more details about temporal analysis in the Appendix~\ref{app:time-sensitivity}.  

To understand the prevalence of these attributes at an aggregate level, we also assign a label of \textbf{Culturally-Sensitive} (CS\statuelibertyemoji) if either \textbf{Dialect Knowledge} \speakersmoji, \textbf{Cultural Knowledge \chopsticksmoji} or \textbf{Geographic Knowledge \worldmapsmoji } are positively attributed to an example. If none of these properties are present, we deem an example to be \textbf{Culturally-Agnostic} (CA\scalesemoji). This enables us to track at an aggregate level the fraction of the entire MMLU that requires \cs{} knowledge.

\subsection{Analysis of MMLU Cultural Biases}
\label{sec:analysis_processing}

Figure~\ref{fig:mmlu_cul_reg_dia_ref} summarizes the results of this extensive annotation process. Our analysis reveals that 28\% of MMLU requires \cs{} knowledge -- defined as requiring knowledge of either geographic knowledge \worldmapsmoji, cultural knowledge \chopsticksmoji{} or dialect knowledge \speakersmoji{} -- to be answered correctly. 
Among these, geographic knowledge \worldmapsmoji{} emerges as the most frequently tagged bias, representing 54.7\% of all \cs{} questions. Cultural knowledge \chopsticksmoji{} follows at 32.7\%, while dialect-specific knowledge \speakersmoji{} accounts for a mere 0.5\% of all questions. Additionally, 10.6\% of questions require both cultural and geographic knowledge, and 1.5\% involve a combination of all three types of nuanced knowledge.

\begin{figure}[ht!]
    \centering\includegraphics[width=1\textwidth]{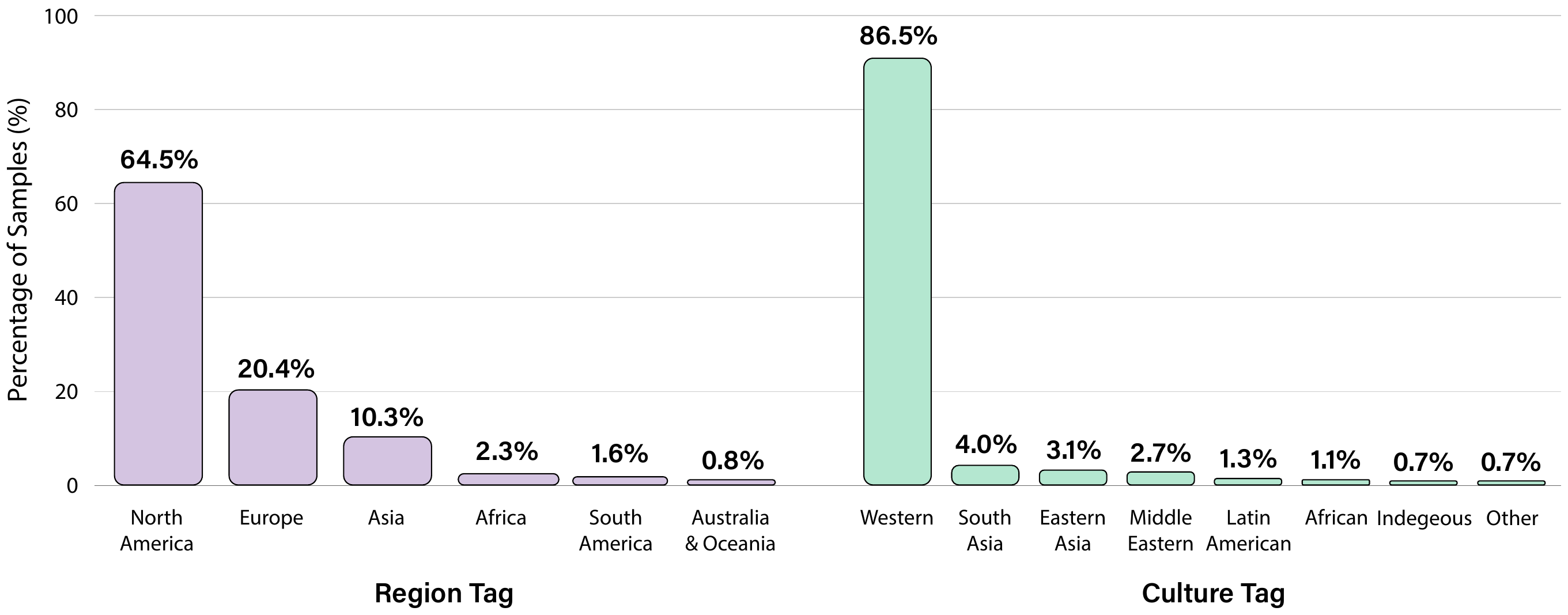}
    \caption{Distribution of region (left) and culture (right) categories found in \cs{} dataset. The majority of Region tags (64.5\%) correspond to North America, while the majority of Culture tags (86.5\%) are classified as Western. We have excluded samples that do not contain any region or culture tags or contain multiple region or culture tags from this figure.}
    \label{fig:mmlu_samples_cultures_regions}
\end{figure}

\textbf{Western-centric culture dominates.} Among the samples identified as requiring culturally sensitive \cs{}, a significant 86.5\% were tagged as specific to \textit{Western} cultural knowledge. In contrast, the next closest category, \emph{South Asian} cultural knowledge, accounted for only 4\% of the cultural tags.
As Figure~\ref{fig:mmlu_samples_cultures_regions} shows, Latin American, African and Indigenous cultures are represented by 1.3\%, 1.1\% and 0.7\% of the tags, respectively. This shows performing well on MMLU heavily depends on mastering Western-centric cultural knowledge.

A similar trend is observed for geographic knowledge: 64.5\% of \cs{} samples were tagged as needing regional knowledge of \textit{North America}, followed by 20.4\% tagged as requiring regional knowledge of \textit{Europe}. This concentration indicates that progress on MMLU predominantly reflects knowledge of Western concepts and regions.

\begin{figure}[ht!]
    \centering\includegraphics[width=1\textwidth]{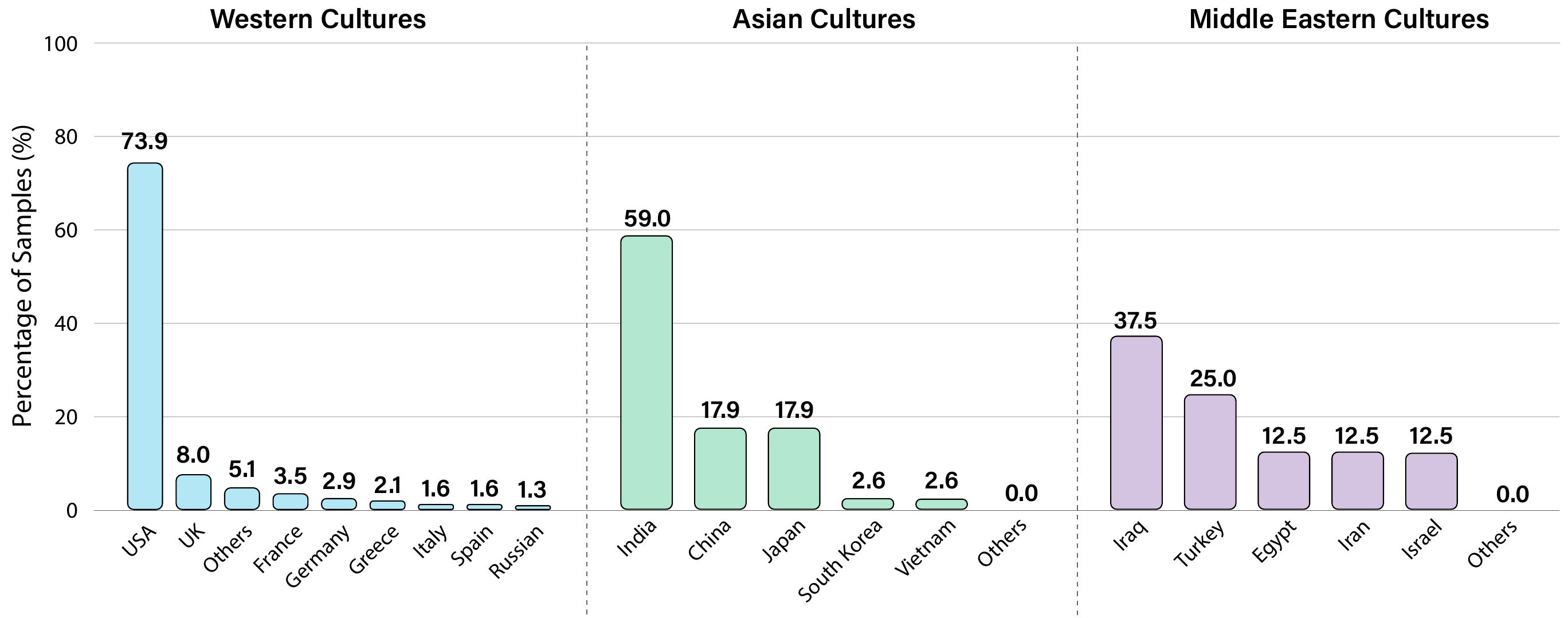}
    \caption{Distribution of cultural and regional tags across countries in the \cs{} dataset. The percentages indicate the representation of each country within the dataset. We have excluded samples that do not contain any country tags or contain multiple country tags from this figure.}
    \label{fig:mmlu_culture_country_1}
\end{figure}

\textbf{Culture-specific knowledge is overfit to a few countries.} 
Figure~\ref{fig:mmlu_culture_country_1} illustrates the distribution of cultural and regional tags across countries within the \cs{} dataset. 
Our analysis reveals that 73.9\% of questions related to Western culture require knowledge about the United States, followed by the United Kingdom at 8\%, with smaller contributions from countries like France and Germany. In contrast, Asian culture tags are predominantly associated with India, accounting for 59\%, while China and Japan represent only 17.9\% each of the questions requiring knowledge of Asian culture. Despite this, the overall representation of Asian cultures remains limited, with only 4.0\% of questions pertaining to South Asia and 3.1\% to East Asia in the MMLU dataset. Similarly, Middle Eastern culture is largely represented by Iraq (37.5\%) and Turkey (25\%), yet its overall presence in the dataset is minimal, with just 2.7\% of questions addressing Middle Eastern cultural knowledge. These findings highlight the dataset's strong bias toward the United States, with a significant portion of cultural tags tied to the U.S. For further analysis of the culture--region relationship and detailed country-level insights, see Appendix~\ref{app:culture-region}.

\textbf{Cultural sensitivity varies considerably across subjects.} The MMLU dataset, introduced by \citet{hendrycks2020measuring}, includes 57 subjects spanning four categories: \emph{STEM, Humanities, Social Sciences}, and \emph{Other}.
From the \emph{Other} category, we selected relevant subjects and further categorized them into \emph{Medical} \citep{chen2023meditron70bscalingmedicalpretraining} and \emph{Business}. 
Additional details about this categorization are provided in Appendix~\ref{app:mmlu_subject_categories}.

Figure~\ref{fig:filtered_mmlu_samples} illustrates the data distribution for the \textbf{CA}\scalesemoji{} subset, revealing significant variation in cultural and regional references between different MMLU subjects and subject categories. Questions from categories in \emph{Humanities} and \emph{Social Sciences} frequently required cultural or regional knowledge, while those from the \emph{STEM} and \emph{Medical} categories generally did not.  Overall for \emph{Humanities}, 68\% of all questions were tagged as \textbf{CS}\statuelibertyemoji. However, this bias was even more pronounced for certain subjects within \emph{Humanities}. Notably, more than 80\% of samples for subjects like Philosophy, Moral Scenarios\footnote{Morals might share universal truths and moral decisions may be well-defined given an underlying belief system, but this does not seem to be the case in this scenario. That is, we observe that Moral Scenarios in MMLU are geared towards Western Culture, and therefore \cs{} knowledge, as it specifies ``moral standards in the US'' in the instruction.}, High School US History, and High School Government and Politics were deemed \textbf{CS}\statuelibertyemoji. Within the \emph{STEM} category, only 30 out of 950 samples (3.15\%) were identified as \cs, and for subjects such as Clinical Knowledge, Computer Security, and Econometrics all question examples were classified as \ca. These findings, detailed in Figure~\ref{fig:filtered_mmlu_samples}, unsurprisingly reveal that certain subjects inherently exhibit more cultural or regional biases. We provide examples of MMLU questions annotated as \textbf{CS}\statuelibertyemoji{} (Culturally Sensitive) and \textbf{CA}\scalesemoji{} (Culturally Agnostic) in the Appendix~\ref{app:ma_data_examples}.

\textbf{Inter-annotator agreement.} Each data point was  reviewed by at least three annotators, and some datapoints had a maximum of 10 annotators. 96.4\% of all data points were reviewed by more than 3 human annotators. Given this rich set of feedback on each data point, we analyze the agreement between ratings from different annotators using \emph{Krippendorff's Alpha} scores~\citep{krippendorff2004reliability}.  We observed high inter-annotator agreement across most subjects, with a unanimous cultural sensitivity agreement in the \emph{Anatomy} subject. Six subjects showed disagreement including High-school US History, while Moral Scenarios showed the most disagreement. Detailed results are presented in Figure~\ref{fig:alpha_krippendorff_score1} and ~\ref{fig:alpha_krippendorff_score2} in Appendix~\ref{app:annotator_agreement}.

\begin{figure}[ht!]
    \centering\includegraphics[width=1\textwidth]{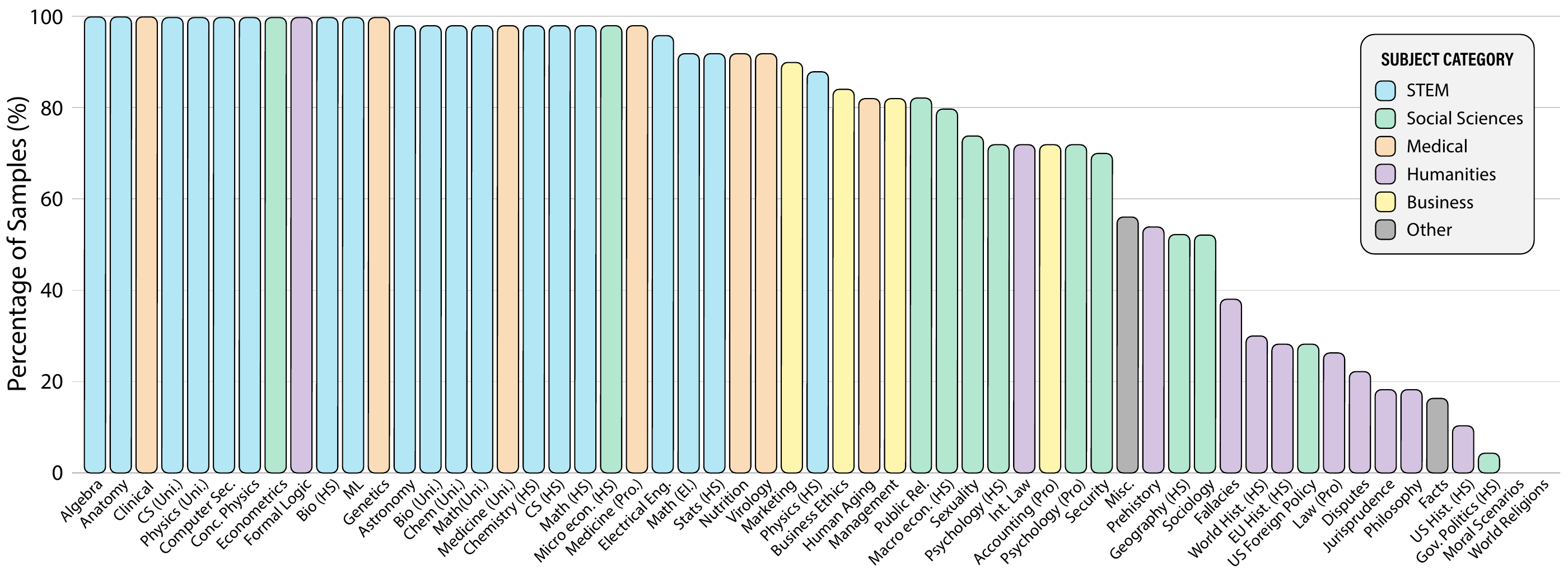}
    \caption{Proportion of samples retained per subject, after excluding those requiring cultural, geographic and dialectic knowledge (selected based on majority agreement).}
    \label{fig:filtered_mmlu_samples}
\end{figure}

\begin{table}[ht!]
\centering
\scalebox{0.88}{
\begin{tabular}{l|ccc|ccc|rrr}
\toprule
                & \multicolumn{3}{c|}{\textbf{Number of Subjects}} & \multicolumn{3}{c|}{\textbf{Number of Samples}}   & \multicolumn{3}{c}{\textbf{Data Proportion}}   \\ 
\toprule
\textbf{Categories}  & MA\memoemoji  &  CS\statuelibertyemoji  &  CA\scalesemoji  &  MA\memoemoji   & CS\statuelibertyemoji   & CA\scalesemoji    &   MA\memoemoji  &  CS\statuelibertyemoji   & CA\scalesemoji  \\
\midrule
STEM            & 19   & \textcolor{red}{11}    & 19      & 950      & 23     &  \textbf{927}    &  33.3\% &  2.9\% \tcbox[colback=LightRed]{\textcolor{black}{$\downarrow$}} & 45.0\%  \tcbox[colback=LightGreen]{\textcolor{black}{$\uparrow$}}  \\
Humanities      & 13   &  \textcolor{red}{12}  & \textcolor{red}{11}    & 650     & \textbf{442}   & 208   & 22.8\% &  55.8\% \tcbox[colback=LightGreen]{\textcolor{black}{$\uparrow$}}  & 10.1\%  \tcbox[colback=LightRed]{\textcolor{black}{$\downarrow$}}  \\
Social Sciences & 12    & \textcolor{red}{11}  & 12     & 600      & 208     &  \textbf{392}   &  21.1\%  &  26.3\% \tcbox[colback=LightGreen]{\textcolor{black}{$\uparrow$}}  & 19.1\%   \tcbox[colback=LightRed]{\textcolor{black}{$\downarrow$}}  \\
Medical         & 7     & \textcolor{red}{5}   & 7       & 350    & 19   &  \textbf{331}   & 12.3\% & 2.4\% \tcbox[colback=LightRed]{\textcolor{black}{$\downarrow$}} &  16.1\%  \tcbox[colback=LightGreen]{\textcolor{black}{$\uparrow$}} \\
Business        & 4        &4         & 4           & 200        & 36    &  \textbf{164}    &  7.0\%  & 4.5\% \tcbox[colback=LightRed]{\textcolor{black}{$\downarrow$}} &  8.0\%  \tcbox[colback=LightGreen]{\textcolor{black}{$\uparrow$}}  \\
Other           & 2        &2        & 2           & 100      &   \textbf{64}     & 36     & 3.5\%  & 8.1\% \tcbox[colback=LightGreen]{\textcolor{black}{$\uparrow$}}  &  1.8\%  \tcbox[colback=LightRed]{\textcolor{black}{$\downarrow$}} \\
\bottomrule
\end{tabular}}
\caption{Statistics for \textbf{MA}\memoemoji, \textbf{CS}\statuelibertyemoji, and \textbf{CA}\scalesemoji{} datasets. The left column displays the number of subjects included in each dataset, the middle column shows the total number of samples per category, and the right column illustrates changes in subject category distributions relative to \textbf{MA}\memoemoji, with arrows indicating increases or decreases in representation.}
\label{tab:data_statistics}
\end{table}

\textbf{Characteristics of \cs{} versus \ca{} subsets.}
Our extensive annotation process resulted in two aggregated annotated subsets of MMLU: \cs{}, which includes all questions labeled as requiring dialect knowledge \speakersmoji, cultural knowledge \chopsticksmoji, or geographic knowledge \worldmapsmoji{} to answer correctly, and \ca{}, comprising questions that do not require knowledge from these categories. Table~\ref{tab:data_statistics} provides a detailed breakdown of the number of subjects and samples in the \textbf{CS}\statuelibertyemoji{} and \textbf{CA}\scalesemoji{} subsets.

We observe notable differences in subject distribution between the \ca{} and \cs{} subsets, leading to shifts in category representation. For instance, while questions from the \emph{Social Sciences} category make up 21.1\% of the MMLU Annotated \memoemoji, a uniformly balanced subsample of the original MMLU, they are over-represented in \cs{}, accounting for 26.3\% of all questions requiring \cs{} knowledge. Conversely, questions from the STEM category, which contribute 33.3\% of the MMLU Annotated \memoemoji, are under-represented in \cs{}, making up only 2.9\% of all questions identified as requiring \cs{} knowledge. These shifts reflect how the nature of the \cs{} subset emphasizes cultural and contextual knowledge over technical or scientific content.

Overall, the proportions of STEM, Medical, and Business categories increase in the \ca{} subset due to their globally relevant content. Conversely, Humanities and Social Sciences are over-represented in the \cs{} subset compared to the original MMLU, as these fields frequently include cultural or regional references. These findings are critical to the model evaluations in Section~\ref{sec:evaluation}, illustrating how cultural references in MMLU influence dataset composition and, ultimately, model performance.

\section{Introducing \textbf{Global-MMLU} \earthmoji}
\label{sec:tranlsation_and_data_mix}

To date, many multilingual evaluations have relied on translated MMLU with the most widely adopted \emph{existing multilingual MMLU translation} dataset being translated into 26 languages using ChatGPT\footnote{\url{https://chat.openai.com/chat}} supported by GPT-3.5~\citep{lai2023okapi}. We release an improved \gmmlu{} benchmark which is both of higher quality and also supports analysis on both \cs{} and \ca{} subsets. 

Here, we improve quality by incorporating professional edits and translations from native speakers for a subset of languages and expanding coverage to 42 languages. We achieve this through a combination of paid professional translations, community contributions, and higher-quality machine translation. This effort involved professionally compensated annotators for four gold-standard languages and a broader pool of community annotators who contributed to translations in 11 additional languages. Where available, we also included the professional human translations from the MMMLU dataset\footnote{\url{https://huggingface.co/datasets/openai/MMMLU}} for 14 languages. We rely as much as possible on human-verified translations to ensure that the translations are reliable and minimize the biases introduced, specifically \emph{translationese} which might be more pronounced in Machine Translation~\citep{bizzoni-etal-2020-human,vanmassenhove-etal-2021-machine,koppel2011translationese}. Alongside these quality improvements through human verification, we include the metadata for the \cs{} and \ca{} annotations developed in the previous sections to allow for analysis on all subsets of data.
Below, we provide further details about our efforts to improve the quality of MMLU and engage compensated human annotators in translating and verifying quality as well as identifying the \cs{} and \ca{} subsets. 

\subsection{Translation Process}
\label{sec:translation_process}

\begin{figure}[ht!]
    \centering\includegraphics[width=0.8\textwidth]{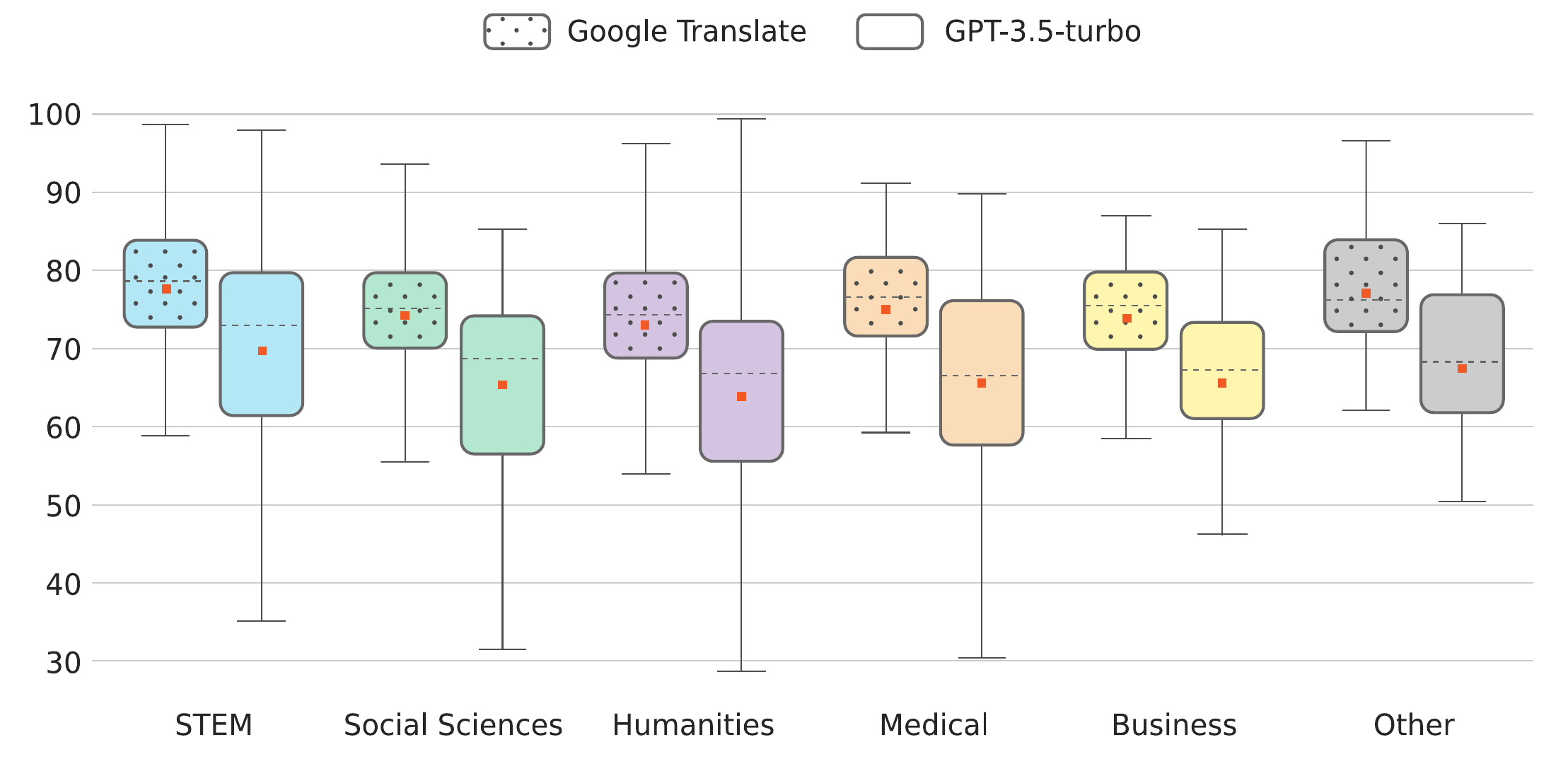}
    \caption{ChrF++ scores for Google Translate and GPT-3.5-Turbo}
    \label{fig:mmlu_gtranslate_gpt3.5_comparison}
\end{figure}

We first translated the English MMLU dataset into 41 languages using the Google Translate API.\footnote{\url{https://cloud.google.com/translate}} 
Despite its cost, we chose to use Google Translate because comprehensive evaluations spanning 102 languages~\citep{zhu-etal-2024-multilingual} demonstrate that Google Translate significantly outperforms alternatives such as NLLB~\citep{nllbteam2022language}, GPT-4, and ChatGPT, on low-resource languages~\citep{robinson2023chatgpt}. Recent work \citep{kocmi2024findings} have shown that LLMs have begun to surpass popular online translation tools like Google Translate for machine translation on specific high-resource languages. However, given that there is a known tendency for models to favor their own generations \citep{panickssery2024llmevaluatorsrecognizefavor,shimabucoro-etal-2024-llm}, we decided to use Google Translate for every language in order to avoid introducing bias into model evaluations.  To empirically validate this choice, we compared Google Translate's outputs with translations performed by GPT-3.5-turbo, which had been previously used to translate the MMLU dataset~\citep{lai2023okapi}. As shown in Figure~\ref{fig:mmlu_gtranslate_gpt3.5_comparison}, we find that Google Translate achieved higher ChrF++ scores~\citep{popovic-2017-chrf} across all subjects and lower deviation in performance across languages, consistent with the findings of previous research \citep{popovic-2017-chrf} about its superiority in translation quality.
Following the translation process, native speakers reviewed and edited the translations to ensure accuracy and fluency, thereby enhancing global representation. These edits were performed by two types of annotators: \emph{professional annotators} and \emph{native community annotators}.

\textbf{Professional Annotators.} We hired compensated professional annotators for four languages: \textit{Arabic}, \textit{French}, \textit{Hindi}, and \textit{Spanish}. These annotators reviewed the machine translations to ensure fluency and cultural appropriateness, making edits where necessary. We refer to this set of translation as our \emph{``Gold Set''}. We include more details about compensated annotation process in the Appendix \ref{app:annotators_gold_standard}.

\textbf{Community Annotators.} In addition to professional annotations for a subset of languages, we also facilitated community contributions to verify translation quality across a broader range of languages, focusing on fluency edits and correcting poor translations. This participatory research approach~\citep{Birhane_2022,corbett2023,Delgado2023ThePT,singh-etal-2024-aya,ustun2024ayamodelinstructionfinetuned} involved collaboration across multiple institutions globally. Such cross-sectional efforts are crucial for gathering linguistic data at scale and fostering community engagement—both essential for developing inclusive language technologies~\citep{joshi-etal-2019-unsung,nekoto2020participatory,singh-etal-2024-aya,romanou2024includeevaluatingmultilinguallanguage}. We established a criterion requiring a minimum of 50 human-translated samples for each language before its inclusion in \gmmlu. This threshold was met by eleven languages: \emph{Amharic, Czech, Malay, Persian, Romanian, Russian, Sinhala, Telugu, Turkish, Ukrainian}, and \emph{Vietnamese}. In the following sections, we refer to this set of languages as \emph{``Community Translated''}.

The participation of native speakers from diverse regions introduced logistical challenges in both data selection and quality control. To overcome these, we adopted Argilla\footnote{\url{https://github.com/argilla-io/argilla}} as our primary annotation platform. In line with our community-based approach, Argilla's collaborative features and customizable workflows enabled us to efficiently manage contributions from various regions while maintaining consistency in translation quality. Annotators were presented with both the original and machine-translated questions and answers, and were asked to edit any translations that did not accurately capture the intent of the original text. The translation interface is shown in Figure~\ref{fig:argilla_tranlation_ui_phase2} in Appendix~\ref{app:translation}.

\begin{figure}[ht!]
    \centering
    \includegraphics[width=0.8\linewidth]{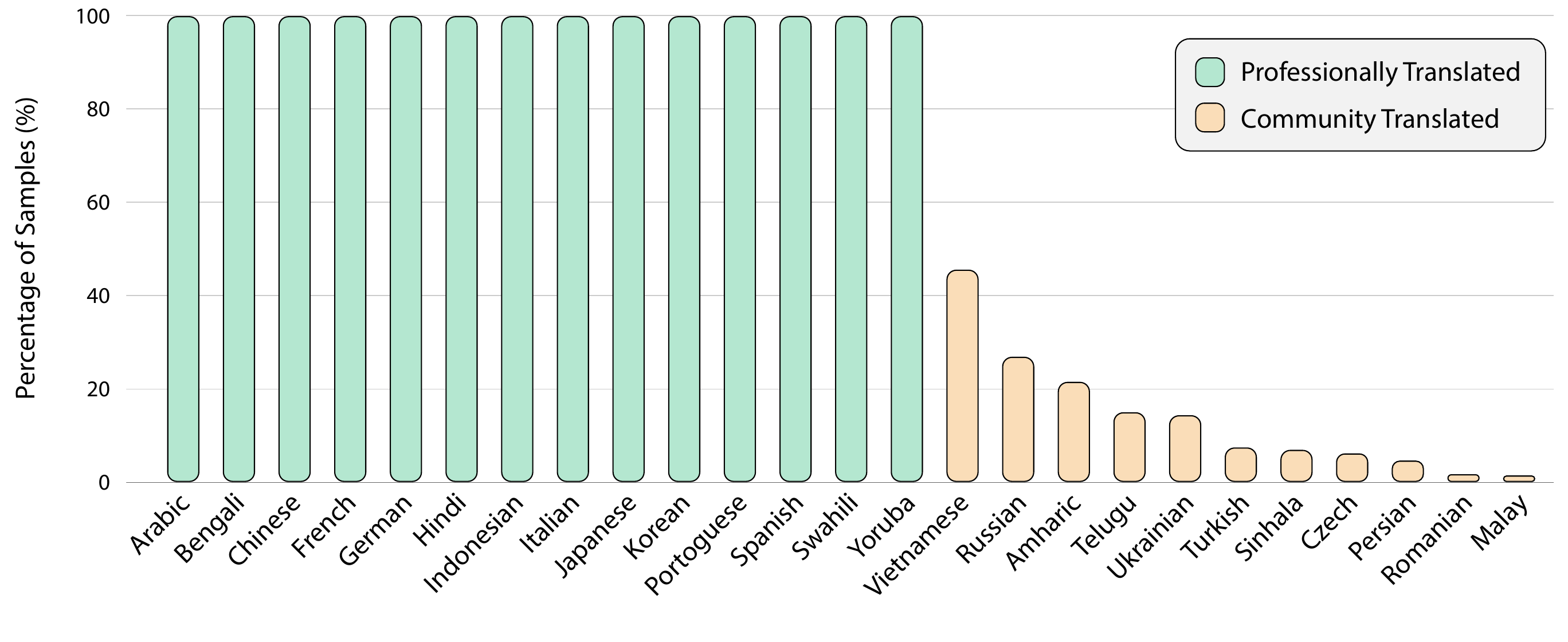}
    \caption{Percentage of Human-Translated Samples in MMLU Annotated \memoemoji.} 
    \label{fig:mmlu_agnostic_human_edited}
\end{figure}

\textbf{MMMLU Translations.} As detailed in the OpenAI-o1 system card,\footnote{\url{https://openai.com/index/openai-o1-system-card/}} MMMLU\footnote{\url{https://hf.co/datasets/openai/MMMLU}} is a professionally human-translated dataset released by OpenAI in 14 languages. To maximize the inclusion of human-translated content in \gmmlu, we incorporated this dataset wherever possible. Since MMMLU overlaps with our \emph{Gold Set}, we utilized the remaining 10 languages: \emph{Bengali, Chinese, German, Indonesian, Italian, Japanese, Korean, Portuguese, Swahili, Yoruba} from this dataset.

Figure~\ref{fig:mmlu_agnostic_human_edited} highlights the number of samples edited by professional annotators and community contributors. A total of 7,565 edits were made, accounting for 36.9\% of the samples reviewed. On average, professional annotators edited 789 samples per language (38.5\% of the total) in the \emph{Gold Set}, while community contributors edited 362 samples per language (17.7\% of the total).
It is important to note that the differences in edit rates likely reflect variations in time and resources available to professional versus community annotators, and cannot be interpreted as differences in translation quality across languages. Additional analyses of question and answer lengths, as well as edit distances across subject categories, are presented in Appendix~\ref{app:translation}.

\subsection{Data Composition of Global-MMLU \earthmoji}
\label{sec:data_comp_gmmlu}

{\gmmlu} is our comprehensive test set  encompassing all 14K samples from MMLU across 42 languages (including English), resulting in a total of 589,764 samples, created by integrating multiple data sources, including human-translated datasets, machine translations, and the original English MMLU.
Throughout the Model Evaluations section, we also report on different subsets of \gmmlu, described as follows:

\textbf{MMLU Annotated \memoemoji.} This subset consists of 2,850 question-answer pairs sampled at uniform from the MMLU dataset (50 questions per subject), representing 20\% of the original data and serving as a representative random sample. These samples are annotated in English to determine whether answering requires cultural, geographic, dialectal, or temporal knowledge. The annotations are then applied to corresponding samples in 41 other languages, resulting in a total of 119,700 samples.

\textbf{Culturally-Sensitive (CS) \statuelibertyemoji.} This subset contains samples identified as requiring dialect knowledge \speakersmoji, cultural knowledge \chopsticksmoji{} or geographic knowledge \worldmapsmoji{} to answer correctly. It includes 792 annotated samples in English based on majority voting by annotators. These annotations are extended to 41 additional languages, creating a dataset with 33,264 entries. This subset is particularly useful for evaluating model performance on culturally contextual tasks.

\textbf{Culturally-Agnostic (CA) \scalesemoji.} This subset includes samples that do not contain cultural, regional, or dialectal references. It serves as a baseline for evaluating models on tasks that do not require specific contextual knowledge. The subset consists of 2,058 annotated samples in English, which are extended to 41 languages for a total of 86,436 entries.

\gmmlulite{}. This is a ``lite'' version of {\gmmlu} covering 15 languages which are fully human translated or post-edited, along with English. It includes 200 CS and 200 CA samples per language, totaling 6,000 samples. Further details on its preparation are in Appendix~\ref{app:gmmlu_lite}.

\section{Model Evaluations}
\label{sec:evaluation}

One of the key findings from Section \ref{sec:analysis_processing} is that MMLU presents severe biases towards \cs{} knowledge. In this section, we seek to understand how these biases may have impacted evaluation of open-weights and closed models. 
To do so, we measure changes to model rankings on 3 subsets of data: \emph{\textbf{Global-MMLU Annotated} \memoemoji}, \emph{\textbf{Global-MMLU Culturally-Agnostic} (\ca)} and \emph{\textbf{Global-MMLU Culturally-Sensitive} (\cs)}. By comparing model performance across these three subsets, we aim to address the following questions: \emph{(1) How do models perform on the MMLU test set when it includes culturally-sensitive samples?} and \emph{(2) How do models perform on samples that do not require specific contextual knowledge, ensuring consistent and fair evaluations across different languages and regions?}

\subsection{Experimental Setup}
\label{sec:exp_setup}

We evaluated 14 recent state-of-the-art language models from 9 model families, focusing on those known for their high multilingual performance. These include \textbf{small models} like Aya Expanse 8B, Gemma2 9B, SEA-LION v3 (9B), Llama 3.1 8B, Mistral Nemo 12B, and Qwen 2.5 7B; \textbf{mid-size models}, comprising Aya Expanse 32B, CommandR (34B), Gemma2 27B, and Qwen 2.5 32B; \textbf{large models}, such as Llama 3.1 70B and CommandR+; and \textbf{closed-weight models}, specifically GPT-4o and Claude Sonnet 3.5. A more detailed description of the models covered is mentioned in the Appendix~\ref{app:models}.
\emph{We note that all these models do not claim to support the same set of languages, and none claim to support the full set of languages we cover.} 

\textbf{Evaluation Setup}. We use \emph{lm-evaluation-harness} \citep{eval-harness} to evaluate the open multilingual models in a 5-shot setting. For closed models (i.e., GPT-4o and Claude-Sonnet 3.5), we also do 5-shot evaluation. However, since log probabilities are not accessible via API for closed models, we send the 5-shot prompt via API and get the corresponding generation from the model. We use a system preamble to make the model respond with only the correct answer option and extract the answer from the output generation. For prompting, we follow the same approach as specified in \citep{hendrycks2020measuring} and use prompt instructions in the same language as the sample.

\textbf{Languages.} We categorize the languages into two main groups for reporting the results. The first group consists of human-translated data only, which covers 10 languages from OpenAI's human-translated MMLU test set and 4 additional languages from our professionally translated set. The second group contains all our data (combining professional, community and machine translations), organized by language resource availability \highemoji{}high-resource, \midemoji mid-resource, and \lowemoji low-resource languages as defined by \citet{joshi-etal-2019-unsung} and categorized in~\citep{singh-etal-2024-aya}.  
We report results for each of these categories. The \emph{\textbf{high-resource}} languages are Arabic, Chinese, Czech, Dutch, English, French, German, Hindi, Italian, Japanese, Persian, Polish, Portuguese, Russian, Spanish, Swedish, Turkish, Vietnamese, \emph{\textbf{mid-resource}} languages are Bengali, Filipino, Greek, Hebrew, Indonesian, Korean, Lithuanian, Malay, Romanian, Serbian, Ukrainian and \emph{\textbf{low-resource}} languages are Amharic, Hausa, Igbo, Kyrgyz, Malagasy, Nepali, Nyanja, Shona, Sinhala, Somali, Swahili, Telugu, Yoruba.

\subsection{Results}
\label{sec:results}

\textbf{Evaluations on Human-Translated Data.}
To assess the performance of models on high-quality, human-translated data, we conducted evaluations using the subset of 14 languages with human-translated data. The analysis focuses on both the \ca{} and \cs{} subsets to explore how models handle tasks with and without cultural context.

\begin{figure}[ht!]
    \centering
    \includegraphics[width=1.0\linewidth]{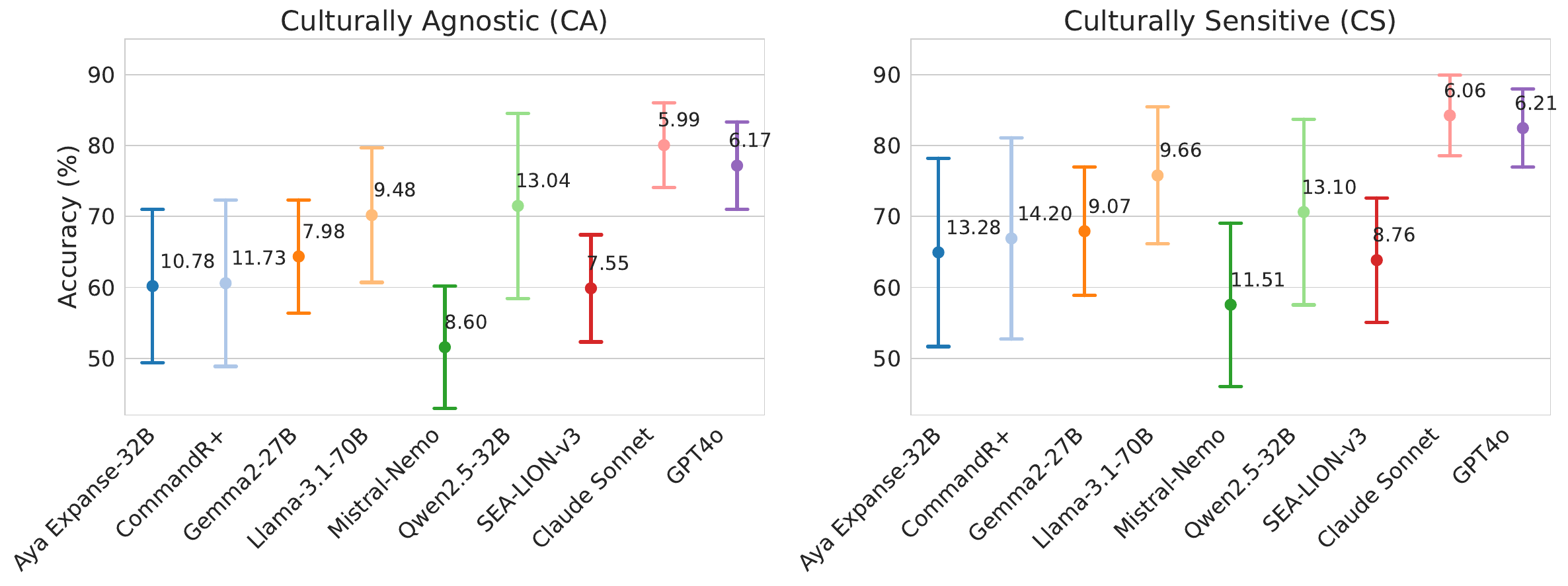}
    \caption{Model evaluations on \ca{} and \cs{} data samples on \textbf{human-translated 14 languages}. The error bars indicates the standard deviation across languages.}
    \label{fig:model_evaluations_on_human_edited_data}
\end{figure}

We evaluated 14 models from 9 different model families, including 2 closed-source models. Figure~\ref{fig:model_evaluations_on_human_edited_data} presents the results aggregated across 14 languages. 
We note that the focus of this evaluation is not to compare model performances directly but to analyze their behaviors on \ca{} and \cs{} datasets. Direct comparisons between proprietary models and open-weight models are not feasible due to significant differences in model sizes (although we note that the parameter sizes of proprietary models have not been officially disclosed) and different evaluation methods. 
Nonetheless, the results show that closed-source proprietary models, such as GPT-4o and Claude 3.5 Sonnet, consistently outperform smaller open-source models. Interestingly, the performance gap between these models is narrower on \cs{} datasets than on \ca{} datasets.

\begin{figure*}[h!]
    \centering
    \includegraphics[width=0.85\linewidth]{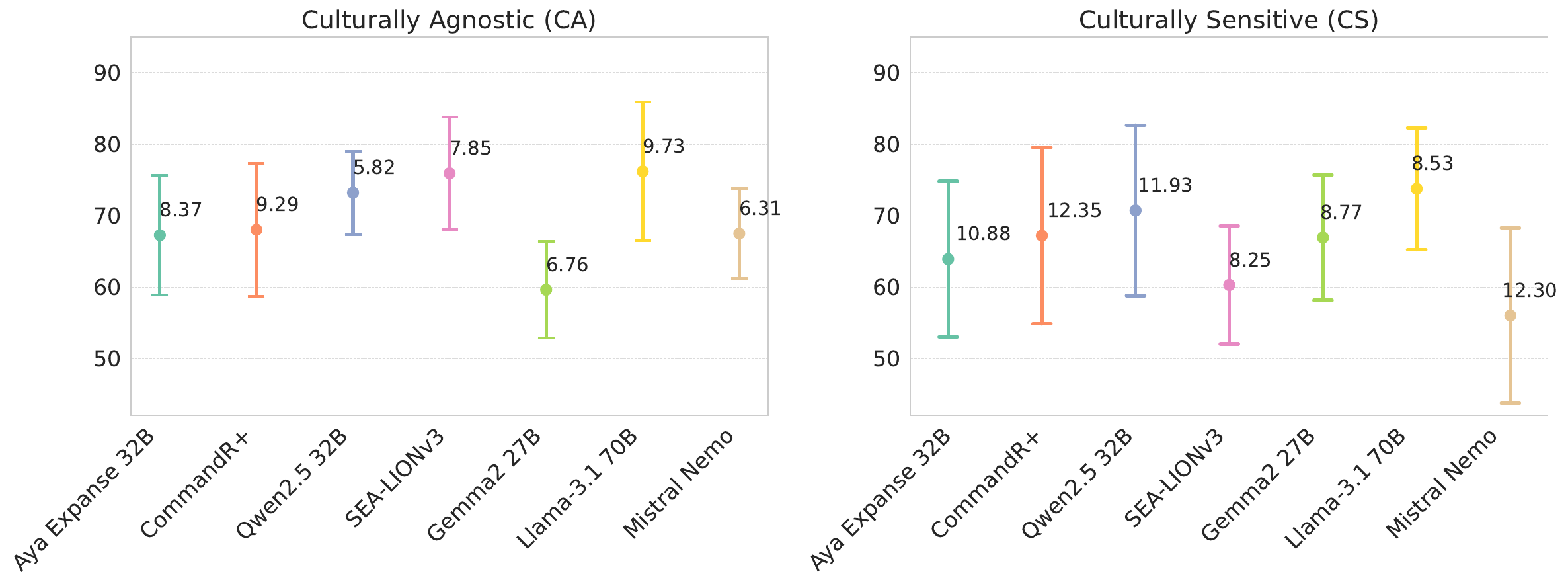}
    \caption{Model evaluations on \ca{} and \cs{} samples in \gmmlulite. Error bars indicate standard deviation across languages.}
    \label{fig:model_evaluations_on_mmlu_lite}
\vspace*{-3mm}
\end{figure*}

Additionally, we assess mid-size and large open-weight models on \gmmlulite, a fully human-translated (or post-edited) subset evenly balanced between \cs{} and \ca{} samples. Unlike the full \gmmlu, this balance enables clearer comparisons. Figure~\ref{fig:model_evaluations_on_mmlu_lite} shows that overall, models perform better on the \ca{} portion.

\textbf{Performance on \cs is higher but presents more variance} Another key observation is that the average accuracy across all models is higher on \cs{} datasets compared to \ca{} datasets. This trend can be attributed to the nature of the \cs{} samples, which are predominantly drawn from Social Sciences and Humanities domains where models generally perform better. In contrast, \ca{} datasets include more challenging categories, such as Medical and STEM, as illustrated in Figure~\ref{fig:subject_level}.

However, the standard deviation in performance across languages is higher for \cs{} data than for \ca{} data for all models. This can be attributed to several factors: culturally sensitive tasks are inherently more challenging and require deeper contextual understanding, making them more susceptible to variations in translation quality. Nuanced cultural, regional, or dialectal references in \cs{} tasks often amplify this sensitivity, as differences in how these references are translated can affect model performance. Furthermore, many large language models are trained predominantly on data from high-resource or Western cultures, leading to biases that favor these contexts and cause inconsistencies when applied to less-represented cultures.

On \gmmlulite, the pattern shifts: \cs{} tasks have lower average accuracies and greater variance than \ca{} tasks. This highlights how cultural specificity increases performance instability, when the \cs{} and \ca{} samples are balanced.

\textbf{Evaluations Across High-, Mid-, and Low-Resource Languages.}
To analyze model performance across languages with varying resource availability, we evaluated the models on \ca{} and \cs{} subsets, categorized into \highemoji{} high-, \midemoji{}mid-, and \lowemoji{} low-resource languages. This evaluation provides insights into how models handle linguistic diversity and cultural nuances across different resource levels.

\begin{figure}[ht!]
    \centering
    \begin{subfigure}[b]{1.0\textwidth}
    \centering
    {\includegraphics[width=\linewidth]{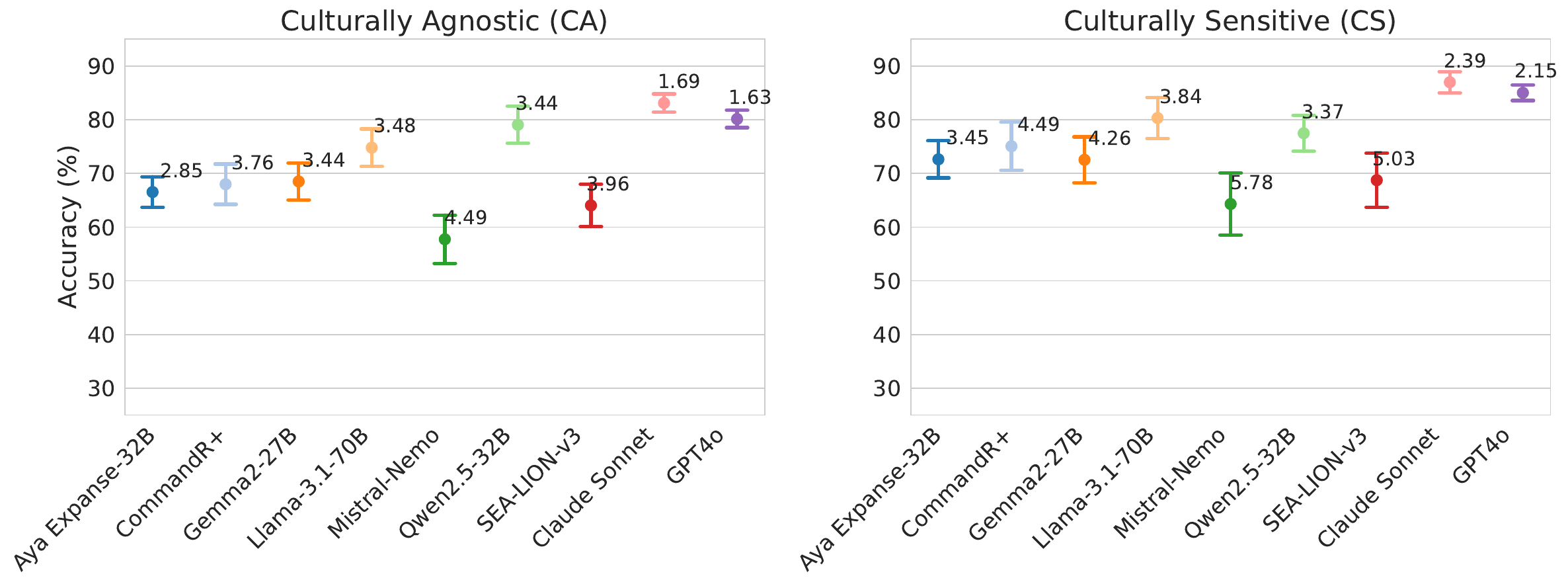}}
    \end{subfigure}
    \vspace{2mm}
    \begin{subfigure}[b]{1.0\textwidth}
    \centering
    {\includegraphics[width=\linewidth]{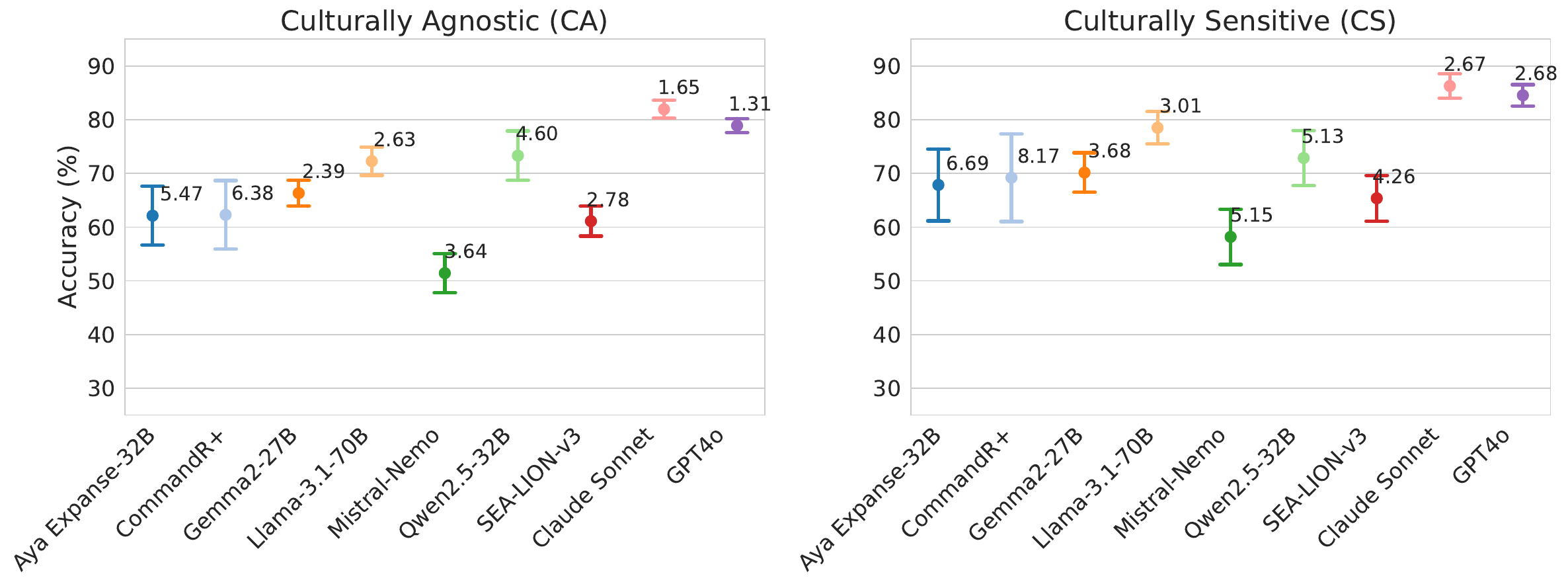}}
    \end{subfigure}
    \vspace{2mm}
    \begin{subfigure}[b]{1.0\textwidth}
    \centering
    {\includegraphics[width=\linewidth]{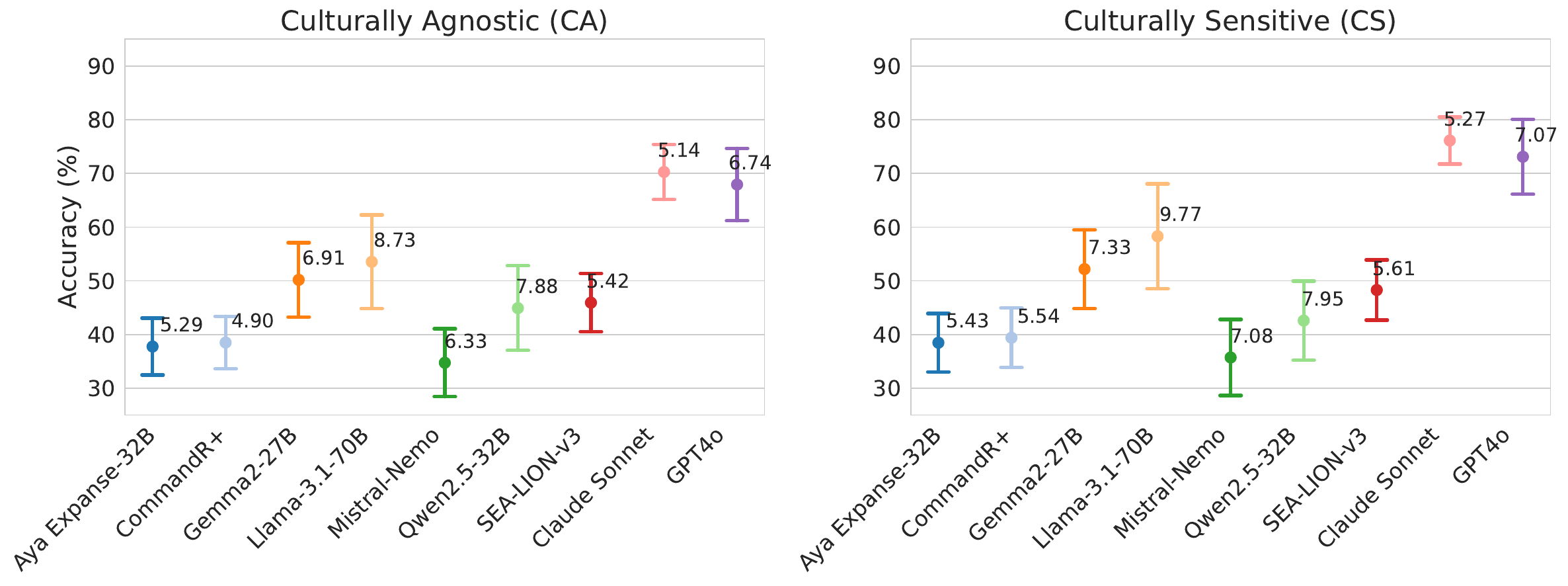}}
    \end{subfigure}
    \caption{Model evaluations on (Top) \highemoji{}high-resource , (Mid) \midemoji {mid-resource} and (Bottom) \lowemoji{low resource} data samples for \ca{} and \cs{} subsets.}
    \label{fig:model_evaluations_on_ca_cs}
\end{figure}

\textbf{Performance degrades on low-resource languages with higher variability} For both \ca{} and \cs{} datasets, \highemoji{}high-resource  languages consistently achieve the highest average accuracy across all models. As expected, performance declines significantly for \lowemoji{low-resource} languages due to the limited availability of high-quality training data, which hinders model generalization. This decline is accompanied by an increase in performance variability, with the standard deviation rising for \midemoji {mid-resource} languages and even more so for \lowemoji{low-resource} languages, particularly on \cs{} datasets.

The average standard deviation for \highemoji{}high-resource  languages is \textbf{3.21} on \ca{} datasets and \textbf{3.86} on \cs{} datasets. For \midemoji mid-resource languages, these values increase to \textbf{3.42} and \textbf{4.6}, respectively.
\lowemoji{Low-resource} languages exhibit significantly higher standard deviations, with averages rising to \textbf{6.37} on \ca{} datasets and \textbf{6.78} on \cs{} datasets. These represent increases of 98\% and 75\% compared to high-resource languages, highlighting the greater variability and sensitivity in low-resource settings.
This increased variability in model performances highlights the challenges of culturally sensitive tasks, which demand a nuanced understanding of regional or dialectal references. Across all level of resourcefulness, performance on \cs{} shows higher variability than \ca.

\textbf{Model Rank Changes.}
This section explores how model performance rankings differ between \ca{} and \cs{} datasets, calculated relative to their ranks on MA \memoemoji, across multiple languages. Table~\ref{tab:rank_changes_human} highlights rank changes for \textbf{human-translated} languages, organized by resource level: \highemoji{}high-resource, \midemoji mid-resource, and \lowemoji low-resource. These rankings offer valuable insights into how dataset type, resource availability and model size impact model performances. Comprehensive rankings for all languages are available in Table~\ref{tab:model_rank_changes_high_res} and Table~\ref{tab:model_rank_changes_mid_low_res} in Appendix~\ref{app:rank_changes}.

\begin{table}[ht!]
\centering
\scalebox{0.65}{
\begin{tabular}{l|c|cccccccccccccc}
\toprule
\textbf{Language} & \textbf{Dataset} & \rotatebox{90}{Aya Exp. 8B} & \rotatebox{90}{Aya Exp. 32B} & \rotatebox{90}{CommandR} & \rotatebox{90}{CommandR+} & \rotatebox{90}{Gemma2 9B} & \rotatebox{90}{Gemma2 27B} & \rotatebox{90}{Llama-3.1 8B} & \rotatebox{90}{Llama-3.1 70B} & \rotatebox{90}{Mistral Nemo} & \rotatebox{90}{Qwen2.5 7B} & \rotatebox{90}{Qwen2.5 32B} & \rotatebox{90}{SEA-LION-v3} & \rotatebox{90}{GPT4o} & \rotatebox{90}{Claude Sonnet} \\
\midrule
\multirow{2}{*}{\textcolor{blue!50}{\highemoji} {Arabic}} &
\scalesemoji &
- & 
- & 
- & 
- & 
- & 
- & 
- & 
- & 
- & 
\tcbox[colback=LightGreen]{\textcolor{black}{$\uparrow$1}} & 
- & 
\tcbox[colback=LightRed]{\textcolor{black}{$\downarrow$1}} & 
- & 
- \\ & 
\statuelibertyemoji & 
- & 
\tcbox[colback=LightGreen]{\textcolor{black}{$\uparrow$1}} & 
- & 
- & 
- & 
\tcbox[colback=LightRed]{\textcolor{black}{$\downarrow$1}} & 
- & 
\tcbox[colback=LightGreen]{\textcolor{black}{$\uparrow$1}} & 
- & 
- & 
\tcbox[colback=LightRed]{\textcolor{black}{$\downarrow$1}} & 
- & 
- & 
- \\ \midrule
\multirow{2}{*}{\highemoji{}{Chinese}} &
\scalesemoji &
- & 
- & 
\tcbox[colback=LightRed]{\textcolor{black}{$\downarrow$1}} & 
- & 
\tcbox[colback=LightGreen]{\textcolor{black}{$\uparrow$1}} & 
- & 
- & 
- & 
- & 
- & 
\tcbox[colback=LightGreen]{\textcolor{black}{$\uparrow$1}} & 
- & 
\tcbox[colback=LightRed]{\textcolor{black}{$\downarrow$1}} & 
- \\ & 
\statuelibertyemoji & 
\tcbox[colback=LightGreen]{\textcolor{black}{$\uparrow$1}} & 
\tcbox[colback=LightGreen]{\textcolor{black}{$\uparrow$1}} & 
\tcbox[colback=LightGreen]{\textcolor{black}{$\uparrow$1}} & 
\tcbox[colback=LightGreen]{\textcolor{black}{$\uparrow$2}} & 
\tcbox[colback=LightGreen]{\textcolor{black}{$\uparrow$1}} & 
- & 
\tcbox[colback=LightRed]{\textcolor{black}{$\downarrow$1}} & 
\tcbox[colback=LightGreen]{\textcolor{black}{$\uparrow$1}} & 
- & 
\tcbox[colback=LightRed]{\textcolor{black}{$\downarrow$3}} & 
\tcbox[colback=LightRed]{\textcolor{black}{$\downarrow$1}} & 
\tcbox[colback=LightRed]{\textcolor{black}{$\downarrow$2}} & 
\tcbox[colback=LightGreen]{\textcolor{black}{$\uparrow$1}} & 
\tcbox[colback=LightRed]{\textcolor{black}{$\downarrow$1}} \\ \midrule
\multirow{2}{*}{\highemoji{}{English}} &
\scalesemoji & 
- & 
- & 
- & 
- & 
- & 
\tcbox[colback=LightRed]{\textcolor{black}{$\downarrow$1}} & 
- & 
- & 
- & 
\tcbox[colback=LightGreen]{\textcolor{black}{$\uparrow$1}} & 
\tcbox[colback=LightGreen]{\textcolor{black}{$\uparrow$1}} & 
- & 
\tcbox[colback=LightRed]{\textcolor{black}{$\downarrow$1}} & 
- \\ & 
\statuelibertyemoji &
- & 
\tcbox[colback=LightGreen]{\textcolor{black}{$\uparrow$1}} & 
- & 
- & 
- & 
- & 
- & 
\tcbox[colback=LightGreen]{\textcolor{black}{$\uparrow$1}} & 
- & 
\tcbox[colback=LightRed]{\textcolor{black}{$\downarrow$1}} & 
\tcbox[colback=LightRed]{\textcolor{black}{$\downarrow$1}} & 
- & 
- & 
- \\ \midrule
\multirow{2}{*}{\highemoji{}{French}} &
\scalesemoji & 
- & 
\tcbox[colback=LightGreen]{\textcolor{black}{$\uparrow$1}} & 
- & 
- & 
- & 
- & 
- & 
- & 
- & 
\tcbox[colback=LightRed]{\textcolor{black}{$\downarrow$1}} & 
- & 
- & 
- & 
- \\ & 
\statuelibertyemoji & 
- & 
\tcbox[colback=LightGreen]{\textcolor{black}{$\uparrow$2}} & 
\tcbox[colback=LightGreen]{\textcolor{black}{$\uparrow$2}} & 
\tcbox[colback=LightGreen]{\textcolor{black}{$\uparrow$1}} & 
- & 
\tcbox[colback=LightRed]{\textcolor{black}{$\downarrow$2}} & 
- & 
\tcbox[colback=LightGreen]{\textcolor{black}{$\uparrow$1}} & 
- & 
\tcbox[colback=LightRed]{\textcolor{black}{$\downarrow$3}} & 
\tcbox[colback=LightRed]{\textcolor{black}{$\downarrow$1}} & 
\tcbox[colback=LightGreen]{\textcolor{black}{$\uparrow$1}} & 
- & 
- \\ \midrule
\multirow{2}{*}{\highemoji{}{German}} & 
\scalesemoji &
- & 
\tcbox[colback=LightRed]{\textcolor{black}{$\downarrow$1}} & 
- & 
\tcbox[colback=LightRed]{\textcolor{black}{$\downarrow$1}} & 
- & 
\tcbox[colback=LightGreen]{\textcolor{black}{$\uparrow$1}} & 
- & 
- & 
- & 
\tcbox[colback=LightGreen]{\textcolor{black}{$\uparrow$1}} & 
- & 
- & 
- & 
- \\ & 
\statuelibertyemoji &
- & 
- & 
\tcbox[colback=LightRed]{\textcolor{black}{$\downarrow$1}} & 
- & 
\tcbox[colback=LightGreen]{\textcolor{black}{$\uparrow$2}} & 
- & 
- & 
\tcbox[colback=LightGreen]{\textcolor{black}{$\uparrow$1}} & 
- & 
\tcbox[colback=LightRed]{\textcolor{black}{$\downarrow$3}} & 
\tcbox[colback=LightRed]{\textcolor{black}{$\downarrow$1}} & 
\tcbox[colback=LightGreen]{\textcolor{black}{$\uparrow$2}} & 
- & 
- \\ \midrule
\multirow{2}{*}{\highemoji{}{Hindi}} &
\scalesemoji & - & 
\tcbox[colback=LightGreen]{\textcolor{black}{$\uparrow$1}} & 
\tcbox[colback=LightRed]{\textcolor{black}{$\downarrow$2}} & 
\tcbox[colback=LightRed]{\textcolor{black}{$\downarrow$1}} & 
\tcbox[colback=LightGreen]{\textcolor{black}{$\uparrow$1}} & 
- & 
- & 
- & 
- & 
- & 
- & 
\tcbox[colback=LightGreen]{\textcolor{black}{$\uparrow$1}} & 
- & 
- \\ & 
\statuelibertyemoji & 
\tcbox[colback=LightGreen]{\textcolor{black}{$\uparrow$1}} & 
\tcbox[colback=LightRed]{\textcolor{black}{$\downarrow$1}} & 
\tcbox[colback=LightGreen]{\textcolor{black}{$\uparrow$1}} & 
\tcbox[colback=LightGreen]{\textcolor{black}{$\uparrow$2}} & 
- & 
\tcbox[colback=LightRed]{\textcolor{black}{$\downarrow$1}} & 
\tcbox[colback=LightGreen]{\textcolor{black}{$\uparrow$1}} & 
- & 
\tcbox[colback=LightGreen]{\textcolor{black}{$\uparrow$1}} & 
\tcbox[colback=LightRed]{\textcolor{black}{$\downarrow$3}} & 
\tcbox[colback=LightRed]{\textcolor{black}{$\downarrow$1}} & 
- & 
\tcbox[colback=LightGreen]{\textcolor{black}{$\uparrow$1}} & 
\tcbox[colback=LightRed]{\textcolor{black}{$\downarrow$1}} \\ \midrule
\multirow{2}{*}{\highemoji{}{Italian}} &
\scalesemoji & 
- & 
- & 
- & 
- & 
- & 
- & 
- & 
- & 
- & 
- & 
- & 
- & 
- & 
- \\ & 
\statuelibertyemoji & 
- & 
- & 
\tcbox[colback=LightGreen]{\textcolor{black}{$\uparrow$1}} & 
\tcbox[colback=LightGreen]{\textcolor{black}{$\uparrow$1}} & 
- & 
\tcbox[colback=LightRed]{\textcolor{black}{$\downarrow$1}} & 
- & 
\tcbox[colback=LightGreen]{\textcolor{black}{$\uparrow$1}} & 
- & 
\tcbox[colback=LightRed]{\textcolor{black}{$\downarrow$2}} & 
\tcbox[colback=LightRed]{\textcolor{black}{$\downarrow$1}} & 
\tcbox[colback=LightGreen]{\textcolor{black}{$\uparrow$1}} & 
- & 
- \\ \midrule
\multirow{2}{*}{\highemoji{}{Japanese}} &
\scalesemoji &
- & 
- & 
- & 
- & 
- & 
- & 
- & 
- & 
- & 
- & 
- & 
- & 
- & 
- \\ & 
\statuelibertyemoji & 
- & 
\tcbox[colback=LightGreen]{\textcolor{black}{$\uparrow$1}} & 
\tcbox[colback=LightGreen]{\textcolor{black}{$\uparrow$1}} & 
\tcbox[colback=LightGreen]{\textcolor{black}{$\uparrow$1}} & 
\tcbox[colback=LightGreen]{\textcolor{black}{$\uparrow$1}} & 
\tcbox[colback=LightRed]{\textcolor{black}{$\downarrow$2}} & 
- & 
\tcbox[colback=LightGreen]{\textcolor{black}{$\uparrow$1}} & 
- & 
\tcbox[colback=LightRed]{\textcolor{black}{$\downarrow$1}} & 
\tcbox[colback=LightRed]{\textcolor{black}{$\downarrow$1}} & 
\tcbox[colback=LightRed]{\textcolor{black}{$\downarrow$1}} & 
- & 
- \\ \midrule
\multirow{2}{*}{\highemoji{}{Portuguese}} &
\scalesemoji & 
- & 
- & 
- & 
- & 
- & 
- & 
- & 
- & 
- & 
- & 
- & 
- & 
- & 
- \\ & 
\statuelibertyemoji &  
- & 
\tcbox[colback=LightGreen]{\textcolor{black}{$\uparrow$1}} & 
\tcbox[colback=LightGreen]{\textcolor{black}{$\uparrow$1}} & 
\tcbox[colback=LightGreen]{\textcolor{black}{$\uparrow$1}} & 
\tcbox[colback=LightGreen]{\textcolor{black}{$\uparrow$1}} & 
\tcbox[colback=LightRed]{\textcolor{black}{$\downarrow$1}} & 
- & 
\tcbox[colback=LightGreen]{\textcolor{black}{$\uparrow$1}} & 
- & 
\tcbox[colback=LightRed]{\textcolor{black}{$\downarrow$2}} & 
\tcbox[colback=LightRed]{\textcolor{black}{$\downarrow$1}} & 
\tcbox[colback=LightRed]{\textcolor{black}{$\downarrow$1}} & 
- & 
- \\ \midrule
\multirow{2}{*}{\highemoji{}{Spanish}} &
\scalesemoji & 
- & 
\tcbox[colback=LightRed]{\textcolor{black}{$\downarrow$1}} & 
- & 
\tcbox[colback=LightRed]{\textcolor{black}{$\downarrow$1}} & 
- & 
\tcbox[colback=LightGreen]{\textcolor{black}{$\uparrow$1}} & 
- & 
- & 
- & 
\tcbox[colback=LightGreen]{\textcolor{black}{$\uparrow$1}} & 
- & 
- & 
- & 
- \\ & 
\statuelibertyemoji & 
- & 
- & 
\tcbox[colback=LightGreen]{\textcolor{black}{$\uparrow$1}} & 
- & 
\tcbox[colback=LightGreen]{\textcolor{black}{$\uparrow$2}} & 
- & 
- & 
\tcbox[colback=LightGreen]{\textcolor{black}{$\uparrow$1}} & 
- & 
\tcbox[colback=LightRed]{\textcolor{black}{$\downarrow$3}} & 
\tcbox[colback=LightRed]{\textcolor{black}{$\downarrow$1}} & 
- & 
- & 
- \\ \midrule \midrule
\multirow{2}{*}{\midemoji  {Bengali}}
& \scalesemoji & - & \tcbox[colback=LightGreen]{\textcolor{black}{$\uparrow$1}} & - & - & - & - & - & \tcbox[colback=LightRed]{\textcolor{black}{$\downarrow$1}} & \tcbox[colback=LightRed]{\textcolor{black}{$\downarrow$1}} & - & - & - & - & - \\ 
& \statuelibertyemoji & - & - & - & - & - & - & - & - & \tcbox[colback=LightGreen]{\textcolor{black}{$\uparrow$1}} & \tcbox[colback=LightRed]{\textcolor{black}{$\downarrow$1}} & - & - & - & - \\ 
\midrule
\multirow{2}{*}{\midemoji  {Indonesian}} &
\scalesemoji  & - &  - & 
\tcbox[colback=LightRed]{\textcolor{black}{$\downarrow$1}} & 
\tcbox[colback=LightRed]{\textcolor{black}{$\downarrow$1}} & 
\tcbox[colback=LightRed]{\textcolor{black}{$\downarrow$1}} & 
\tcbox[colback=LightGreen]{\textcolor{black}{$\uparrow$1}} & 
- &  - &  - & 
\tcbox[colback=LightGreen]{\textcolor{black}{$\uparrow$2}} &  - &  - &  - &  - \\ & 
\statuelibertyemoji & 
- & 
- & 
\tcbox[colback=LightGreen]{\textcolor{black}{$\uparrow$1}} & 
- & 
- & 
- & 
\tcbox[colback=LightRed]{\textcolor{black}{$\downarrow$1}} & 
\tcbox[colback=LightGreen]{\textcolor{black}{$\uparrow$1}} & 
\tcbox[colback=LightGreen]{\textcolor{black}{$\uparrow$1}} & 
- & 
\tcbox[colback=LightRed]{\textcolor{black}{$\downarrow$1}} & 
\tcbox[colback=LightRed]{\textcolor{black}{$\downarrow$1}} & 
- & 
- \\ \midrule
\multirow{2}{*}{\midemoji  {Korean}} &
\scalesemoji &
\tcbox[colback=LightRed]{\textcolor{black}{$\downarrow$1}} & 
\tcbox[colback=LightRed]{\textcolor{black}{$\downarrow$1}} & 
\tcbox[colback=LightRed]{\textcolor{black}{$\downarrow$1}} & 
- & 
- & 
\tcbox[colback=LightGreen]{\textcolor{black}{$\uparrow$1}} & 
\tcbox[colback=LightGreen]{\textcolor{black}{$\uparrow$1}} & 
- & 
- & 
\tcbox[colback=LightGreen]{\textcolor{black}{$\uparrow$1}} & 
- & 
- & 
- & 
- \\ & 
\statuelibertyemoji & 
- & 
\tcbox[colback=LightGreen]{\textcolor{black}{$\uparrow$1}} & 
\tcbox[colback=LightGreen]{\textcolor{black}{$\uparrow$1}} & 
\tcbox[colback=LightRed]{\textcolor{black}{$\downarrow$1}} & 
- & 
\tcbox[colback=LightRed]{\textcolor{black}{$\downarrow$1}} & 
- & 
\tcbox[colback=LightGreen]{\textcolor{black}{$\uparrow$1}} & 
- & 
- & 
\tcbox[colback=LightRed]{\textcolor{black}{$\downarrow$1}} & 
- & 
- & 
- \\ \midrule \midrule
\multirow{2}{*}{\lowemoji {Sinhala}} &
\scalesemoji &
- & 
\tcbox[colback=LightGreen]{\textcolor{black}{$\uparrow$1}} & 
- & 
- & 
- & 
- & 
\tcbox[colback=LightRed]{\textcolor{black}{$\downarrow$3}} & 
- & 
- & 
\tcbox[colback=LightGreen]{\textcolor{black}{$\uparrow$2}} & 
- & 
- & 
- & 
- \\ & 
\statuelibertyemoji & 
- & 
\tcbox[colback=LightRed]{\textcolor{black}{$\downarrow$1}} & 
\tcbox[colback=LightGreen]{\textcolor{black}{$\uparrow$1}} & 
\tcbox[colback=LightGreen]{\textcolor{black}{$\uparrow$1}} & 
- & 
- & 
- & 
- & 
- & 
\tcbox[colback=LightRed]{\textcolor{black}{$\downarrow$1}} & 
- & 
- & 
- & 
- \\ 
\midrule
\multirow{2}{*}{\lowemoji {Swahili}} &
\scalesemoji &
- & 
\tcbox[colback=LightRed]{\textcolor{black}{$\downarrow$1}} & 
- & 
- & 
- & 
- & 
\tcbox[colback=LightGreen]{\textcolor{black}{$\uparrow$1}} & 
- & 
- & 
- & 
- & 
- & 
- & 
- \\ & 
\statuelibertyemoji & 
- & 
- & 
\tcbox[colback=LightGreen]{\textcolor{black}{$\uparrow$1}} & 
- & 
- & 
- & 
\tcbox[colback=LightRed]{\textcolor{black}{$\downarrow$1}} & 
- & 
- & 
- & 
- & 
- & 
\tcbox[colback=LightRed]{\textcolor{black}{$\downarrow$1}} & 
\tcbox[colback=LightGreen]{\textcolor{black}{$\uparrow$1}} \\ \midrule
\multirow{2}{*}{\lowemoji {Yoruba}} &
\scalesemoji &
- & 
\tcbox[colback=LightGreen]{\textcolor{black}{$\uparrow$1}} & 
\tcbox[colback=LightRed]{\textcolor{black}{$\downarrow$2}} & 
- & 
\tcbox[colback=LightRed]{\textcolor{black}{$\downarrow$1}} & 
- & 
- & 
- & 
- & 
\tcbox[colback=LightGreen]{\textcolor{black}{$\uparrow$2}} & 
\tcbox[colback=LightGreen]{\textcolor{black}{$\uparrow$1}} & 
\tcbox[colback=LightRed]{\textcolor{black}{$\downarrow$1}} & 
- & 
- \\ & 
\statuelibertyemoji & 
- & 
\tcbox[colback=LightRed]{\textcolor{black}{$\downarrow$1}} & 
\tcbox[colback=LightGreen]{\textcolor{black}{$\uparrow$1}} & 
\tcbox[colback=LightGreen]{\textcolor{black}{$\uparrow$1}} & 
\tcbox[colback=LightGreen]{\textcolor{black}{$\uparrow$1}} & 
- & 
- & 
- & 
- & 
- & 
\tcbox[colback=LightRed]{\textcolor{black}{$\downarrow$2}} & 
- & 
- & 
- \\ 
\bottomrule
\end{tabular}}
\caption{Changes in model rankings on \ca{} and \cs{} datasets, based on MA \memoemoji, across \textbf{human-translated} languages, including English. Languages are categorized as \highemoji{high-}, \midemoji {mid-}, and \lowemoji{low}-resource. Color-coded boxes indicate increases (\tcbox[colback=LightGreen]{\textcolor{black}{$\uparrow$}}) and decreases (\tcbox[colback=LightRed]{\textcolor{black}{$\downarrow$}}) in rank.}
\label{tab:rank_changes_human}
\end{table}

The rank changes reveal three key findings:

\textbf{1) Models perform differently across \ca{} and \cs{} datasets, with the latter showing greater variation.} Rankings on \ca{} datasets exhibit minimal changes. For instance, Italian, Japanese, and Portuguese show no rank changes, while Arabic and French each experience only two shifts, each by one position.

On the other hand, model performance varies significantly on \cs{} datasets. Chinese and Hindi emerge as the most sensitive languages to culture-specific knowledge, with models showing both increases and decreases in rankings. Similar variations are evident in French, German, Italian, Japanese, and Portuguese. Notably, models from the Aya Expanse and CommandR families tend to show positive trends on \cs{} datasets, particularly for these languages.
On average, across all languages, \ca{} datasets see 3.4 rank changes and 3.7 position changes, whereas \cs{} datasets experience markedly higher volatility, with 5.7 rank changes and 7.3 position changes.

\textbf{2) The difference between performances on \ca{} and \cs{} datasets are less on low-resource languages.}
\highemoji{High-resource} languages demonstrate relatively stable rankings on \ca{} datasets, with an average of 3.3 rank changes and a maximum shift of 3 positions. However, on \cs{} datasets, ranking changes are more pronounced, with an average of 6.8 rank changes and 9.1 position shifts. In contrast, \midemoji {mid-resource} languages display moderate variability. 
While \emph{small models} face slightly greater fluctuations on \cs{} datasets, their performance on \ca{} datasets remains more consistent. 
For \midemoji {mid-resource} languages, the average rank changes are 3.7 on \ca{} and 4.7 on \cs{}, with corresponding position changes of 4.7 and 4.9. Among the three resource groups, \midemoji {mid-resource} languages show the smallest difference between \ca{} and \cs{} performance.

\lowemoji{Low-resource} languages show an increase in the difference between \ca{} and \cs{} rank changes compared to \midemoji {mid-resource}. Average rank changes are 3.3 on \ca{} datasets and 3.7 on \cs{}, with position changes rising to 5.7 on \ca{} and 7.9 on \cs{}. Notably, this group also experiences the largest rank changes. Table~\ref{tab:rank_changes_low_2} highlights the most significant changes across all languages, including \textbf{rank shifts of up to 5 positions} for Malagasy, and \textbf{13 ranking changes} for the models on Ukrainian. These findings underscore how resource levels amplify rank changes, even within \ca{} datasets.

\textbf{3) Model size influences performance variations.} We analyzed performance variations across three model groups, as defined in the \emph{Model} section (excluding closed-weight models due to unknown sizes). Our findings highlight distinct trends for large, mid-size, and small models:

\emph{Large models} demonstrate higher consistency across datasets and resource levels. The average rank changes for large models are minimal, at 0.21 for \ca{} and 0.67 for \cs{}. The maximum position shift for models in this group is 3 while it can be 5 for \emph{small-models}. This consistency reflects their robustness and higher capacity to generalize across diverse datasets.

\emph{Mid-size models}, on the other hand, show much bigger variability. Their average rank changes are 0.33 for \ca{} and 1.97 for \cs{}, indicating they are more sensitive to dataset characteristics, particularly in the \cs{} datasets that requires cultural knowledge.

\emph{Small models} exhibit the smallest difference in rank change between \ca{} and \cs{} (0.35 and 0.45, respectively). However, this apparent stability stems from their weaker overall performance across both datasets. For instance, the average accuracy for small models is 51.3\% on \ca{} and 54.8\% on \cs{}, while mid-size models achieve 59.1\% and 61.7\%, and large models perform at 61.6\% and 66.8\% on \ca{} and \cs{}, respectively.

\begin{table}[ht!]
\centering
\scalebox{0.65}{
\begin{tabular}{l|c|cccccccccccccc}
\toprule
\textbf{Language} & \textbf{Dataset} & \rotatebox{90}{Aya Exp. 8B} & \rotatebox{90}{Aya Exp. 32B} & \rotatebox{90}{CommandR} & \rotatebox{90}{CommandR+} & \rotatebox{90}{Gemma2 9B} & \rotatebox{90}{Gemma2 27B} & \rotatebox{90}{Llama-3.1 8B} & \rotatebox{90}{Llama-3.1 70B} & \rotatebox{90}{Mistral Nemo} & \rotatebox{90}{Qwen2.5 7B} & \rotatebox{90}{Qwen2.5 32B} & \rotatebox{90}{SEA-LION-v3} & \rotatebox{90}{GPT4o} & \rotatebox{90}{Claude Sonnet} \\
\midrule
\multirow{2}{*}{\midemoji  Greek} &
\scalesemoji &
\tcbox[colback=LightRed]{\textcolor{black}{$\downarrow$1}} & 
\tcbox[colback=LightRed]{\textcolor{black}{$\downarrow$1}} & 
- & 
- & 
- & 
\tcbox[colback=LightGreen]{\textcolor{black}{$\uparrow$1}} & 
- & 
- & 
\tcbox[colback=LightRed]{\textcolor{black}{$\downarrow$1}} & 
\tcbox[colback=LightGreen]{\textcolor{black}{$\uparrow$2}} & 
- & 
- & 
- & 
- \\ 
& \statuelibertyemoji & 
- & 
- & 
\tcbox[colback=LightGreen]{\textcolor{black}{$\uparrow$2}} & 
\tcbox[colback=LightGreen]{\textcolor{black}{$\uparrow$3}} & 
- & 
\tcbox[colback=LightRed]{\textcolor{black}{$\downarrow$1}} & 
\tcbox[colback=LightGreen]{\textcolor{black}{$\uparrow$1}} & 
- & 
- & 
\tcbox[colback=LightRed]{\textcolor{black}{$\downarrow$1}} & 
\tcbox[colback=LightRed]{\textcolor{black}{$\downarrow$4}} & 
- & 
- & 
- \\ 
\midrule
\multirow{2}{*}{\midemoji  Ukrainian} &
\scalesemoji &
- & 
\tcbox[colback=LightGreen]{\textcolor{black}{$\uparrow$1}} & 
- & 
\tcbox[colback=LightRed]{\textcolor{black}{$\downarrow$1}} & 
\tcbox[colback=LightRed]{\textcolor{black}{$\downarrow$1}} & 
- & 
- & 
- & 
- & 
\tcbox[colback=LightGreen]{\textcolor{black}{$\uparrow$1}} & 
- & 
- & 
- & 
- \\ 
& \statuelibertyemoji & 
- & 
\tcbox[colback=LightGreen]{\textcolor{black}{$\uparrow$1}} & 
- & 
\tcbox[colback=LightGreen]{\textcolor{black}{$\uparrow$1}} & 
- & 
\tcbox[colback=LightRed]{\textcolor{black}{$\downarrow$2}} & 
- & 
\tcbox[colback=LightGreen]{\textcolor{black}{$\uparrow$1}} & 
\tcbox[colback=LightGreen]{\textcolor{black}{$\uparrow$1}} & 
\tcbox[colback=LightRed]{\textcolor{black}{$\downarrow$1}} & 
\tcbox[colback=LightRed]{\textcolor{black}{$\downarrow$1}} & 
- & 
\tcbox[colback=LightGreen]{\textcolor{black}{$\uparrow$1}} & 
\tcbox[colback=LightRed]{\textcolor{black}{$\downarrow$1}} \\ 
\midrule
\multirow{2}{*}{\lowemoji Malagasy} &
\scalesemoji &
- & 
\tcbox[colback=LightRed]{\textcolor{black}{$\downarrow$1}} & 
- & 
- & 
- & 
- & 
- & 
- & 
- & 
\tcbox[colback=LightGreen]{\textcolor{black}{$\uparrow$1}} & 
- & 
- & 
- & 
- \\ 
& \statuelibertyemoji & 
- & 
\tcbox[colback=LightGreen]{\textcolor{black}{$\uparrow$1}} & 
\tcbox[colback=LightGreen]{\textcolor{black}{$\uparrow$4}} & 
\tcbox[colback=LightGreen]{\textcolor{black}{$\uparrow$1}} & 
- & 
- & 
\tcbox[colback=LightRed]{\textcolor{black}{$\downarrow$1}} & 
- & 
\tcbox[colback=LightGreen]{\textcolor{black}{$\uparrow$1}} & 
\tcbox[colback=LightRed]{\textcolor{black}{$\downarrow$1}} & 
\tcbox[colback=LightRed]{\textcolor{black}{$\downarrow$5}} & 
- & 
- & 
- \\ 
\midrule
\multirow{2}{*}{\lowemoji Shona} &
\scalesemoji &
- & 
- & 
- & 
- & 
\tcbox[colback=LightRed]{\textcolor{black}{$\downarrow$1}} & 
- & 
- & 
- & 
- & 
- & 
\tcbox[colback=LightGreen]{\textcolor{black}{$\uparrow$1}} & 
- & 
\tcbox[colback=LightRed]{\textcolor{black}{$\downarrow$1}} & 
\tcbox[colback=LightGreen]{\textcolor{black}{$\uparrow$1}} \\ 
& \statuelibertyemoji & 
\tcbox[colback=LightGreen]{\textcolor{black}{$\uparrow$2}} & 
- & 
\tcbox[colback=LightGreen]{\textcolor{black}{$\uparrow$1}} & 
\tcbox[colback=LightGreen]{\textcolor{black}{$\uparrow$1}} & 
- & 
- & 
\tcbox[colback=LightGreen]{\textcolor{black}{$\uparrow$1}} & 
- & 
- & 
\tcbox[colback=LightRed]{\textcolor{black}{$\downarrow$4}} & 
\tcbox[colback=LightRed]{\textcolor{black}{$\downarrow$1}} & 
- & 
- & 
- \\ 
\bottomrule
\end{tabular}}
\caption{Changes in model rankings on \ca{} and \cs{} datasets, based on MA \memoemoji{} on Greek, Ukrainian, Malagasy, and Shona.}
\label{tab:rank_changes_low_2}
\end{table}

Overall, we can conclude that dataset characteristics significantly impact model performance across all model sizes, though the magnitude of variability differs. Across all groups, models demonstrate sensitivity to the diverse cultural and linguistic nuances present in \cs{} datasets, with performance variations reflecting their capacity to adapt to dataset-specific nuances. This pattern holds consistently, regardless of model size, though the magnitude of variability differs.

A similar trend appears in \gmmlulite{}, where despite being smaller and balanced, performance volatility is still higher on \cs{} datasets, particularly for low-resource languages as shown in Table~\ref{tab:rank_changes_mmlu_lite}.

\begin{center}
\footnotesize
\begin{table*}[h!]
\begin{minipage}{0.48\textwidth}
\centering
\scalebox{0.65}{
\begin{tabular}{l|c|cccccccccccccc}
\toprule
\textbf{Language} & \textbf{Dataset} & \rotatebox{90}{Aya Exp. 32B} & \rotatebox{90}{CommandR+}  & \rotatebox{90}{Gemma2 27B} & \rotatebox{90}{Llama-3.1 70B} & \rotatebox{90}{Mistral Nemo}  & \rotatebox{90}{Qwen2.5 32B} & \rotatebox{90}{SEA-LION-v3} \\
\midrule
\multirow{2}{*}{\textcolor{blue!50}{\highemoji} {Arabic}} &
CA &
- & 
\tcbox[colback=LightRed]{\textcolor{black}{$\downarrow$1}} & 
\tcbox[colback=LightGreen]{\textcolor{black}{$\uparrow$1}} & 
- & 
- & 
- & 
- \\ & 
CS & 
\tcbox[colback=LightGreen]{\textcolor{black}{$\uparrow$1}} & 
- & 
\tcbox[colback=LightRed]{\textcolor{black}{$\downarrow$1}} & 
- & 
- & 
- & 
- \\ \midrule
\multirow{2}{*}{\textcolor{blue!50}{\highemoji} {Chinese}} &
CA &

\tcbox[colback=LightGreen]{\textcolor{black}{$\uparrow$1}} & 
\tcbox[colback=LightRed]{\textcolor{black}{$\downarrow$1}} & 
- & 
- & 
- & 
- & 
- \\ & 
CS & 
- & 
\tcbox[colback=LightGreen]{\textcolor{black}{$\uparrow$1}} & 
\tcbox[colback=LightRed]{\textcolor{black}{$\downarrow$1}} & 
- & 
- & 
- & 
- \\ \midrule
\multirow{2}{*}{\textcolor{blue!50}{\highemoji} {English}} &
CA &

\tcbox[colback=LightRed]{\textcolor{black}{$\downarrow$1}} & 
\tcbox[colback=LightRed]{\textcolor{black}{$\downarrow$1}} & 
\tcbox[colback=LightGreen]{\textcolor{black}{$\uparrow$1}} & 
\tcbox[colback=LightRed]{\textcolor{black}{$\downarrow$1}} & 
- & 
\tcbox[colback=LightGreen]{\textcolor{black}{$\uparrow$1}} & 
\tcbox[colback=LightGreen]{\textcolor{black}{$\uparrow$1}} \\ & 
CS & 
\tcbox[colback=LightGreen]{\textcolor{black}{$\uparrow$1}} & 
- & 
\tcbox[colback=LightRed]{\textcolor{black}{$\downarrow$1}} & 
- & 
\tcbox[colback=LightGreen]{\textcolor{black}{$\uparrow$1}} & 
- & 
\tcbox[colback=LightRed]{\textcolor{black}{$\downarrow$1}} \\ \midrule
\multirow{2}{*}{\textcolor{blue!50}{\highemoji} {French}} &
CA &

\tcbox[colback=LightGreen]{\textcolor{black}{$\uparrow$1}} & 
\tcbox[colback=LightRed]{\textcolor{black}{$\downarrow$1}} & 
- & 
- & 
- & 
- & 
- \\ & 
CS & 
\tcbox[colback=LightRed]{\textcolor{black}{$\downarrow$1}} & 
\tcbox[colback=LightGreen]{\textcolor{black}{$\uparrow$1}} & 
\tcbox[colback=LightRed]{\textcolor{black}{$\downarrow$1}} & 
\tcbox[colback=LightGreen]{\textcolor{black}{$\uparrow$1}} & 
\tcbox[colback=LightGreen]{\textcolor{black}{$\uparrow$2}} & 
\tcbox[colback=LightRed]{\textcolor{black}{$\downarrow$1}} & 
\tcbox[colback=LightRed]{\textcolor{black}{$\downarrow$1}} \\ \midrule
\multirow{2}{*}{\textcolor{blue!50}{\highemoji} {German}} &
CA &
- & 
\tcbox[colback=LightRed]{\textcolor{black}{$\downarrow$1}} & 
- & 
\tcbox[colback=LightRed]{\textcolor{black}{$\downarrow$1}} & 
- & 
\tcbox[colback=LightGreen]{\textcolor{black}{$\uparrow$2}} & 
- \\ & 
CS & 
- & 
\tcbox[colback=LightGreen]{\textcolor{black}{$\uparrow$1}} & 
- & 
\tcbox[colback=LightRed]{\textcolor{black}{$\downarrow$1}} & 
- & 
- & 
- \\ \midrule
\multirow{2}{*}{\textcolor{blue!50}{\highemoji} {Hindi}} &
CA &

\tcbox[colback=LightRed]{\textcolor{black}{$\downarrow$1}} & 
- & 
- & 
- & 
- & 
\tcbox[colback=LightRed]{\textcolor{black}{$\downarrow$2}} & 
\tcbox[colback=LightGreen]{\textcolor{black}{$\uparrow$3}} \\ & 
CS & 
- & 
- & 
- & 
- & 
- & 
- & 
- \\ \midrule
\multirow{2}{*}{\textcolor{blue!50}{\highemoji} {Italian}} &
CA &

\tcbox[colback=LightGreen]{\textcolor{black}{$\uparrow$2}} & 
\tcbox[colback=LightRed]{\textcolor{black}{$\downarrow$3}} & 
- & 
- & 
- & 
- & 
\tcbox[colback=LightGreen]{\textcolor{black}{$\uparrow$1}} \\ & 
CS & 
- & 
- & 
- & 
- & 
\tcbox[colback=LightGreen]{\textcolor{black}{$\uparrow$1}} & 
- & 
\tcbox[colback=LightRed]{\textcolor{black}{$\downarrow$1}} \\ \midrule
\multirow{2}{*}{\textcolor{blue!50}{\highemoji} {Japanese}}  &
CA &

\tcbox[colback=LightGreen]{\textcolor{black}{$\uparrow$1}} & 
\tcbox[colback=LightRed]{\textcolor{black}{$\downarrow$1}} & 
- & 
- & 
- & 
- & 
- \\ & 
CS & 
- & 
- & 
- & 
- & 
- & 
- & 
- \\ \bottomrule
\end{tabular}}
\end{minipage}
\hspace{5mm}
\begin{minipage}{0.48\textwidth}
\centering
\vspace{-10mm}
\scalebox{0.65}{
\begin{tabular}{l|c|cccccccccccccc}
\toprule
\textbf{Language} & \textbf{Dataset} & \rotatebox{90}{Aya Exp. 32B} & \rotatebox{90}{CommandR+}  & \rotatebox{90}{Gemma2 27B} & \rotatebox{90}{Llama-3.1 70B} & \rotatebox{90}{Mistral Nemo}  & \rotatebox{90}{Qwen2.5 32B} & \rotatebox{90}{SEA-LION-v3} \\
\midrule
\multirow{2}{*}{\textcolor{blue!50}{\highemoji} {Portuguese}}  &
CA &

\tcbox[colback=LightRed]{\textcolor{black}{$\downarrow$1}} & 
\tcbox[colback=LightRed]{\textcolor{black}{$\downarrow$2}} & 
\tcbox[colback=LightGreen]{\textcolor{black}{$\uparrow$1}} & 
\tcbox[colback=LightRed]{\textcolor{black}{$\downarrow$1}} & 
- & 
\tcbox[colback=LightGreen]{\textcolor{black}{$\uparrow$1}} & 
\tcbox[colback=LightGreen]{\textcolor{black}{$\uparrow$2}} \\ & 
CS & 
\tcbox[colback=LightGreen]{\textcolor{black}{$\uparrow$1}} & 
- & 
\tcbox[colback=LightRed]{\textcolor{black}{$\downarrow$1}} & 
- & 
- & 
- & 
- \\ \midrule
\multirow{2}{*}{\textcolor{blue!50}{\highemoji} {Spanish}}  &
CA &

- & 
- & 
- & 
- & 
- & 
- & 
- \\ & 
CS & 
- & 
- & 
- & 
\tcbox[colback=LightGreen]{\textcolor{black}{$\uparrow$1}} & 
- & 
\tcbox[colback=LightRed]{\textcolor{black}{$\downarrow$1}} & 
- \\ \midrule

\multirow{2}{*}{\textcolor{blue!50}{\midemoji} {Bengali}}  &
CA & \tcbox[colback=LightGreen]{\textcolor{black}{$\uparrow$1}} & 
- & 
- & 
- & 
\tcbox[colback=LightRed]{\textcolor{black}{$\downarrow$1}} & 
- & 
- \\ & 
CS & 
- & 
- & 
- & 
- & 
- & 
- & 
- \\ \midrule
\multirow{2}{*}{\textcolor{blue!50}{\midemoji} {Indonesian}}  &
CA &

- & 
- & 
- & 
- & 
- & 
- & 
- \\ & 
CS & 
\tcbox[colback=LightGreen]{\textcolor{black}{$\uparrow$1}} & 
\tcbox[colback=LightGreen]{\textcolor{black}{$\uparrow$1}} & 
\tcbox[colback=LightRed]{\textcolor{black}{$\downarrow$2}} & 
- & 
- & 
- & 
- \\ \midrule
\multirow{2}{*}{\textcolor{blue!50}{\midemoji} {Korean}}  &
CA &
\tcbox[colback=LightRed]{\textcolor{black}{$\downarrow$1}} & 
\tcbox[colback=LightGreen]{\textcolor{black}{$\uparrow$1}} & 
- & 
- & 
- & 
- & 
- \\ & 
CS & 
- & 
- & 
- & 
- & 
- & 
- & 
- \\ \midrule
\multirow{2}{*}{\textcolor{blue!50}{\lowemoji} {Swahili}}  &
CA &

\tcbox[colback=LightRed]{\textcolor{black}{$\downarrow$1}} & 
\tcbox[colback=LightGreen]{\textcolor{black}{$\uparrow$1}} & 
\tcbox[colback=LightGreen]{\textcolor{black}{$\uparrow$1}} & 
\tcbox[colback=LightRed]{\textcolor{black}{$\downarrow$1}} & 
\tcbox[colback=LightGreen]{\textcolor{black}{$\uparrow$1}} & 
\tcbox[colback=LightRed]{\textcolor{black}{$\downarrow$1}} & 
- \\ & 
CS & 
\tcbox[colback=LightGreen]{\textcolor{black}{$\uparrow$1}} & 
\tcbox[colback=LightRed]{\textcolor{black}{$\downarrow$1}} & 
- & 
- & 
- & 
- & 
- \\ \midrule
\multirow{2}{*}{\textcolor{blue!50}{\lowemoji} {Yoruba}}  &
CA &

- & 
\tcbox[colback=LightRed]{\textcolor{black}{$\downarrow$2}} & 
- & 
\tcbox[colback=LightRed]{\textcolor{black}{$\downarrow$2}} & 
- & 
\tcbox[colback=LightGreen]{\textcolor{black}{$\uparrow$1}} & 
\tcbox[colback=LightGreen]{\textcolor{black}{$\uparrow$3}} \\ & 
CS & 
\tcbox[colback=LightGreen]{\textcolor{black}{$\uparrow$3}} & 
\tcbox[colback=LightGreen]{\textcolor{black}{$\uparrow$1}} & 
\tcbox[colback=LightRed]{\textcolor{black}{$\downarrow$4}} & 
\tcbox[colback=LightGreen]{\textcolor{black}{$\uparrow$1}} & 
- & 
- & 
\tcbox[colback=LightRed]{\textcolor{black}{$\downarrow$1}} \\ 
\bottomrule
\end{tabular}}
\end{minipage}
\caption{Changes in model rankings on \ca{} and \cs{} datasets, based on total accuracy on \gmmlulite. Languages are categorized as \highemoji{high-}, \midemoji {mid-}, and \lowemoji{low}-resource. Color-coded boxes indicate increases (\tcbox[colback=LightGreen]{\textcolor{black}{$\uparrow$}}) and decreases (\tcbox[colback=LightRed]{\textcolor{black}{$\downarrow$}}) in rank.}
\label{tab:rank_changes_mmlu_lite}
\end{table*}
\end{center}

\begin{figure}[ht!]
    \centering
    \begin{subfigure}[b]{0.60\textwidth}
    \centering
    {\includegraphics[width=\linewidth]{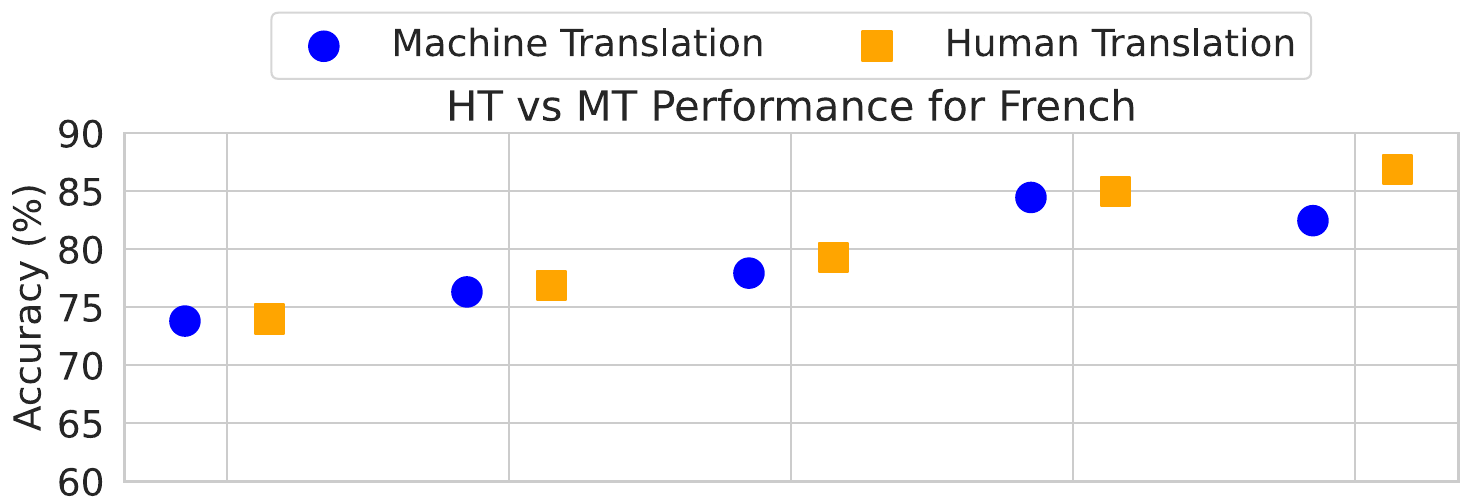}}
    \end{subfigure}
    \begin{subfigure}[b]{0.60\textwidth}
    \centering
    {\includegraphics[width=\linewidth]{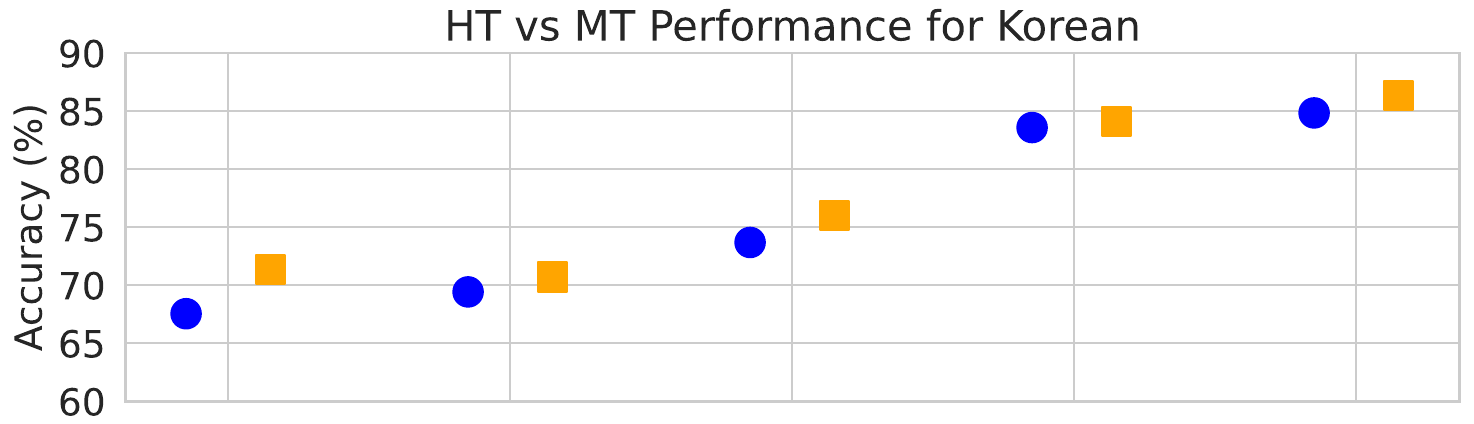}}
    \end{subfigure}
    \begin{subfigure}[b]{0.60\textwidth}
    \centering
    {\includegraphics[width=\linewidth]{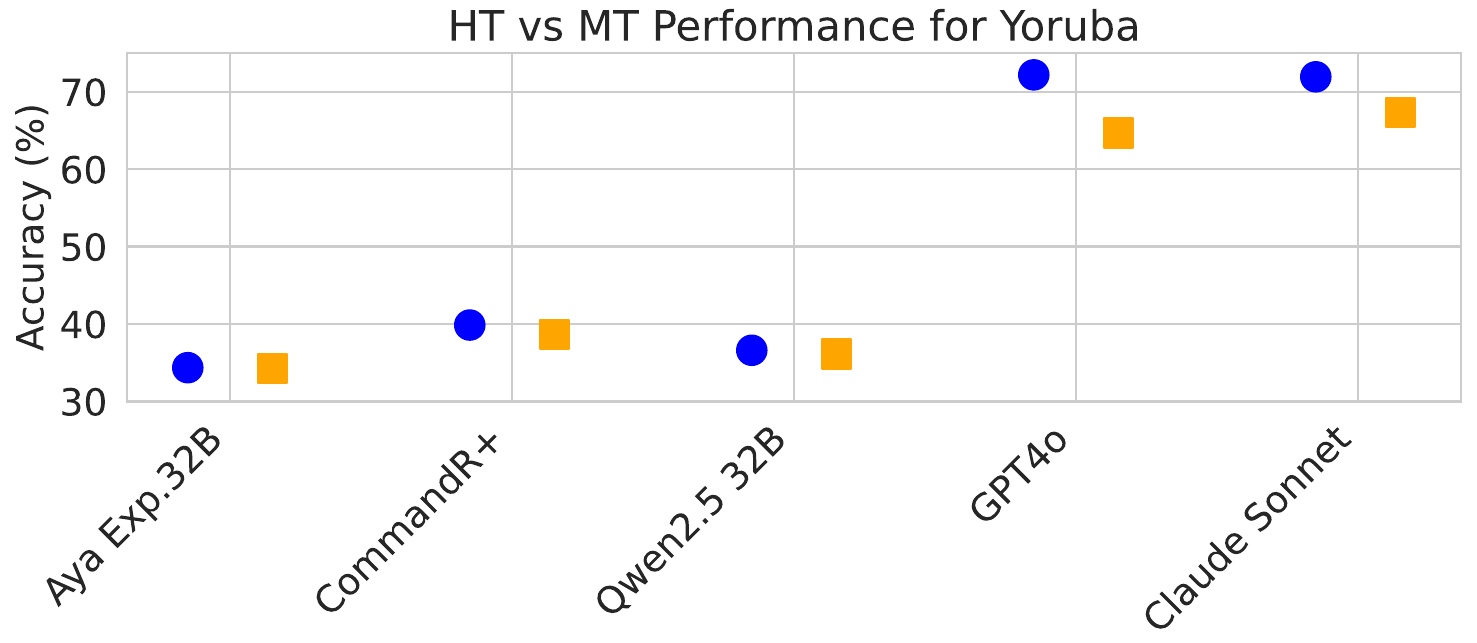}}
    \end{subfigure}
    \caption{Comparison of model performance on \emph{human-translated} and \emph{machine-translated} \cs{} in French, Korean, and Yoruba.}
    \label{fig:model_evaluations_on_ht_mt_cs}
\end{figure}

\textbf{Human Translated vs. Machine Translated.} 
We compared models on Human-Translated (HT) and Machine-Translated (MT) \cs{} datasets to gain deeper insights into model behavior. Figure~\ref{fig:model_evaluations_on_ht_mt_cs} illustrates the model performances for one \highemoji{}high-resource  language (French), one \midemoji {mid-resource} language (Korean), one \lowemoji{low-resource} language (Yoruba).

The key finding is that models generally perform better on human-translated data for \highemoji{}high-resource  languages. This is likely because these languages benefit from extensive in-language training data. However, this trend shifts for \midemoji {mid-resource} languages. The figure reveals that the performance gap between HT and MT narrows for models such as Claude Sonnet and Qwen2.5 32B. Conversely, models like CommandR+ and Aya Expanse 32B continue to perform better on HT data. Notably, these two models have strong Korean language support, which can be attributed to a substantial amount of in-language training data.

For \lowemoji{low-resource} languages, a distinct pattern emerges. As shown in the figure, models such as Claude Sonnet and GPT-4o perform significantly better on MT data than on HT data. Similarly, CommandR+ and Qwen2.5 32B also show improved performance on MT data, albeit with less pronounced differences. This behavior is likely because these models primarily rely on machine-translated data for low-resource languages during training, and the distribution of the machine-translated test set aligns more closely with their training data. Notably, the only model demonstrating consistent performance across both HT and MT datasets is Aya Expanse 32B, which can be attributed to its broad coverage and strong support for low-resource languages.

These results underscore the importance of in-language or human-translated datasets for evaluating low-resource languages. The \gmmlu{} dataset provides a valuable tool for assessing the in-language performance of large language models (LLMs) on low-resource languages, offering insights into their capabilities and limitations in such contexts.

\section{Related Work}
\label{sec:related_work}

\subsection{Multilingual Knowledge Evaluation}

As the MMLU benchmark has become a standard for evaluating LLMs \citep{beeching2023open,openai2024gpt4technicalreport,dubey2024llama,ustun2024ayamodelinstructionfinetuned,aryabumi2024aya23openweight}, addressing its limitations and introducing enhancements are essential to maintaining high evaluation standards.
For English, \citet{gema2024mmlu} manually re-annotated 3K questions across 30 MMLU subjects to identify quality or problematic questions and released it as MMLU-redux. \citet{wang2024mmluprorobustchallengingmultitask} introduced an extended version of this dataset, MMLU-Pro, which adds more challenging, reasoning-focused questions and expands the answer choice set from four to ten options.
MMLU-Pro+ extends the previous work by incorporating questions with multiple correct answers across diverse domains and evaluating higher-order reasoning in LLMs \citep{taghanaki2024mmluproevaluatinghigherorderreasoning}.While these efforts enhance the difficulty and diversity of tasks, they remain restricted to English alone.

Language-specific variants of comprehensive multiple-choice exam benchmarks are typically centered around a single language. 
Examples include ArabicMMLU \citep{koto2024arabicmmluassessingmassivemultitask}, CMMLU \citep{li2024cmmlumeasuringmassivemultitask}, IndoMMLU \citep{koto2023largelanguagemodelspass}, ThaiExam \citep{pipatanakul2023typhoonthailargelanguage},
, TurkishMMLU \citep{yüksel2024turkishmmlumeasuringmassivemultitask},  AfriMMLU \citep{Adelani2024IrokoBenchAN}, 
Khayyam Challenge \citep{ghahroodi2024khayyamchallengepersianmmlullm}, KMMLU \citep{son2024kmmlumeasuringmassivemultitask}, HAE-RAE \citep{son2024haeraebenchevaluationkorean} and
VNHSGE \citep{dao2023vnhsgevietnamesehighschool} covering Arabic, Chinese, Indonesian, Thai, Turkish, Persian, Korean, and Vietnamese, respectively. 

There have been multiple efforts to design and construct evaluation datasets that cater to multilingual settings. AGIEval is a compilation of human-centric standardized exams to assess language model performance in English and Chinese \citep{zhong2023agievalhumancentricbenchmarkevaluating}. BEnQ is similar but for English and Bengali \citep{shafayat2024benq}.  EXAMS is a multilingual high school examination collection covering 16 languages \citep{hardalov-etal-2020-exams}.
M3EXAMS is a multimodal multilingual benchmark supporting 9 languages with three educational levels \citep{zhang2023m3exammultilingualmultimodalmultilevel}. 
Both evaluation sets process exams on various topics in different countries and build per-language benchmarks. These initiatives strive to evaluate the performance of language models across various languages; however, they often support a small number of languages and lack a consistent, standardized framework for direct comparison between languages. We note recent work INCLUDE as an exception to this as one of the most extensive evaluation benchmarks, compiled from local exams across various countries and languages, covering 44 languages \citep{romanou2024includeevaluatingmultilinguallanguage}.

To enable evaluation across a wider range of languages, efforts have also been made to translate the MMLU dataset into multiple languages.
\citet{lai2023okapi} use ChatGPT to translate the English MMLU dataset into 26 languages. However, the quality of translations produced by ChatGPT can vary significantly across different languages and is not always reliable \citep{robinson2023chatgpt}. More recently OpenAI released MMMLU by translating MMLU into 14 languages using professional human translators, and we incorporate this high-quality dataset into our benchmark.

\subsection{Culturally-aware Evaluation}

Recent research has increasingly focused on examining the cultural alignment of LLMs. Studies such as \citet{arora2022probing} and \citet{cao2023assessing} have explored LLMs’ ability to understand cross-cultural differences in values and beliefs.  To ensure accurate cross-cultural and cross-linguistic representation, SEA-HELM\footnote{An acronym for \textbf{S}outh\textbf{E}ast \textbf{A}sian \textbf{H}olistic \textbf{E}valuation of \textbf{L}anguage \textbf{M}odels.} (previously known as BHASA \citep{leong2023bhasa})\footnote{ \url{https://leaderboard.sea-lion.ai}} is an evaluation suite which emphasizes Southeast Asian languages and contains a variety of tasks, including manually handcrafted linguistic diagnostics as well as manually translated and validated SEA-IFEval and SEA-MTBench. \citet{wang2023not} and \citet{masoud2024culturalalignmentlargelanguage} demonstrate that LLMs often reflect values and opinions aligned with Western culture, a trend that persists across multiple languages. 
Additionally, benchmarks like those introduced by \citet{naous2024havingbeerprayermeasuring} and \citet{rao2024normadframeworkmeasuringcultural} aim to measure cultural biases in LLMs, while \citet{ventura2024navigatingculturalchasmsexploring} investigates cultural biases within text-to-image diffusion models, proposing a comprehensive suite of cultural evaluation techniques. \citet{aakanksha-etal-2024-multilingual} studied aligning language models balancing dual objectives: addressing and optimizing for a non-homogeneous set of languages and cultural preferences based upon annotations from professional multilingual annotators while minimizing both global and local harms. Some studies focus on specific cultural aspects, such as \citet{myung2024blend}, \citet{magomere2024you}, and \citet{montalan2024kalahihandcraftedgrassrootscultural}, which evaluate LLMs' understanding of everyday cultural knowledge across diverse cultures and regions. 

In addition, several studies have explored evaluating multilingual visual language models (VLMs).
PangeaBench is a holistic evaluation suite encompassing 14 pre-existing datasets covering 47 languages \citep{yue2024pangeafullyopenmultilingual}. \cite{romero2024cvqa} presents CVQA, a culturally diverse multilingual Visual Question Answering benchmark that includes culturally-driven images and questions across 30 countries and 31 languages. \citet{vayani2024languagesmatterevaluatinglmms} introduces a multimodal benchmark including culturally diverse images paired with text across 100 languages. 

Numerous studies have also explored the role of pre-training in shaping the cultural biases present in LLMs. For example, \cite{chen2024gooddatamultilingualinstruction} examines the impact of native versus translated data on LLM instruction tuning and evaluation. Their findings reveal that models fine-tuned with native instructions typically outperform those trained using translated data. Similarly, \cite{choenni2024evaluationpracticesmultilingualnlp} investigates the reliability of machine translation as a substitute for human translation in large-scale multilingual evaluations, highlighting its effectiveness across a diverse set of languages. \citet{ustun2024ayamodelinstructionfinetuned} released the Aya-101 model and focused on in-language prompting and using a comprehensive dataset of human-written data for instruction tuning large language models across 114 languages to reflect local culture and preferences \citep{singh-etal-2024-aya}. Additionally, significant efforts have been made to incorporate knowledge from various cultures into LLMs to achieve broader cultural alignment. For instance, \cite{li2024culturegenrevealingglobalcultural} proposes a cost-effective fine-tuning strategy to embed cultural differences into LLMs, facilitating better representation and understanding of global cultural nuances.
Meanwhile, \citet{alkhamissi2024investigating} introduces ``Anthropological Prompting'' a novel method that employs anthropological reasoning to enhance the cultural alignment of LLMs. 

\subsection{Participatory Open Science Projects}

Participatory research empowers diverse communities to actively contribute to the research process, ensuring that outcomes are inclusive, contextually relevant, and address real-world needs.
Previous participatory research efforts have primarily focused on specific regions or tasks such as translation, character recognition, audio segmentation, and transcription. For instance, \citet{clanuwat2018deep} addressed the challenge of reading and understanding Kuzushiji, an old cursive style of Japanese writing no longer commonly used. Another notable example of culturally diverse data collection is MaRVL (Multicultural Reasoning over Vision and Language; \citealp{liu-etal-2021-visually}), where native speakers of five typologically, genealogically, and geographically diverse languages (\textit{Indonesian, Swahili, Tamil, Turkish}, and \texttt{Mandarin Chinese}) contributed images reflecting their cultures. Professional linguists fluent in these languages then wrote captions for the images. However, MaRVL’s dataset is relatively small, with fewer than 8,000 data points, limiting its use to evaluation purposes. 
Similarly, \citet{hernandez-mena-meza-ruiz-2022-creating} developed eight open-access resources for Mexican and Latin American Spanish by establishing a social service program where students voluntarily contributed to tasks like audio segmentation and transcription. Notably, these efforts are largely concentrated on image and speech, unlike our work, which focuses on text.
\citet{CaneteCFP2020} spearheaded the collection of a Latin American Spanish dataset to train a language model. 
\citet{guevara-rukoz-etal-2020-crowdsourcing} explored the development of a crowd-sourced corpus for Latin American Spanish dialects to address resource scarcity for these languages.
Masakhane utilized a participatory research framework to curate  NLP datasets and build models for several underrepresented African languages~\citep{forall-nekoto-etal-2020-participatory,adelani-etal-2021-masakhaner,adelani-etal-2023-masakhanews}.
Aligned with the goals of having a participatory framework and open-access resources, Project SEALD,\footnote{An acronym for \textbf{S}outh\textbf{e}ast \textbf{A}sian \textbf{L}anguages in One Network \textbf{D}ata.}{$^{,}$}\footnote{\url{https://aisingapore.org/aiproducts/southeast-asian-languages-in-one-network-data-seald/}} a collaboration between AI Singapore and Google Research, pioneered multilingual data collection for Large Language Models (LLMs) in Southeast Asia (SEA). The output of this project continues to contribute to the development of open-source multilingual models in this region, namely SEA-LION\footnote{\url{https://sea-lion.ai}} and its derivatives, such as WangchanLion \citep{phatthiyaphaibun2024wangchanlion} and Sahabat-AI.\footnote{\url{https://sahabat-ai.com}} Similarly, the NusaCrowd initiative by \citet{cahyawijaya-etal-2023-nusacrowd} focused on aggregating and standardizing data sources for Indonesian languages. The ongoing SEACrowd project\footnote{\url{https://github.com/SEACrowd}} represents a similar effort, aiming to standardize data resources for all Southeast Asian languages \citep{lovenia2024seacrowd}. The Aya Initiative, through a global community effort of 3,000 contributors, collected instruction data in 114 languages, fostering linguistic diversity and inclusivity to create one of the largest multilingual datasets for advancing state-of-the-art language models \citep{singh-etal-2024-aya,ustun2024ayamodelinstructionfinetuned}.

\section{Conclusion}
\label{sec:conclusion}

We evaluate the cultural biases present in MMLU and find that 28\% of all questions require culturally-sensitive knowledge. In particular, progress on MMLU depends heavily on learning Western-centric concepts. For questions requiring geographic knowledge, the vast majority focus on North America and Europe. This cultural bias remains in translated variants of MMLU that are widely used for multilingual LLM evaluation, which reduces the dataset’s practical effectiveness as a global benchmark and risks over-indexing evaluations on Western-centric idioms and knowledge.

We examine the impact of translation artifacts and cultural bias on multilingual model rankings. We introduce \gmmlu{} and \gmmlulite{}, multilingual multi-domain datasets that distinguish between culturally-sensitive (\cs{}) and culturally-agnostic (\ca{}) knowledge. By incorporating professional and crowd-sourced annotations, these subsets enable rigorous multilingual model evaluation.

Finally, we evaluate a large group of state-of-the-art open-weight and proprietary models to understand performance differences on both these subsets. We find that model rankings change depending on whether models are assessed on culturally-sensitive or culturally-agnostic subsets, highlighting that progress on translated MMLU is insufficient as an indicator of performance. Instead, we recommend evaluations for multilingual reports on \gmmlu{} and both \ca{} and \cs{} subsets as part of the holistic evaluation of progress in multilingual LLM capabilities. As part of our commitment to the research ecosystem, we release \gmmlu{} and \gmmlulite{} under a fully permissive license for use in evaluations at \url{https://hf.co/datasets/CohereForAI/Global-MMLU} and \url{https://huggingface.co/datasets/CohereForAI/Global-MMLU-Lite}.

\section{Limitations}
\label{sec:limitation}

\textbf{Uneven distribution of contributions} Beyond the gold standard languages where we engaged with compensated annotators, participation from community annotators was heavily skewed across languages. Despite a large volume of community annotators, there was a ‘long tail’ of annotators only contributing one or two annotations. Similarly, there is a huge gap between languages with the highest number of contributions and ones with the lowest number of contributions. Consequently, this suggests potential unevenness in dataset distributions across different languages and a lack of annotator diversity within some languages dominated by one or two frequent contributors. 

\textbf{Language and dialect coverage} We focus on 42 lanugages for \gmmlu. However, this is still only a tiny fraction of the world’s linguistic diversity. Of the world’s approximately 7,000 languages, only half of them are captured in any sort of written form \citep{ADDA20168}. Of this half, only a few hundred are included on the internet in machine readable corpora \citep{ADDA20168}. Future work is needed to continue to improve evaluations beyond these 42 languages and to take into account how technology serves different dialects (a topic we do not address here). Geo-cultural variation within a language often gives rise to new dialects or creoles over time \citep{zampieri2020natural, wolfram1997issues} and, as such, dialects can serve an important function in establishing and maintaining cultural identity\citep{falck2012dialects}. Many different dialects that are generally recognized as belonging to a single parent language are not represented in this evaluation dataset.

\textbf{Toxic or offensive speech} Our annotation interface does not contain specific flags for toxic, harmful, or offensive speech, so it is possible that \gmmlu{} contains some data that could be considered harmful. We believe this is of relatively low risk because of the nature of the original MMLU and the focus on examination material. However, we did not monitor or track this explicitly during our cultural sensitivity annotations or translation post-edits.

\textbf{Region Category Assignment:} For the annotation of geographically sensitive questions, we classified regions into six geographic regions (Africa, Asia, Europe, North America, Oceania, and South America).\footnote{\url{https://www.pewresearch.org/global/2013/06/04/regional-categorization/}} However, based upon discussions we would going forward recommend switching to the taxonomy proposed by the World Bank which is more granular and includes separate designations for Central America and Sub-Saharan Africa.\footnote{\url{https://ourworldindata.org/world-region-map-definitions}}

\textbf{Identifying cultural sensitivity does not guarantee cultural inclusion.} We acknowledge that efforts like the proposed Global-MMLU highlight important limitations in current datasets by identifying gaps in non-Western cultural representation. Identifying whether a dataset is culturally agnostic or not is highly relevant as mere translations may create the illusion that datasets are being more culturally inclusive and validating models in that sense, while this is not the real case. However, it must be noted that they do not fully resolve the issue. Future work must prioritize the integration of diverse culturally grounded knowledge to achieve true inclusivity and fairness in multilingual AI evaluation.

\section{Acknowledgments}
\label{sec:acknowledgements}

We would like to thank members of the Cohere For AI community who championed this initiative
and helped with annotating samples for cultural sensitivity as well as improving translation quality across many languages. In particular, we recognize Ashay Srivastava, Aurélien-Morgan Claudon, Bevnm SaiAsrit, Danylo Boiko, Hanna Yukhymenko, Sai Vineetha Baddepudi Venkata Naga Sri, Sangyeon Kim, Tadesse Destaw Belay, Alperen Ünlü, Mohammed Hamdy, Muhammad Rafi Sudrajat, Olusanya Joy Naomi, Vu Trong Ki, Yiyang Nan, Abdelmoneim Shahd, Arwa ALaya, Bimasena Putra, Emad Alghamdi, Fabian Farestam, Mridul Sharma, Sayuru Bopitiya, Surya Abhinai who contributed a significant amount to each of their languages. A special thank you to Claire Cheng and Trisha Starostina for helping to coordinate the Cohere professional annotators who contributed to this project. We thank all these compensated experts who provided their language knowledge to comprehensively improve quality over our gold languages.

\bibliography{main}

\appendix

\section{Global-MMLU Languages}
\label{app:mmlu_global_languages}

In this work we will refer to groups of languages to be ``lower-'', ``mid-'' or ``higher''-resourced according to their recorded, written, and catalogued NLP resources~\citep{joshi2020state}.
We group these 5 distinct clusters following the groupings in~\citep{singh-etal-2024-aya} into a rough taxonomy of \textbf{lower-resourced (LR)}, \textbf{mid-resourced (MR)} and \textbf{higher-resourced (HR)}.
We note that this grouping is inevitably imperfect; languages and their varieties cannot absolutely nor universally be classified based on this single dimension~\citep{H_m_l_inen_2021,bird-2022-local}. The categorization in our case serves the purpose of aggregation in our analysis of the data distribution.

\begin{center}
\scriptsize
\begin{longtable}{lllllcc}
\toprule
ISO Code & Language & Script & Resource & Type \\
\midrule 
am & Amharic & Ge'ez & Low & {\dia} \spade \\
ar & Arabic & Arabic                & High           & \spade \\
bn & Bengali                   & Bengali       & Mid           &  \spade \\
cs & Czech & Latin  & High & {\dia} \spade \\ 
de & German                    & Latin       & High          & \spade \\ 
el & Greek                     & Greek        & Mid           & {\dia} \\ 
en & English                   & Latin        & High          & {\dia} \spade \\ 
fil & Filipino                  & Latin        & Mid           & {\dia} \\
fr & French                    & Latin         & High          & \spade \\ 
ha & Hausa                     & Latin        & Low           & {\dia} \\ 
he & Hebrew                    & Hebrew       & Mid           & {\dia} \\ 
hi & Hindi                     & Devanagari     & High           & \spade \\ 
ig & Igbo                      & Latin         & Low           & {\dia} \\
id & Indonesian                & Latin       & Mid           &  \spade \\ 
it & Italian                   & Latin          & High           & \spade \\ 
ja & Japanese                  & Japanese      & High          & \spade \\ 
ky & Kyrgyz                    & Cyrillic       & Low           & {\dia} \\
ko & Korean                    & Hangul        & Mid           & \spade \\ 
lt & Lithuanian                & Latin     & Mid           & {\dia} \\
mg & Malagasy          & Latin        & Low           & {\dia} \\
ms & Malay            & Latin           & Mid           & {\dia} \spade \\ 
ne & Nepali        & Devanagari                & Low           & {\dia} \\
nl & Dutch          & Latin          & High   & {\dia} \\ 
ny & Nyanja & Latin   & Low           & {\dia} \\
fa & Persian           & Arabic            & High           & {\dia} \spade \\ 
pl & Polish      & Latin            & High           & {\dia} \\ 
pt & Portuguese                & Latin           & High           & \spade \\ 
ro & Romanian & Latin  & Mid & {\dia} \spade \\ 
ru & Russian                   & Cyrillic       & High & {\dia} \spade \\ 
sin & Sinhala  & Sinhala              & Low           & {\dia} \spade \\
sn & Shona                     & Latin         & Low           & {\dia} \\ 
som & Somali                    & Latin     & Low           & {\dia}  \\ 
es & Spanish                   & Latin         & High          & \spade \\ 
sr & Serbian      & Cyrillic             & High           & {\dia} \\ 
sw & Swahili                   & Latin          & Low           & \spade \\ 
sv & Swedish         & Latin            & High           & {\dia} \\ 
te & Telugu                    & Telugu        & Low           & {\dia} \spade \\ 
tr & Turkish                   & Latin       & High           & {\dia} \spade \\ 
uk & Ukrainian       & Cyrillic       & Mid           & {\dia} \spade \\ 
vi & Vietnamese                & Latin        & High           & {\dia} \spade \\ 
yo & Yorùbá                    & Latin        & Low           & \spade \\ 
zh & Chinese       & Hans        & High          & \spade \\
\bottomrule
\caption{42 languages in \gmmlu, along with each language's script and resource category. We followed \citep{singh-etal-2024-aya} and categorized languages as low, mid and high resource based on language classes proposed by \citep{joshi2020state} (low: [0, 1, 2], mid: [3], high: [4, 5]). In \emph{Global-MMLU}, the language is either fully machine translated \specialdia, fully human translated \specialspade, or contains both machine and human translated data \specialdia \specialspade.}
\label{tab:language_codes}
\end{longtable}
\end{center}

\section{\gmmlu{} Subject Categories}
\label{app:mmlu_subject_categories}

\gmmlu{} covers six diverse subject categories: STEM, Humanities, Social Sciences, Medical, Business, and Other. For a consistent approach, we adopt the classification proposed by \cite{hendrycks2020measuring} for the MMLU dataset to categorize subjects as STEM, Humanities, and Social Sciences. 
However, we further refine the 'Other' category from the original MMLU dataset by breaking it down into two distinct categories: Medical and Business. Within the 'Other' category, subjects such as clinical knowledge, college medicine, human aging, medical genetics, nutrition, professional medicine, and virology are classified under the Medical category. 
Meanwhile, business ethics, management, marketing, and professional accounting fall under the Business category. It's worth noting that the 'Other' category in \gmmlu{}, sometimes referred to as 'General Knowledge', includes the remaining two subjects from the original MMLU 'Other' category: global facts and miscellaneous.

\section{\gmmlulite{}}
\label{app:gmmlu_lite}

\begin{figure}[ht!]
    \centering\includegraphics[width=0.75\columnwidth]{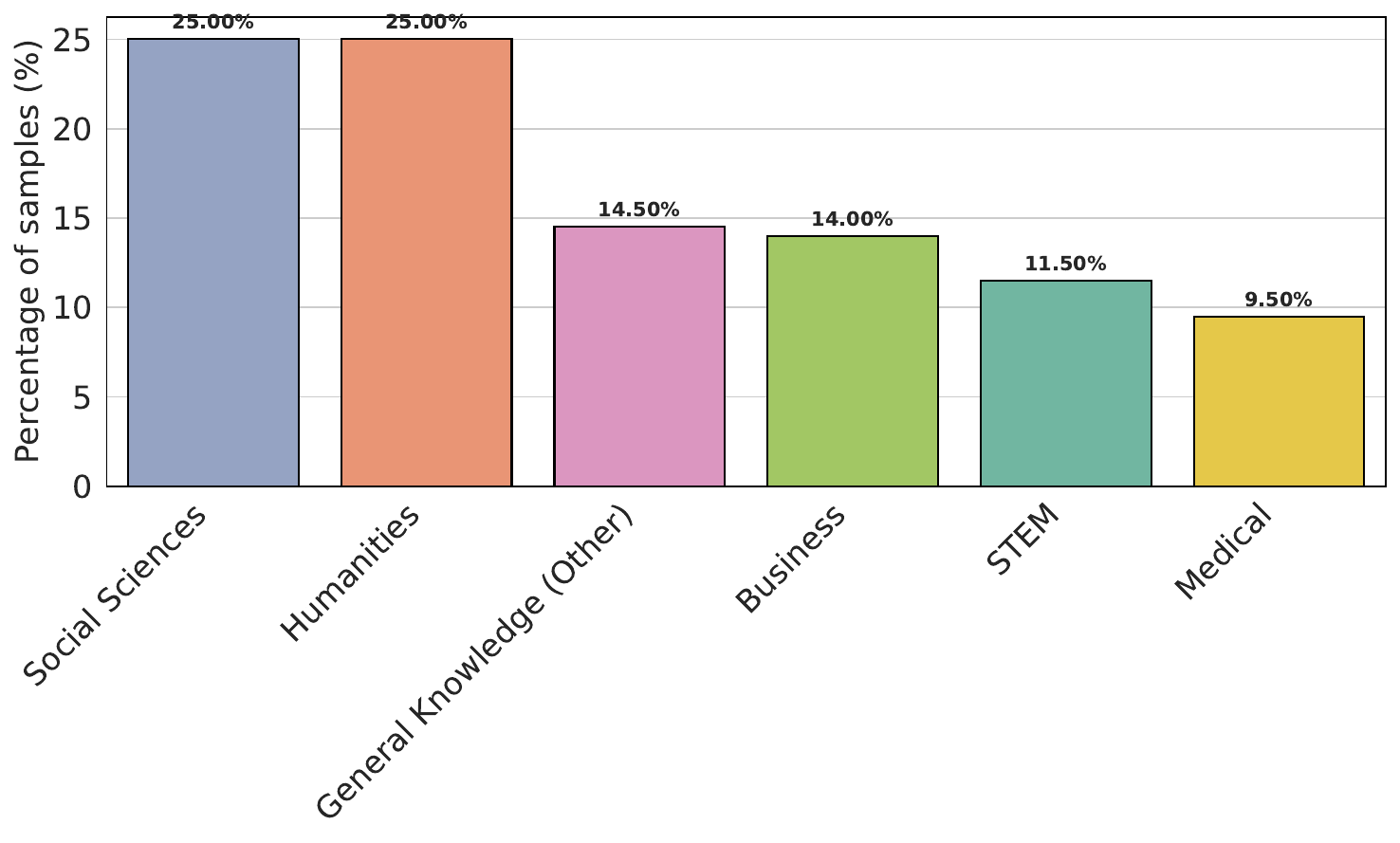}
    \caption{Distribution of samples across subject categories in \gmmlulite{}}
    \label{fig:gmmlu_lite_subject_cat_proportion}
\end{figure}

As mentioned in section~\ref{sec:data_comp_gmmlu}, \gmmlulite{} is a lighter version of \gmmlu{} containing 200 \cs{} and 200 \ca{} samples per language for 15 human-translated or post-edited languages, including English. 

For preparing \gmmlulite{}, we took the MA subset of \gmmlu{} containing 50 samples per subject and looked at proportion of CS and CA samples available per subject. Subjects exclusively tagged as \cs{} or \ca{} (14 in total) were excluded to ensure both categories were represented within each subject. Consequently, Social Sciences and Humanities subjects are more prevalent in \gmmlulite{}, as shown in Figure~\ref{fig:gmmlu_lite_subject_cat_proportion}. 

However, we aimed for a balanced distribution across subject categories. Social Science subjects like High School Geography and Sociology had higher proportion of CS samples whereas STEM subjects like Abstract Algebra had higher number of CA samples. To maintain balance, we sampled five \cs{} and five \ca{} samples per subject where available. Few subjects like Anatomy or High School Mathematics had only one CS sample available, so for such subjects, only one CS and one CA sample was taken. Samples from few subjects of Business and Medical categories were slightly upsampled to ensure adequate representation.

The General Knowledge category, comprising only Miscellaneous and Global Facts, was also upsampled, with 22 samples from Miscellaneous and 8 from Global Facts per category. This adjustment ensures sufficient coverage for evaluating general knowledge capabilities.
The overall goal with \gmmlulite{} is to have a balanced dataset for efficient multilingual evaluation across multiple languages.

\section{Temporal Knowledge}
\label{app:time-sensitivity}

\begin{figure}[ht!]
    \centering\includegraphics[width=0.75\textwidth]{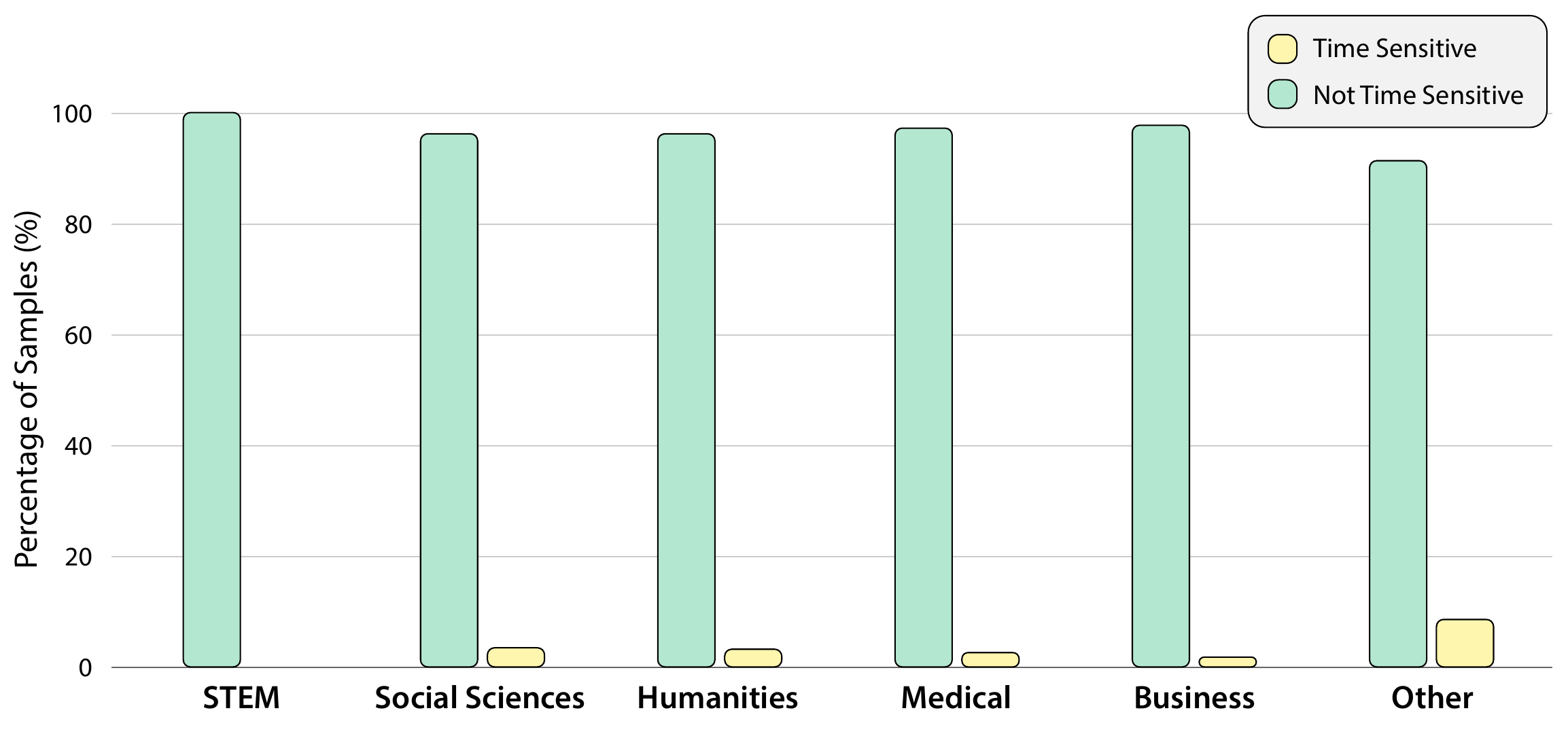}
    \caption{Distribution of time-sensitive samples across subject categories. Note that STEM subjects do not include any temporal knowledge.}
    \label{fig:mmlu_time_sensitive}
\end{figure}

As part of the annotation process, annotators were also asked to label samples for temporal or time-sensitive knowledge \timeremoji. This applies to questions where the correct answer may change over time due to factors such as current political leaders or economic statistics.
Figure~\ref{fig:mmlu_time_sensitive} shows the distribution of time sensitive samples in \textbf{MMLU Annotated\memoemoji}. Overall it is observed that only 2.4\% of the dataset is tagged as time-sensitive and majority of these samples fall under Social Sciences, Humanities, Medical and Other categories. STEM is the only category with no time sensitive samples at all.

\section{Models Covered}
\label{app:models}

Details of each model are descibed below:
\begin{itemize}
\item \textbf{Aya Expanse}\footnote{\url{https://hf.co/blog/aya-expanse}} family of models include 8B\footnote{\url{https://hf.co/CohereForAI/aya-expanse-8b}} and 32B\footnote{\url{https://hf.co/CohereForAI/aya-expanse-32b}} parameter models. 
Aya Expanse models support 23 languages including Arabic, Chinese (simplified \& traditional), Czech, Dutch, English, French, German, Greek, Hebrew, Hindi, Indonesian, Italian, Japanese, Korean, Persian, Polish, Portuguese, Romanian, Russian, Spanish, Turkish, Ukrainian, and Vietnamese. Aya Expanse builds on the Aya initiative which includes multilingual first releases like Aya 101 \citep{ustun2024ayamodelinstructionfinetuned}, Aya 23 \citep{aryabumi2024aya23openweight} and extensive multilingual datasets such as Aya collection \citep{singh-etal-2024-aya}.
\item \textbf{Command R and R+} are open-weight models of size 34B\footnote{\url{https://hf.co/CohereForAI/c4ai-command-r-08-2024}} and 104B\footnote{\url{https://hf.co/CohereForAI/c4ai-command-r-plus-08-2024}} respectively which both support 10 languages: \emph{English, French, Spanish, Italian, German, Brazilian Portuguese, Japanese, Korean, Arabic, Simplified Chinese}. We use Command-R 08-2024 and Command-R+ 08-2024 for evaluation.
\item \textbf{Gemma2} ~\citep{team2024gemma} is part of the Gemma model family. The languages targeted are not explicitly reported. We evaluate the instruct-tuned 9B (gemma-2-9b-it) and 27B (gemma-2-27b-it) variants. 
\item \textbf{Gemma2-9B-CPT-SEA-LIONv3}\footnote{\url{https://hf.co/aisingapore/gemma2-9b-cpt-sea-lionv3-instruct}} is part of the SEA-LION\footnote{An acronym for \textbf{S}outh\textbf{e}ast \textbf{A}sian \textbf{L}anguages \textbf{i}n \textbf{O}ne \textbf{N}etwork.}{$^{,}$}\footnote{\url{https://github.com/aisingapore/sealion}} collection of models trained for Southeast Asian (SEA) languages, including Burmese, Chinese, English, Filipino, Indonesian, Javanese, Khmer, Lao, Malay, Sundanese, Tamil, Thai, and Vietnamese. We use Gemma2-9B-CPT-SEA-LIONv3-Instruct for evaluation.
\item \textbf{Llama 3.1} ~\citep{dubey2024llama} Llama 3.1 is a series of open LLM models that come in three sizes: 8B, 70B, and 405B parameters. All variants support 8 languages, including English, German, French, Italian, Portuguese, Hindi, Spanish, and Thai. We use Llama-3.1-8B-Instruct and Llama-3.1-70B-Instruct for evaluation.
\item \textbf{Mistral Nemo}\footnote{\url{https://hf.co/mistralai/Mistral-Nemo-Instruct-2407}} is a 12B model which supports 11 languages including English, French, German, Spanish, Italian, Portuguese, Chinese, Japanese, Korean, Arabic, and Hindi. 
\item \textbf{Qwen 2.5}\footnote{\url{https://huggingface.co/collections/Qwen/qwen25-66e81a666513e518adb90d9e}} models support up to 29 languages, including Chinese, English, French, Spanish, and Portuguese. We evaluate Qwen2.5-7B-Instruct and Qwen2.5-32B-Instruct variants of Qwen 2.5.
\item \textbf{GPT-4o} \citep{hurst2024gpt} is a multilingual, multimodal closed-model and is part of the GPT-4 family. The languages targeted are not explicitly reported.
\item \textbf{Claude Sonnet 3.5} is also a multilingual, multimodal closed-model from the Claude 3.5 family. The languages supported by this model are also unknown. 

\end{itemize}

\section{Additional Results}

\subsection{Model Rank Changes}
\label{app:rank_changes}

Table~\ref{tab:model_rank_changes_high_res} presents the rank changes and corresponding position shifts (indicated next to the arrows) for high-resource languages, while Table~\ref{tab:model_rank_changes_mid_low_res} provides similar data for mid- and low-resource languages. The rightmost columns in each table summarize the total number of models that changed ranks (\textit{Total Rank Change}) and the total number of position shifts in the rankings (\textit{Total Position Change}). A detailed analysis of these results is provided in Section~\ref{sec:results}.

\begin{table}[ht!]
\centering
\scalebox{0.60}{
\begin{tabular}{l|c|cccccccccccccc|cc}
\toprule
\textbf{Language} &  \textbf{Dataset}  & \rotatebox{90}{Aya Exp. 8B} & \rotatebox{90}{Aya Exp. 32B} & \rotatebox{90}{CommandR} & \rotatebox{90}{CommandR+} & \rotatebox{90}{Gemma2 9B} & \rotatebox{90}{Gemma2 27B} & \rotatebox{90}{Llama-3.1 8B} & \rotatebox{90}{Llama-3.1 70B} & \rotatebox{90}{Mistral Nemo} & \rotatebox{90}{Qwen2.5 7B} & \rotatebox{90}{Qwen2.5 32B} & \rotatebox{90}{SEA-LION-v3} & \rotatebox{90}{GPT4o} & \rotatebox{90}{Claude Sonnet} & \rotatebox{90}{\textbf{Total rank change}} & \rotatebox{90}{\textbf{Total position change}} \\
\midrule
\multirow{2}{*}{\highemoji Arabic} &
\scalesemoji &
- & 
- & 
- & 
- & 
- & 
- & 
- & 
- & 
- & 
\tcbox[colback=LightGreen]{\textcolor{black}{$\uparrow$1}} & 
- & 
\tcbox[colback=LightRed]{\textcolor{black}{$\downarrow$1}} & 
- & 
- & 2 & 2 \\ & 
\statuelibertyemoji & 
- & 
\tcbox[colback=LightGreen]{\textcolor{black}{$\uparrow$1}} & 
- & 
- & 
- & 
\tcbox[colback=LightRed]{\textcolor{black}{$\downarrow$1}} & 
- & 
\tcbox[colback=LightGreen]{\textcolor{black}{$\uparrow$1}} & 
- & 
- & 
\tcbox[colback=LightRed]{\textcolor{black}{$\downarrow$1}} & 
- & 
- & 
- &4 & 4 \\ \midrule
\multirow{2}{*}{\highemoji Chinese} &
\scalesemoji &
- & 
- & 
\tcbox[colback=LightRed]{\textcolor{black}{$\downarrow$1}} & 
- & 
\tcbox[colback=LightGreen]{\textcolor{black}{$\uparrow$1}} & 
- & 
- & 
- & 
- & 
- & 
\tcbox[colback=LightGreen]{\textcolor{black}{$\uparrow$1}} & 
- & 
\tcbox[colback=LightRed]{\textcolor{black}{$\downarrow$1}} & 
- & 4 & 4 \\ & 
\statuelibertyemoji & 
\tcbox[colback=LightGreen]{\textcolor{black}{$\uparrow$1}} & 
\tcbox[colback=LightGreen]{\textcolor{black}{$\uparrow$1}} & 
\tcbox[colback=LightGreen]{\textcolor{black}{$\uparrow$1}} & 
\tcbox[colback=LightGreen]{\textcolor{black}{$\uparrow$2}} & 
\tcbox[colback=LightGreen]{\textcolor{black}{$\uparrow$1}} & 
- & 
\tcbox[colback=LightRed]{\textcolor{black}{$\downarrow$1}} & 
\tcbox[colback=LightGreen]{\textcolor{black}{$\uparrow$1}} & 
- & 
\tcbox[colback=LightRed]{\textcolor{black}{$\downarrow$3}} & 
\tcbox[colback=LightRed]{\textcolor{black}{$\downarrow$1}} & 
\tcbox[colback=LightRed]{\textcolor{black}{$\downarrow$2}} & 
\tcbox[colback=LightGreen]{\textcolor{black}{$\uparrow$1}} & 
\tcbox[colback=LightRed]{\textcolor{black}{$\downarrow$1}} & 12 & 16 \\ \midrule
\multirow{2}{*}{\highemoji Czech} &
\scalesemoji &
- & 
- & 
- & 
- & 
- & 
- & 
- & 
\tcbox[colback=LightRed]{\textcolor{black}{$\downarrow$1}} & 
- & 
- & 
\tcbox[colback=LightGreen]{\textcolor{black}{$\uparrow$1}} & 
- & 
- & 
- & 2 & 2 \\ & 
\statuelibertyemoji & 
\tcbox[colback=LightGreen]{\textcolor{black}{$\uparrow$2}} & 
\tcbox[colback=LightRed]{\textcolor{black}{$\downarrow$1}} & 
- & 
\tcbox[colback=LightGreen]{\textcolor{black}{$\uparrow$3}} & 
- & 
\tcbox[colback=LightRed]{\textcolor{black}{$\downarrow$1}} & 
\tcbox[colback=LightRed]{\textcolor{black}{$\downarrow$2}} & 
- & 
- & 
- & 
\tcbox[colback=LightRed]{\textcolor{black}{$\downarrow$1}} & 
- & 
- & 
- & 6 & 10 \\ \midrule
\multirow{2}{*}{\highemoji Dutch} &
\scalesemoji &
- & 
- & 
- & 
- & 
- & 
- & 
- & 
- & 
- & 
- & 
- & 
- & 
- & 
-  & 0 & 0 \\ & 
\statuelibertyemoji & 
- & 
- & 
- & 
\tcbox[colback=LightGreen]{\textcolor{black}{$\uparrow$1}} & 
\tcbox[colback=LightGreen]{\textcolor{black}{$\uparrow$2}} & 
\tcbox[colback=LightRed]{\textcolor{black}{$\downarrow$1}} & 
- & 
\tcbox[colback=LightGreen]{\textcolor{black}{$\uparrow$1}} & 
- & 
\tcbox[colback=LightRed]{\textcolor{black}{$\downarrow$2}} & 
\tcbox[colback=LightRed]{\textcolor{black}{$\downarrow$1}} & 
- & 
- & 
-  & 6 & 8  \\ \midrule
\multirow{2}{*}{\highemoji English} &
\scalesemoji &
- & 
- & 
- & 
- & 
- & 
\tcbox[colback=LightRed]{\textcolor{black}{$\downarrow$1}} & 
- & 
- & 
- & 
\tcbox[colback=LightGreen]{\textcolor{black}{$\uparrow$1}} & 
\tcbox[colback=LightGreen]{\textcolor{black}{$\uparrow$1}} & 
- & 
\tcbox[colback=LightRed]{\textcolor{black}{$\downarrow$1}} & 
-  & 4 & 4 \\ & 
\statuelibertyemoji & 
- & 
\tcbox[colback=LightGreen]{\textcolor{black}{$\uparrow$1}} & 
- & 
- & 
- & 
- & 
- & 
\tcbox[colback=LightGreen]{\textcolor{black}{$\uparrow$1}} & 
- & 
\tcbox[colback=LightRed]{\textcolor{black}{$\downarrow$1}} & 
\tcbox[colback=LightRed]{\textcolor{black}{$\downarrow$1}} & 
- & 
- & 
-  & 4 & 4 \\ \midrule
\multirow{2}{*}{\highemoji French} &
\scalesemoji &
- & 
\tcbox[colback=LightGreen]{\textcolor{black}{$\uparrow$1}} & 
- & 
- & 
- & 
- & 
- & 
- & 
- & 
\tcbox[colback=LightRed]{\textcolor{black}{$\downarrow$1}} & 
- & 
- & 
- & 
-   & 2 & 2 \\ & 
\statuelibertyemoji & 
- & 
\tcbox[colback=LightGreen]{\textcolor{black}{$\uparrow$2}} & 
\tcbox[colback=LightGreen]{\textcolor{black}{$\uparrow$2}} & 
\tcbox[colback=LightGreen]{\textcolor{black}{$\uparrow$1}} & 
- & 
\tcbox[colback=LightRed]{\textcolor{black}{$\downarrow$2}} & 
- & 
\tcbox[colback=LightGreen]{\textcolor{black}{$\uparrow$1}} & 
- & 
\tcbox[colback=LightRed]{\textcolor{black}{$\downarrow$3}} & 
\tcbox[colback=LightRed]{\textcolor{black}{$\downarrow$1}} & 
\tcbox[colback=LightGreen]{\textcolor{black}{$\uparrow$1}} & 
- & 
-  & 8 & 13  \\ \midrule
\multirow{2}{*}{\highemoji German} &
\scalesemoji &
- & 
\tcbox[colback=LightRed]{\textcolor{black}{$\downarrow$1}} & 
- & 
\tcbox[colback=LightRed]{\textcolor{black}{$\downarrow$1}} & 
- & 
\tcbox[colback=LightGreen]{\textcolor{black}{$\uparrow$1}} & 
- & 
- & 
- & 
\tcbox[colback=LightGreen]{\textcolor{black}{$\uparrow$1}} & 
- & 
- & 
- & 
-  & 4 & 4 \\ & 
\statuelibertyemoji & 
- & 
- & 
\tcbox[colback=LightRed]{\textcolor{black}{$\downarrow$1}} & 
- & 
\tcbox[colback=LightGreen]{\textcolor{black}{$\uparrow$2}} & 
- & 
- & 
\tcbox[colback=LightGreen]{\textcolor{black}{$\uparrow$1}} & 
- & 
\tcbox[colback=LightRed]{\textcolor{black}{$\downarrow$3}} & 
\tcbox[colback=LightRed]{\textcolor{black}{$\downarrow$1}} & 
\tcbox[colback=LightGreen]{\textcolor{black}{$\uparrow$2}} & 
- & 
-  & 6 & 10  \\ \midrule
\multirow{2}{*}{\highemoji Hindi} &
\scalesemoji &
- & 
\tcbox[colback=LightGreen]{\textcolor{black}{$\uparrow$1}} & 
\tcbox[colback=LightRed]{\textcolor{black}{$\downarrow$2}} & 
\tcbox[colback=LightRed]{\textcolor{black}{$\downarrow$1}} & 
\tcbox[colback=LightGreen]{\textcolor{black}{$\uparrow$1}} & 
- & 
- & 
- & 
- & 
- & 
- & 
\tcbox[colback=LightGreen]{\textcolor{black}{$\uparrow$1}} & 
- & 
-  & 5 & 6 \\ & 
\statuelibertyemoji & 
\tcbox[colback=LightGreen]{\textcolor{black}{$\uparrow$1}} & 
\tcbox[colback=LightRed]{\textcolor{black}{$\downarrow$1}} & 
\tcbox[colback=LightGreen]{\textcolor{black}{$\uparrow$1}} & 
\tcbox[colback=LightGreen]{\textcolor{black}{$\uparrow$2}} & 
- & 
\tcbox[colback=LightRed]{\textcolor{black}{$\downarrow$1}} & 
\tcbox[colback=LightGreen]{\textcolor{black}{$\uparrow$1}} & 
- & 
\tcbox[colback=LightGreen]{\textcolor{black}{$\uparrow$1}} & 
\tcbox[colback=LightRed]{\textcolor{black}{$\downarrow$3}} & 
\tcbox[colback=LightRed]{\textcolor{black}{$\downarrow$1}} & 
- & 
\tcbox[colback=LightGreen]{\textcolor{black}{$\uparrow$1}} & 
\tcbox[colback=LightRed]{\textcolor{black}{$\downarrow$1}}  & 11 & 14  \\ \midrule
\multirow{2}{*}{\highemoji Italian} &
\scalesemoji &
- & 
- & 
- & 
- & 
- & 
- & 
- & 
- & 
- & 
- & 
- & 
- & 
- & 
-  & 0 & 0 \\ & 
\statuelibertyemoji & 
- & 
- & 
\tcbox[colback=LightGreen]{\textcolor{black}{$\uparrow$1}} & 
\tcbox[colback=LightGreen]{\textcolor{black}{$\uparrow$1}} & 
- & 
\tcbox[colback=LightRed]{\textcolor{black}{$\downarrow$1}} & 
- & 
\tcbox[colback=LightGreen]{\textcolor{black}{$\uparrow$1}} & 
- & 
\tcbox[colback=LightRed]{\textcolor{black}{$\downarrow$2}} & 
\tcbox[colback=LightRed]{\textcolor{black}{$\downarrow$1}} & 
\tcbox[colback=LightGreen]{\textcolor{black}{$\uparrow$1}} & 
- & 
-  & 7 & 8 \\ \midrule
\multirow{2}{*}{\highemoji Japanese} &
\scalesemoji &
- & 
- & 
- & 
- & 
- & 
- & 
- & 
- & 
- & 
- & 
- & 
- & 
- & 
-  & 0 & 0 \\ & 
\statuelibertyemoji & 
- & 
\tcbox[colback=LightGreen]{\textcolor{black}{$\uparrow$1}} & 
\tcbox[colback=LightGreen]{\textcolor{black}{$\uparrow$1}} & 
\tcbox[colback=LightGreen]{\textcolor{black}{$\uparrow$1}} & 
\tcbox[colback=LightGreen]{\textcolor{black}{$\uparrow$1}} & 
\tcbox[colback=LightRed]{\textcolor{black}{$\downarrow$2}} & 
- & 
\tcbox[colback=LightGreen]{\textcolor{black}{$\uparrow$1}} & 
- & 
\tcbox[colback=LightRed]{\textcolor{black}{$\downarrow$1}} & 
\tcbox[colback=LightRed]{\textcolor{black}{$\downarrow$1}} & 
\tcbox[colback=LightRed]{\textcolor{black}{$\downarrow$1}} & 
- & 
-  & 9 & 10 \\ \midrule
\multirow{2}{*}{\highemoji Persian}
& \scalesemoji
& \tcbox[colback=LightGreen]{\textcolor{black}{$\uparrow$1}} & 
\tcbox[colback=LightGreen]{\textcolor{black}{$\uparrow$1}} & 
- & 
\tcbox[colback=LightRed]{\textcolor{black}{$\downarrow$1}} & 
- & 
- & 
\tcbox[colback=LightGreen]{\textcolor{black}{$\uparrow$1}} & 
- & 
\tcbox[colback=LightRed]{\textcolor{black}{$\downarrow$2}} & 
- & 
- & 
- & 
- & 
- & 5 & 6 \\ 
& \statuelibertyemoji
& - & 
- & 
- & \tcbox[colback=LightGreen]{\textcolor{black}{$\uparrow$2}} & \tcbox[colback=LightGreen]{\textcolor{black}{$\uparrow$1}} & \tcbox[colback=LightRed]{\textcolor{black}{$\downarrow$2}} & 
- & 
- & 
\tcbox[colback=LightGreen]{\textcolor{black}{$\uparrow$1}} & 
\tcbox[colback=LightGreen]{\textcolor{black}{$\uparrow$1}} & 
\tcbox[colback=LightRed]{\textcolor{black}{$\downarrow$1}} & 
\tcbox[colback=LightGreen]{\textcolor{black}{$\uparrow$1}} & 
- & 
- & 7 & 9 \\ \midrule
\multirow{2}{*}{\highemoji Polish}
& \scalesemoji
& \tcbox[colback=LightGreen]{\textcolor{black}{$\uparrow$2}} & 
\tcbox[colback=LightGreen]{\textcolor{black}{$\uparrow$1}} & 
\tcbox[colback=LightGreen]{\textcolor{black}{$\uparrow$2}} & 
\tcbox[colback=LightRed]{\textcolor{black}{$\downarrow$1}} & 
\tcbox[colback=LightRed]{\textcolor{black}{$\downarrow$1}} & 
- & 
\tcbox[colback=LightRed]{\textcolor{black}{$\downarrow$1}} & 
- & 
\tcbox[colback=LightRed]{\textcolor{black}{$\downarrow$1}} & 
\tcbox[colback=LightGreen]{\textcolor{black}{$\uparrow$2}} & 
- & 
\tcbox[colback=LightRed]{\textcolor{black}{$\downarrow$1}} & 
- & 
- & 9 & 12 \\ 
& \statuelibertyemoji
& - &  - & \tcbox[colback=LightGreen]{\textcolor{black}{$\uparrow$2}} &  \tcbox[colback=LightGreen]{\textcolor{black}{$\uparrow$2}} & - & \tcbox[colback=LightRed]{\textcolor{black}{$\downarrow$1}} & 
- & \tcbox[colback=LightGreen]{\textcolor{black}{$\uparrow$1}} & \tcbox[colback=LightGreen]{\textcolor{black}{$\uparrow$1}} & \tcbox[colback=LightRed]{\textcolor{black}{$\downarrow$1}} & 
\tcbox[colback=LightRed]{\textcolor{black}{$\downarrow$1}} & - & 
- & 
- & 7 & 9 \\ \midrule
\multirow{2}{*}{\highemoji Portuguese} &
\scalesemoji &
- & 
- & 
- & 
- & 
- & 
- & 
- & 
- & 
- & 
- & 
- & 
- & 
- & 
- & 0 & 0  \\ & 
\statuelibertyemoji & 
- & 
\tcbox[colback=LightGreen]{\textcolor{black}{$\uparrow$1}} & 
\tcbox[colback=LightGreen]{\textcolor{black}{$\uparrow$1}} & 
\tcbox[colback=LightGreen]{\textcolor{black}{$\uparrow$1}} & 
\tcbox[colback=LightGreen]{\textcolor{black}{$\uparrow$1}} & 
\tcbox[colback=LightRed]{\textcolor{black}{$\downarrow$1}} & 
- & 
\tcbox[colback=LightGreen]{\textcolor{black}{$\uparrow$1}} & 
- & 
\tcbox[colback=LightRed]{\textcolor{black}{$\downarrow$2}} & 
\tcbox[colback=LightRed]{\textcolor{black}{$\downarrow$1}} & 
\tcbox[colback=LightRed]{\textcolor{black}{$\downarrow$1}} & 
- & 
- & 9 & 10 \\ \midrule
\multirow{2}{*}{\highemoji Russian}
& \scalesemoji
& - & 
\tcbox[colback=LightRed]{\textcolor{black}{$\downarrow$1}} & 
\tcbox[colback=LightRed]{\textcolor{black}{$\downarrow$1}} & 
\tcbox[colback=LightRed]{\textcolor{black}{$\downarrow$1}} & 
\tcbox[colback=LightGreen]{\textcolor{black}{$\uparrow$1}} & - & - & 
- & 
- & 
\tcbox[colback=LightGreen]{\textcolor{black}{$\uparrow$2}} & 
- & 
- & 
- & 
- & 5 & 6 \\ 
& \statuelibertyemoji
& \tcbox[colback=LightGreen]{\textcolor{black}{$\uparrow$1}} & 
- & 
- & 
\tcbox[colback=LightGreen]{\textcolor{black}{$\uparrow$2}} & 
\tcbox[colback=LightRed]{\textcolor{black}{$\downarrow$1}} & 
\tcbox[colback=LightRed]{\textcolor{black}{$\downarrow$1}} & 
\tcbox[colback=LightGreen]{\textcolor{black}{$\uparrow$1}}  & 
- & 
- & 
\tcbox[colback=LightRed]{\textcolor{black}{$\downarrow$2}} & 
\tcbox[colback=LightRed]{\textcolor{black}{$\downarrow$1}} & 
\tcbox[colback=LightGreen]{\textcolor{black}{$\uparrow$3}} & 
- & 
- & 8 & 12 \\ \midrule
\multirow{2}{*}{\highemoji Serbian}
& \scalesemoji
& - & 
\tcbox[colback=LightRed]{\textcolor{black}{$\downarrow$1}} & 
- & 
\tcbox[colback=LightGreen]{\textcolor{black}{$\uparrow$1}} & 
- & 
- & 
- & 
\tcbox[colback=LightRed]{\textcolor{black}{$\downarrow$1}} & 
- & 
\tcbox[colback=LightRed]{\textcolor{black}{$\downarrow$1}} & 
\tcbox[colback=LightGreen]{\textcolor{black}{$\uparrow$1}} & 
- & 
- & 
- & 5 & 5 \\ 
& \statuelibertyemoji
& - & 
\tcbox[colback=LightGreen]{\textcolor{black}{$\uparrow$2}} & 
\tcbox[colback=LightGreen]{\textcolor{black}{$\uparrow$1}} & 
\tcbox[colback=LightRed]{\textcolor{black}{$\downarrow$1}} & 
\tcbox[colback=LightGreen]{\textcolor{black}{$\uparrow$1}} & 
- & 
- & 
- & 
- & 
- & 
- & 
- & 
- & 
- & 4 & 5 \\ \midrule
\multirow{2}{*}{\highemoji Spanish} &
\scalesemoji &

- & 
\tcbox[colback=LightRed]{\textcolor{black}{$\downarrow$1}} & 
- & 
\tcbox[colback=LightRed]{\textcolor{black}{$\downarrow$1}} & 
- & 
\tcbox[colback=LightGreen]{\textcolor{black}{$\uparrow$1}} & 
- & 
- & 
- & 
\tcbox[colback=LightGreen]{\textcolor{black}{$\uparrow$1}} & 
- & 
- & 
- & 
- & 4 & 4 \\ & 
\statuelibertyemoji & 
- & 
- & 
\tcbox[colback=LightGreen]{\textcolor{black}{$\uparrow$1}} & 
- & 
\tcbox[colback=LightGreen]{\textcolor{black}{$\uparrow$2}} & 
- & 
- & 
\tcbox[colback=LightGreen]{\textcolor{black}{$\uparrow$1}} & 
- & 
\tcbox[colback=LightRed]{\textcolor{black}{$\downarrow$3}} & 
\tcbox[colback=LightRed]{\textcolor{black}{$\downarrow$1}} & 
- & 
- & 
- & 5 & 8 \\ \midrule
\multirow{2}{*}{\highemoji Swedish}
& \scalesemoji
& - & \tcbox[colback=LightRed]{\textcolor{black}{$\downarrow$1}} & - &  \tcbox[colback=LightRed]{\textcolor{black}{$\downarrow$1}} & - & \tcbox[colback=LightGreen]{\textcolor{black}{$\uparrow$1}} & 
- & - & 
- & \tcbox[colback=LightRed]{\textcolor{black}{$\downarrow$1}} & - & 
- & 
- & 
- & 4 & 4 \\ 
& \statuelibertyemoji
& - & - & \tcbox[colback=LightGreen]{\textcolor{black}{$\uparrow$1}} & - & \tcbox[colback=LightGreen]{\textcolor{black}{$\uparrow$2}} & 
- & 
- & 
\tcbox[colback=LightGreen]{\textcolor{black}{$\uparrow$1}} & 
- & 
\tcbox[colback=LightRed]{\textcolor{black}{$\downarrow$3}} & 
\tcbox[colback=LightRed]{\textcolor{black}{$\downarrow$1}} & 
- & 
- & 
- & 5 & 8 \\ \midrule
\multirow{2}{*}{\highemoji Turkish}
& \scalesemoji
& - & 
-& 
- & 
\tcbox[colback=LightRed]{\textcolor{black}{$\downarrow$1}} & 
\tcbox[colback=LightGreen]{\textcolor{black}{$\uparrow$1}} & 
- & 
\tcbox[colback=LightRed]{\textcolor{black}{$\downarrow$1}} & 
- & 
- & 
\tcbox[colback=LightGreen]{\textcolor{black}{$\uparrow$1}} & 
- & 
- & 
- & & 4 & 4 \\ 
& \statuelibertyemoji
& - & 
\tcbox[colback=LightGreen]{\textcolor{black}{$\uparrow$2}} & 
\tcbox[colback=LightRed]{\textcolor{black}{$\downarrow$1}} & 
\tcbox[colback=LightGreen]{\textcolor{black}{$\uparrow$1}} & 
- & 
\tcbox[colback=LightRed]{\textcolor{black}{$\downarrow$1}} & 
- & 
- & 
- & 
- & 
\tcbox[colback=LightRed]{\textcolor{black}{$\downarrow$2}} & 
- & 
- & 
- & 5 & 7 \\ \midrule
\multirow{2}{*}{\highemoji Vietnamese}
& \scalesemoji
& - & - & - & \tcbox[colback=LightRed]{\textcolor{black}{$\downarrow$1}} & - & \tcbox[colback=LightGreen]{\textcolor{black}{$\uparrow$1}}  & \tcbox[colback=LightRed]{\textcolor{black}{$\downarrow$1}} & - &  -\tcbox[colback=LightGreen]{\textcolor{black}{$\uparrow$1}} & - & - & - & - & -  & 4 & 4 \\
& \statuelibertyemoji
& - & \tcbox[colback=LightRed]{\textcolor{black}{$\downarrow$1}} & 
\tcbox[colback=LightGreen]{\textcolor{black}{$\uparrow$3}} & - & - & \tcbox[colback=LightRed]{\textcolor{black}{$\downarrow$1}} & \tcbox[colback=LightRed]{\textcolor{black}{$\downarrow$1}} & - & \tcbox[colback=LightGreen]{\textcolor{black}{$\uparrow$1}} & \tcbox[colback=LightRed]{\textcolor{black}{$\downarrow$1}} & - & - & - & -  & 6 & 8 \\
\bottomrule
\end{tabular}}
\caption{Model rankings with MA \memoemoji rank as the reference for high-resource languages (\highemoji). First row indicates changes in CA \scalesemoji{} ranks, while second row shows the changes in CS \statuelibertyemoji{} ranks relative to MA. Color-coded boxes highlight increases (\tcbox[colback=LightGreen]{\textcolor{black}{$\uparrow$}}) and decreases (\tcbox[colback=LightRed]{\textcolor{black}{$\downarrow$}}).}
\label{tab:model_rank_changes_high_res}
\end{table}
\FloatBarrier

\begin{table}[ht!]
\centering
\scalebox{0.60}{
\begin{tabular}{l|c|cccccccccccccc|cc}
\toprule
\textbf{Language} &  \textbf{Dataset}  & \rotatebox{90}{Aya Exp. 8B} & \rotatebox{90}{Aya Exp. 32B} & \rotatebox{90}{CommandR} & \rotatebox{90}{CommandR+} & \rotatebox{90}{Gemma2 9B} & \rotatebox{90}{Gemma2 27B} & \rotatebox{90}{Llama-3.1 8B} & \rotatebox{90}{Llama-3.1 70B} & \rotatebox{90}{Mistral Nemo} & \rotatebox{90}{Qwen2.5 7B} & \rotatebox{90}{Qwen2.5 32B} & \rotatebox{90}{SEA-LION-v3} & \rotatebox{90}{GPT4o} & \rotatebox{90}{Claude Sonnet} & \rotatebox{90}{\textbf{Total rank change}} & \rotatebox{90}{\textbf{Total position change}} \\
\midrule
\multirow{2}{*}{\midemoji  Bengali}
& \scalesemoji & - & \tcbox[colback=LightGreen]{\textcolor{black}{$\uparrow$1}} & - & - & - & - & - & \tcbox[colback=LightRed]{\textcolor{black}{$\downarrow$1}} & \tcbox[colback=LightRed]{\textcolor{black}{$\downarrow$1}} & - & - & - & - & - & 3 & 3 \\ 
& \statuelibertyemoji & - & - & - & - & - & - & - & - & \tcbox[colback=LightGreen]{\textcolor{black}{$\uparrow$1}} & \tcbox[colback=LightRed]{\textcolor{black}{$\downarrow$1}} & - & - & - & - & 2 & 2  \\ 
\midrule
\multirow{2}{*}{\midemoji  Filipino}
& \scalesemoji & - & - & - & - & - & - & - & - & - & - & - & - & - & - & 0 & 0 \\ 
& \statuelibertyemoji & - & - & - & - & - & \tcbox[colback=LightGreen]{\textcolor{black}{$\uparrow$1}} & \tcbox[colback=LightGreen]{\textcolor{black}{$\uparrow$1}} & - & \tcbox[colback=LightRed]{\textcolor{black}{$\downarrow$1}} & - & \tcbox[colback=LightRed]{\textcolor{black}{$\downarrow$1}} & - & \tcbox[colback=LightRed]{\textcolor{black}{$\downarrow$1}} & \tcbox[colback=LightGreen]{\textcolor{black}{$\uparrow$1}} & 6 & 6 \\ 
\midrule
\multirow{2}{*}{\midemoji  Greek} &
\scalesemoji &
\tcbox[colback=LightRed]{\textcolor{black}{$\downarrow$1}} & 
\tcbox[colback=LightRed]{\textcolor{black}{$\downarrow$1}} & 
- & 
- & 
- & 
\tcbox[colback=LightGreen]{\textcolor{black}{$\uparrow$1}} & 
- & 
- & 
\tcbox[colback=LightRed]{\textcolor{black}{$\downarrow$1}} & 
\tcbox[colback=LightGreen]{\textcolor{black}{$\uparrow$2}} & 
- & 
- & 
- & 
- & 5  &  6 \\ & 
\statuelibertyemoji & 
- & 
- & 
\tcbox[colback=LightGreen]{\textcolor{black}{$\uparrow$2}} & 
\tcbox[colback=LightGreen]{\textcolor{black}{$\uparrow$3}} & 
- & 
\tcbox[colback=LightRed]{\textcolor{black}{$\downarrow$1}} & 
\tcbox[colback=LightGreen]{\textcolor{black}{$\uparrow$1}} & 
- & 
- & 
\tcbox[colback=LightRed]{\textcolor{black}{$\downarrow$1}} & 
\tcbox[colback=LightRed]{\textcolor{black}{$\downarrow$4}} & 
- & 
- & 
- & 6 & 12 \\ \midrule
\multirow{2}{*}{\midemoji  Hebrew}
& \scalesemoji 
& \tcbox[colback=LightRed]{\textcolor{black}{$\downarrow$1}} & \tcbox[colback=LightGreen]{\textcolor{black}{$\uparrow$1}} & - & \tcbox[colback=LightRed]{\textcolor{black}{$\downarrow$1}} & - & - & - & -  & \tcbox[colback=LightGreen]{\textcolor{black}{$\uparrow$1}} & - & -  & - & - & - & 4 &  4 \\ 
& \statuelibertyemoji & - & \tcbox[colback=LightGreen]{\textcolor{black}{$\uparrow$2}} & - & \tcbox[colback=LightGreen]{\textcolor{black}{$\uparrow$2}} & - & \tcbox[colback=LightRed]{\textcolor{black}{$\downarrow$2}} & - & - & - & - & \tcbox[colback=LightRed]{\textcolor{black}{$\downarrow$2}} & - & - & - & 4 & 8 \\ 
\midrule
\multirow{2}{*}{\midemoji  Indonesian} &
\scalesemoji &
- & 
- & 
\tcbox[colback=LightRed]{\textcolor{black}{$\downarrow$1}} & 
\tcbox[colback=LightRed]{\textcolor{black}{$\downarrow$1}} & 
\tcbox[colback=LightRed]{\textcolor{black}{$\downarrow$1}} & 
\tcbox[colback=LightGreen]{\textcolor{black}{$\uparrow$1}} & 
- & 
- & 
- & 
\tcbox[colback=LightGreen]{\textcolor{black}{$\uparrow$2}} & 
- & 
- & 
- & 
- & 5 & 6 \\ & 
\statuelibertyemoji & 
- & 
- & 
\tcbox[colback=LightGreen]{\textcolor{black}{$\uparrow$1}} & 
- & 
- & 
- & 
\tcbox[colback=LightRed]{\textcolor{black}{$\downarrow$1}} & 
\tcbox[colback=LightGreen]{\textcolor{black}{$\uparrow$1}} & 
\tcbox[colback=LightGreen]{\textcolor{black}{$\uparrow$1}} & 
- & 
\tcbox[colback=LightRed]{\textcolor{black}{$\downarrow$1}} & 
\tcbox[colback=LightRed]{\textcolor{black}{$\downarrow$1}} & 
- & 
- & 6 & 6 \\ \midrule
\multirow{2}{*}{\midemoji  Korean} &
\scalesemoji &
\tcbox[colback=LightRed]{\textcolor{black}{$\downarrow$1}} & 
\tcbox[colback=LightRed]{\textcolor{black}{$\downarrow$1}} & 
\tcbox[colback=LightRed]{\textcolor{black}{$\downarrow$1}} & 
- & 
- & 
\tcbox[colback=LightGreen]{\textcolor{black}{$\uparrow$1}} & 
\tcbox[colback=LightGreen]{\textcolor{black}{$\uparrow$1}} & 
- & 
- & 
\tcbox[colback=LightGreen]{\textcolor{black}{$\uparrow$1}} & 
- & 
- & 
- & 
- & 6 & 6 \\ & 
\statuelibertyemoji & 
- & 
\tcbox[colback=LightGreen]{\textcolor{black}{$\uparrow$1}} & 
\tcbox[colback=LightGreen]{\textcolor{black}{$\uparrow$1}} & 
\tcbox[colback=LightRed]{\textcolor{black}{$\downarrow$1}} & 
- & 
\tcbox[colback=LightRed]{\textcolor{black}{$\downarrow$1}} & 
- & 
\tcbox[colback=LightGreen]{\textcolor{black}{$\uparrow$1}} & 
- & 
- & 
\tcbox[colback=LightRed]{\textcolor{black}{$\downarrow$1}} & 
- & 
- & 
-  & 6 & 6 \\ \midrule
\multirow{2}{*}{\midemoji  Malay}
& \scalesemoji 
& - & 
- & 
- & 
- & 
\tcbox[colback=LightRed]{\textcolor{black}{$\downarrow$1}} & 
- & 
- & 
\tcbox[colback=LightRed]{\textcolor{black}{$\downarrow$1}} & 
- & 
\tcbox[colback=LightGreen]{\textcolor{black}{$\uparrow$1}} & 
\tcbox[colback=LightGreen]{\textcolor{black}{$\uparrow$1}} & 
- & 
- & 
- & 4 & 4 \\ 
& \statuelibertyemoji
& - & 
\tcbox[colback=LightGreen]{\textcolor{black}{$\uparrow$1}} & 
\tcbox[colback=LightGreen]{\textcolor{black}{$\uparrow$1}} & 
\tcbox[colback=LightRed]{\textcolor{black}{$\downarrow$1}} & 
- & 
- & 
- & 
- & 
- & 
\tcbox[colback=LightRed]{\textcolor{black}{$\downarrow$1}} & 
- & 
- & 
- & 
- & 4 & 4 \\ \midrule
\multirow{2}{*}{\midemoji  Lithuanian}
& \scalesemoji 
& - & 
- & 
- & 
- & 
- & 
- & 
- & 
- & 
- & 
- & 
- & 
- & 
- & 
- & 0 & 0 \\ 
& \statuelibertyemoji
& - & 
- & 
- & 
\tcbox[colback=LightGreen]{\textcolor{black}{$\uparrow$2}} & 
- & 
- & 
- & 
- & 
- & 
- & 
- & 
\tcbox[colback=LightRed]{\textcolor{black}{$\downarrow$2}} & 
- & 
- & 2 & 4 \\ \midrule
\multirow{2}{*}{\midemoji  Romanian} &
\scalesemoji &
- & 
\tcbox[colback=LightGreen]{\textcolor{black}{$\uparrow$1}} & 
- & 
\tcbox[colback=LightRed]{\textcolor{black}{$\downarrow$1}} & 
- & 
- & 
\tcbox[colback=LightGreen]{\textcolor{black}{$\uparrow$1}} & 
\tcbox[colback=LightRed]{\textcolor{black}{$\downarrow$1}} & 
\tcbox[colback=LightRed]{\textcolor{black}{$\downarrow$1}} & 
- & 
\tcbox[colback=LightGreen]{\textcolor{black}{$\uparrow$1}} & 
- & 
- & 
- & 6 &  6 \\ & 
\statuelibertyemoji & 
- & 
- & 
- & 
\tcbox[colback=LightGreen]{\textcolor{black}{$\uparrow$2}} & 
- & 
\tcbox[colback=LightRed]{\textcolor{black}{$\downarrow$1}} & 
- & 
- & 
- & 
- & 
- & 
- & 
- & 
- & 2 & 3 \\ \midrule
\multirow{2}{*}{\midemoji  Ukrainian} &
\scalesemoji &
- & 
\tcbox[colback=LightGreen]{\textcolor{black}{$\uparrow$1}} & 
- & 
\tcbox[colback=LightRed]{\textcolor{black}{$\downarrow$1}} & 
\tcbox[colback=LightRed]{\textcolor{black}{$\downarrow$1}} & 
- & 
- & 
- & 
- & 
\tcbox[colback=LightGreen]{\textcolor{black}{$\uparrow$1}} & 
- & 
- & 
- & 
- & 4 & 4 \\ & 
\statuelibertyemoji & 
- & 
\tcbox[colback=LightGreen]{\textcolor{black}{$\uparrow$1}} & 
- & 
\tcbox[colback=LightGreen]{\textcolor{black}{$\uparrow$1}} & 
- & 
\tcbox[colback=LightRed]{\textcolor{black}{$\downarrow$2}} & 
- & 
\tcbox[colback=LightGreen]{\textcolor{black}{$\uparrow$1}} & 
\tcbox[colback=LightGreen]{\textcolor{black}{$\uparrow$1}} & 
\tcbox[colback=LightRed]{\textcolor{black}{$\downarrow$1}} & 
\tcbox[colback=LightRed]{\textcolor{black}{$\downarrow$1}} & 
- & 
\tcbox[colback=LightGreen]{\textcolor{black}{$\uparrow$1}} & 
\tcbox[colback=LightRed]{\textcolor{black}{$\downarrow$1}} & 9 & 10 \\ \midrule
\midrule 
\multirow{2}{*}{\lowemoji Amharic} &
\scalesemoji &
- & 
- & 
\tcbox[colback=LightRed]{\textcolor{black}{$\downarrow$1}} & 
\tcbox[colback=LightGreen]{\textcolor{black}{$\uparrow$1}} & 
\tcbox[colback=LightRed]{\textcolor{black}{$\downarrow$1}} & 
- & 
- & 
- & 
- & 
- & 
\tcbox[colback=LightGreen]{\textcolor{black}{$\uparrow$1}} & 
- & 
- & 
- & 4 & 4 \\ & 
\statuelibertyemoji & 
\tcbox[colback=LightRed]{\textcolor{black}{$\downarrow$1}} & 
\tcbox[colback=LightGreen]{\textcolor{black}{$\uparrow$2}} & 
\tcbox[colback=LightGreen]{\textcolor{black}{$\uparrow$2}} & 
\tcbox[colback=LightRed]{\textcolor{black}{$\downarrow$1}} & 
- & 
- & 
- & 
- & 
\tcbox[colback=LightGreen]{\textcolor{black}{$\uparrow$1}} & 
\tcbox[colback=LightRed]{\textcolor{black}{$\downarrow$3}} & 
- & 
- & 
- & 
- & 6 & 10 \\ \midrule
\multirow{2}{*}{\lowemoji Hausa} &
\scalesemoji &
- & 
- & 
- & 
- & 
- & 
- & 
- & 
- & 
- & 
- & 
- & 
- & 
- & 
- & 0 &  0 \\ & 
\statuelibertyemoji & 
\tcbox[colback=LightGreen]{\textcolor{black}{$\uparrow$1}} & 
\tcbox[colback=LightRed]{\textcolor{black}{$\downarrow$1}} & 
\tcbox[colback=LightGreen]{\textcolor{black}{$\uparrow$3}} & 
\tcbox[colback=LightRed]{\textcolor{black}{$\downarrow$1}} & 
- & 
- & 
\tcbox[colback=LightRed]{\textcolor{black}{$\downarrow$1}} & 
- & 
\tcbox[colback=LightRed]{\textcolor{black}{$\downarrow$1}} & 
\tcbox[colback=LightRed]{\textcolor{black}{$\downarrow$1}} & 
\tcbox[colback=LightGreen]{\textcolor{black}{$\uparrow$1}} & 
- & 
- & 
- & 8 & 10 \\ \midrule
\multirow{2}{*}{\lowemoji Igbo} &
\scalesemoji &
- & 
- & 
\tcbox[colback=LightRed]{\textcolor{black}{$\downarrow$1}} & 
- & 
- & 
- & 
\tcbox[colback=LightRed]{\textcolor{black}{$\downarrow$1}} & 
- & 
- & 
\tcbox[colback=LightGreen]{\textcolor{black}{$\uparrow$1}} & 
\tcbox[colback=LightGreen]{\textcolor{black}{$\uparrow$1}} & 
- & 
- & 
-  & 4 & 4 \\ & 
\statuelibertyemoji & 
- & 
\tcbox[colback=LightGreen]{\textcolor{black}{$\uparrow$1}} & 
- & 
- & 
\tcbox[colback=LightGreen]{\textcolor{black}{$\uparrow$1}} & 
- & 
- & 
- & 
\tcbox[colback=LightGreen]{\textcolor{black}{$\uparrow$2}} & 
\tcbox[colback=LightRed]{\textcolor{black}{$\downarrow$3}} & 
- & 
\tcbox[colback=LightRed]{\textcolor{black}{$\downarrow$1}} & 
- & 
- & 5 & 8 \\ \midrule
\multirow{2}{*}{\lowemoji Kyrgyz} &
\scalesemoji &
- & 
- & 
- & 
- & 
- & 
\tcbox[colback=LightRed]{\textcolor{black}{$\downarrow$1}} & 
- & 
- & 
- & 
- & 
\tcbox[colback=LightGreen]{\textcolor{black}{$\uparrow$1}} & 
- & 
- & 
- & 2 & 2 \\ & 
\statuelibertyemoji & 
- & 
\tcbox[colback=LightRed]{\textcolor{black}{$\downarrow$1}} & 
\tcbox[colback=LightGreen]{\textcolor{black}{$\uparrow$1}} & 
\tcbox[colback=LightGreen]{\textcolor{black}{$\uparrow$1}} & 
- & 
- & 
\tcbox[colback=LightGreen]{\textcolor{black}{$\uparrow$1}} & 
- & 
- & 
\tcbox[colback=LightRed]{\textcolor{black}{$\downarrow$2}} & 
- & 
- & 
- & 
- & 5 & 6 \\ \midrule
\multirow{2}{*}{\lowemoji Malagasy} &
\scalesemoji &
- & 
\tcbox[colback=LightRed]{\textcolor{black}{$\downarrow$1}} & 
- & 
- & 
- & 
- & 
- & 
- & 
- & 
\tcbox[colback=LightGreen]{\textcolor{black}{$\uparrow$1}} & 
- & 
- & 
- & 
- & 2 & 2 \\ & 
\statuelibertyemoji & 
- & 
\tcbox[colback=LightGreen]{\textcolor{black}{$\uparrow$1}} & 
\tcbox[colback=LightGreen]{\textcolor{black}{$\uparrow$4}} & 
\tcbox[colback=LightGreen]{\textcolor{black}{$\uparrow$1}} & 
- & 
- & 
\tcbox[colback=LightRed]{\textcolor{black}{$\downarrow$1}} & 
- & 
\tcbox[colback=LightGreen]{\textcolor{black}{$\uparrow$1}} & 
\tcbox[colback=LightRed]{\textcolor{black}{$\downarrow$1}} & 
\tcbox[colback=LightRed]{\textcolor{black}{$\downarrow$5}} & 
- & 
- & 
- & 7 & 14 \\ \midrule
\multirow{2}{*}{\lowemoji Nepali} &
\scalesemoji &
- & 
- & 
- & 
- & 
- & 
- & 
- & 
- & 
\tcbox[colback=LightRed]{\textcolor{black}{$\downarrow$1}} & 
\tcbox[colback=LightGreen]{\textcolor{black}{$\uparrow$1}} & 
- & 
- & 
- & 
-  & 2 & 2 \\ & 
\statuelibertyemoji & 
- & 
- & 
- & 
- & 
- & 
\tcbox[colback=LightGreen]{\textcolor{black}{$\uparrow$1}} & 
\tcbox[colback=LightRed]{\textcolor{black}{$\downarrow$1}} & 
- & 
\tcbox[colback=LightGreen]{\textcolor{black}{$\uparrow$1}} & 
- & 
\tcbox[colback=LightRed]{\textcolor{black}{$\downarrow$1}} & 
- & 
\tcbox[colback=LightGreen]{\textcolor{black}{$\uparrow$1}} & 
\tcbox[colback=LightRed]{\textcolor{black}{$\downarrow$1}}  & 6 & 6 \\ \midrule
\multirow{2}{*}{\lowemoji Nyanja} &
\scalesemoji &
- & 
- & 
- & 
\tcbox[colback=LightRed]{\textcolor{black}{$\downarrow$1}} & 
\tcbox[colback=LightRed]{\textcolor{black}{$\downarrow$1}} & 
- & 
- & 
- & 
- & 
- & 
\tcbox[colback=LightGreen]{\textcolor{black}{$\uparrow$2}} & 
- & 
\tcbox[colback=LightGreen]{\textcolor{black}{$\uparrow$1}} & 
\tcbox[colback=LightRed]{\textcolor{black}{$\downarrow$1}} & 5 & 6 \\ & 
\statuelibertyemoji & 
- & 
\tcbox[colback=LightRed]{\textcolor{black}{$\downarrow$1}} & 
\tcbox[colback=LightGreen]{\textcolor{black}{$\uparrow$1}} & 
- & 
- & 
- & 
- & 
- & 
- & 
- & 
- & 
- & 
- & 
- & 2 & 2 \\ \midrule
\multirow{2}{*}{\lowemoji Shona} &
\scalesemoji &
- & 
- & 
- & 
- & 
\tcbox[colback=LightRed]{\textcolor{black}{$\downarrow$1}} & 
- & 
- & 
- & 
- & 
- & 
\tcbox[colback=LightGreen]{\textcolor{black}{$\uparrow$1}} & 
- & 
\tcbox[colback=LightRed]{\textcolor{black}{$\downarrow$1}} & 
\tcbox[colback=LightGreen]{\textcolor{black}{$\uparrow$1}} & 4 & 4  \\ & 
\statuelibertyemoji & 
\tcbox[colback=LightGreen]{\textcolor{black}{$\uparrow$2}} & 
- & 
\tcbox[colback=LightGreen]{\textcolor{black}{$\uparrow$1}} & 
\tcbox[colback=LightGreen]{\textcolor{black}{$\uparrow$1}} & 
- & 
- & 
\tcbox[colback=LightGreen]{\textcolor{black}{$\uparrow$1}} & 
- & 
- & 
\tcbox[colback=LightRed]{\textcolor{black}{$\downarrow$4}} & 
\tcbox[colback=LightRed]{\textcolor{black}{$\downarrow$1}} & 
- & 
- & 
- & 6 & 10 \\ \midrule
\multirow{2}{*}{\lowemoji Sinhala} &
\scalesemoji &
- & 
\tcbox[colback=LightGreen]{\textcolor{black}{$\uparrow$1}} & 
- & 
- & 
- & 
- & 
\tcbox[colback=LightRed]{\textcolor{black}{$\downarrow$3}} & 
- & 
- & 
\tcbox[colback=LightGreen]{\textcolor{black}{$\uparrow$2}} & 
- & 
- & 
- & 
- & 3 & 6 \\ & 
\statuelibertyemoji & 
- & 
\tcbox[colback=LightRed]{\textcolor{black}{$\downarrow$1}} & 
\tcbox[colback=LightGreen]{\textcolor{black}{$\uparrow$1}} & 
\tcbox[colback=LightGreen]{\textcolor{black}{$\uparrow$1}} & 
- & 
- & 
- & 
- & 
- & 
\tcbox[colback=LightRed]{\textcolor{black}{$\downarrow$1}} & 
- & 
- & 
- & 
- & 4 & 4 \\ \midrule
\multirow{2}{*}{\lowemoji Somali} &
\scalesemoji &
- & 
\tcbox[colback=LightRed]{\textcolor{black}{$\downarrow$2}} & 
- & 
\tcbox[colback=LightGreen]{\textcolor{black}{$\uparrow$1}} & 
- & 
- & 
- & 
- & 
- & 
\tcbox[colback=LightGreen]{\textcolor{black}{$\uparrow$1}} & 
- & 
- & 
- & 
- & 3 & 4 \\ & 
\statuelibertyemoji & 
- & 
\tcbox[colback=LightGreen]{\textcolor{black}{$\uparrow$1}} & 
\tcbox[colback=LightGreen]{\textcolor{black}{$\uparrow$2}} & 
\tcbox[colback=LightRed]{\textcolor{black}{$\downarrow$2}} & 
- & 
- & 
\tcbox[colback=LightGreen]{\textcolor{black}{$\uparrow$2}} & 
- & 
- & 
\tcbox[colback=LightRed]{\textcolor{black}{$\downarrow$2}} & 
\tcbox[colback=LightRed]{\textcolor{black}{$\downarrow$1}} & 
- & 
\tcbox[colback=LightRed]{\textcolor{black}{$\downarrow$1}} & 
\tcbox[colback=LightGreen]{\textcolor{black}{$\uparrow$1}} & 8 & 12 \\ \midrule
\multirow{2}{*}{\lowemoji Swahili} &
\scalesemoji &
- & 
\tcbox[colback=LightRed]{\textcolor{black}{$\downarrow$1}} & 
- & 
- & 
- & 
- & 
\tcbox[colback=LightGreen]{\textcolor{black}{$\uparrow$1}} & 
- & 
- & 
- & 
- & 
- & 
- & 
- & 2 & 2 \\ & 
\statuelibertyemoji & 
- & 
- & 
\tcbox[colback=LightGreen]{\textcolor{black}{$\uparrow$1}} & 
- & 
- & 
- & 
\tcbox[colback=LightRed]{\textcolor{black}{$\downarrow$1}} & 
- & 
- & 
- & 
- & 
- & 
\tcbox[colback=LightRed]{\textcolor{black}{$\downarrow$1}} & 
\tcbox[colback=LightGreen]{\textcolor{black}{$\uparrow$1}}  & 4 & 4 \\ \midrule
\multirow{2}{*}{\lowemoji Telugu} &
\scalesemoji &
- & 
\tcbox[colback=LightRed]{\textcolor{black}{$\downarrow$1}} & 
- & 
- & 
- & 
- & 
- & 
- & 
- & 
\tcbox[colback=LightGreen]{\textcolor{black}{$\uparrow$1}} & 
\tcbox[colback=LightGreen]{\textcolor{black}{$\uparrow$1}} & 
\tcbox[colback=LightRed]{\textcolor{black}{$\downarrow$1}} & 
- & 
- & 4 & 4 \\ & 
\statuelibertyemoji & 
- & 
\tcbox[colback=LightRed]{\textcolor{black}{$\downarrow$1}} & 
\tcbox[colback=LightGreen]{\textcolor{black}{$\uparrow$2}} & 
\tcbox[colback=LightGreen]{\textcolor{black}{$\uparrow$1}} & 
\tcbox[colback=LightGreen]{\textcolor{black}{$\uparrow$1}} & 
- & 
\tcbox[colback=LightGreen]{\textcolor{black}{$\uparrow$1}} & 
- & 
\tcbox[colback=LightRed]{\textcolor{black}{$\downarrow$1}} & 
\tcbox[colback=LightRed]{\textcolor{black}{$\downarrow$2}} & 
\tcbox[colback=LightRed]{\textcolor{black}{$\downarrow$1}} & 
- & 
- & 
- & 8 & 10 \\ \midrule
\multirow{2}{*}{\lowemoji Yoruba} &
\scalesemoji &
- & 
\tcbox[colback=LightGreen]{\textcolor{black}{$\uparrow$1}} & 
\tcbox[colback=LightRed]{\textcolor{black}{$\downarrow$2}} & 
- & 
\tcbox[colback=LightRed]{\textcolor{black}{$\downarrow$1}} & 
- & 
- & 
- & 
- & 
\tcbox[colback=LightGreen]{\textcolor{black}{$\uparrow$2}} & 
\tcbox[colback=LightGreen]{\textcolor{black}{$\uparrow$1}} & 
\tcbox[colback=LightRed]{\textcolor{black}{$\downarrow$1}} & 
- & 
- & 6 & 8 \\ & 
\statuelibertyemoji & 
- & 
\tcbox[colback=LightRed]{\textcolor{black}{$\downarrow$1}} & 
\tcbox[colback=LightGreen]{\textcolor{black}{$\uparrow$1}} & 
\tcbox[colback=LightGreen]{\textcolor{black}{$\uparrow$1}} & 
\tcbox[colback=LightGreen]{\textcolor{black}{$\uparrow$1}} & 
- & 
- & 
- & 
- & 
- & 
\tcbox[colback=LightRed]{\textcolor{black}{$\downarrow$2}} & 
- & 
- & 
- & 5 & 6 \\ \bottomrule
\end{tabular}}
\caption{Model rankings with MA \memoemoji rank as the reference for mid (\midemoji ) and low (\lowemoji) resource languages. First row indicates changes in CA \scalesemoji{} ranks, while second row shows the changes in CS \statuelibertyemoji{} ranks relative to MA. Color-coded boxes highlight increases (\tcbox[colback=LightGreen]{\textcolor{black}{$\uparrow$}}) and decreases (\tcbox[colback=LightRed]{\textcolor{black}{$\downarrow$}}).}
\label{tab:model_rank_changes_mid_low_res}
\end{table}
\FloatBarrier

\subsection{Subject-level Performance}
\label{app:sl_perf}

\begin{figure}[ht!]
    \centering
    \centering
    {\includegraphics[width=0.95\linewidth]{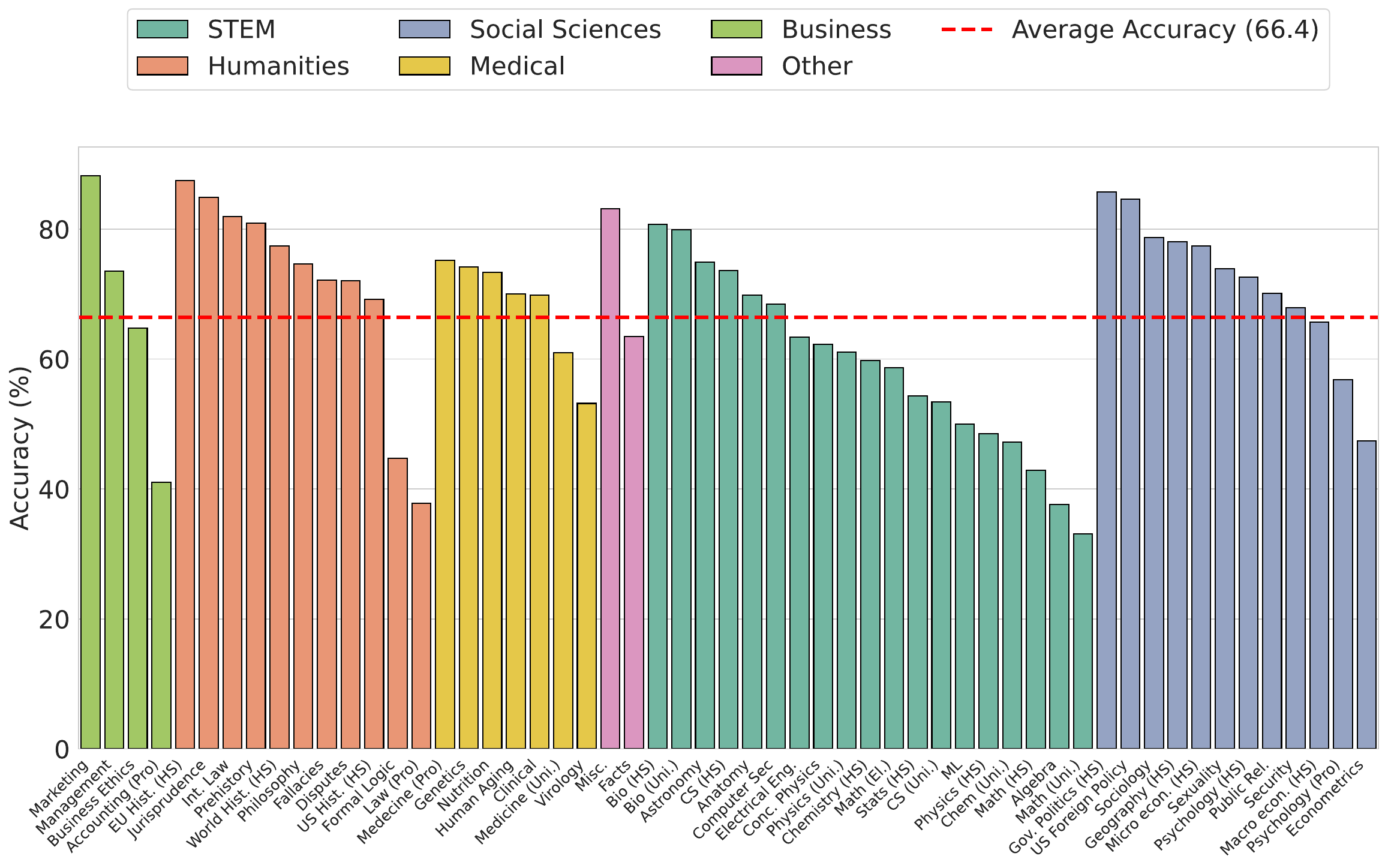}}
    \caption{Aya Expanse 32B performance on each subjects.}
    \label{fig:subject_level}
\end{figure}

Figure~\ref{fig:subject_level} illustrates the performance of the Aya Expanse 32 model across various subjects, with an average accuracy of 66.4\%. Notably, most \emph{STEM} subjects fall below this average, whereas the majority of \emph{Social Sciences} and \emph{Humanities} subjects exceed it.
 
\section{Relationship between cultural and geographical tags }
\label{app:culture-region}

\subsection{Culture--Region Relations}
\label{app:culture-region2}

\begin{figure}[ht!]
    \centering
    \centering
    {\includegraphics[width=0.95\linewidth]{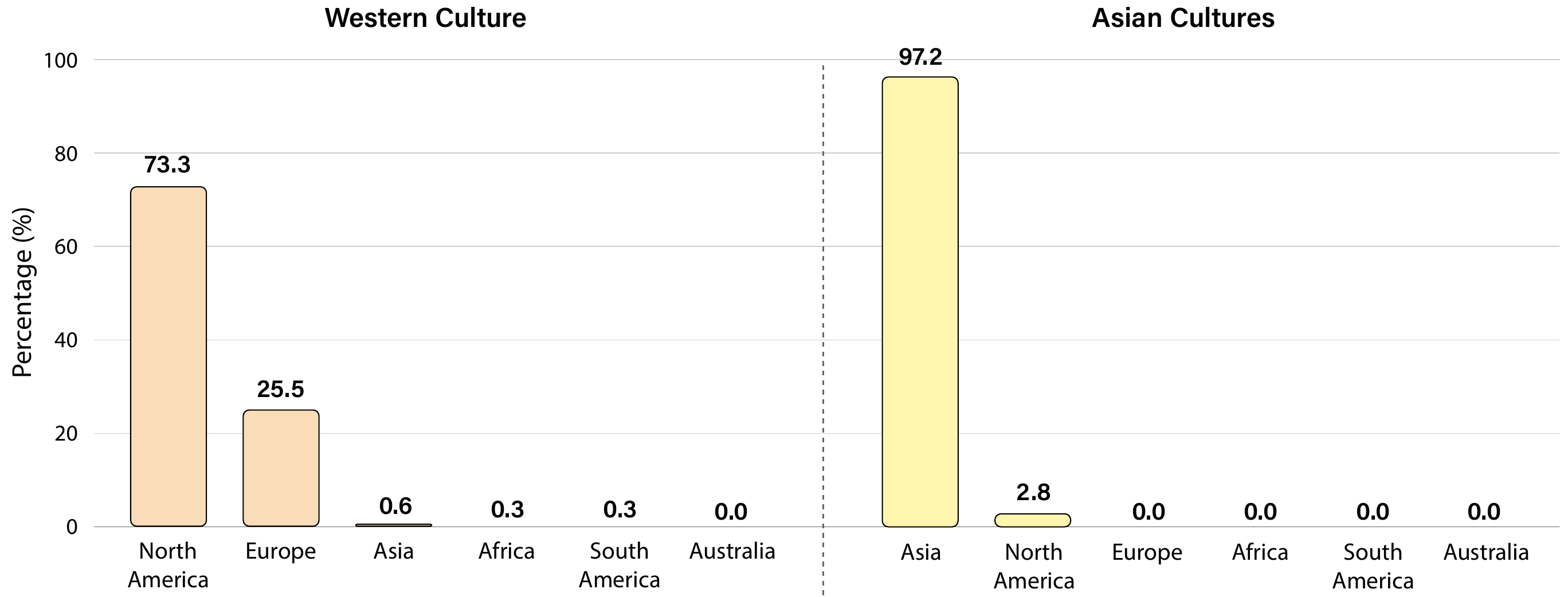}}
    \caption{Relationship between Western and Asia cultures and region tags.}
    \label{fig:mmlu_culture_region_1}
\end{figure}

We analyzed the samples in the \cs{} dataset. Figure~\ref{fig:mmlu_culture_region_1} illustrates the relationship between Western and Asian cultures and their associated regions. Among the samples labeled with a Western culture tag, 73.3\% are also tagged with North America, followed by 25.5\% with Europe. Similarly, 97.2\% of samples labeled with Asian cultures are associated with the Asia region. 

\subsection{Culture Country Relations}
\label{app:culture-country}

\begin{figure}[ht!]
    \centering
    \centering
    {\includegraphics[width=0.95\linewidth]{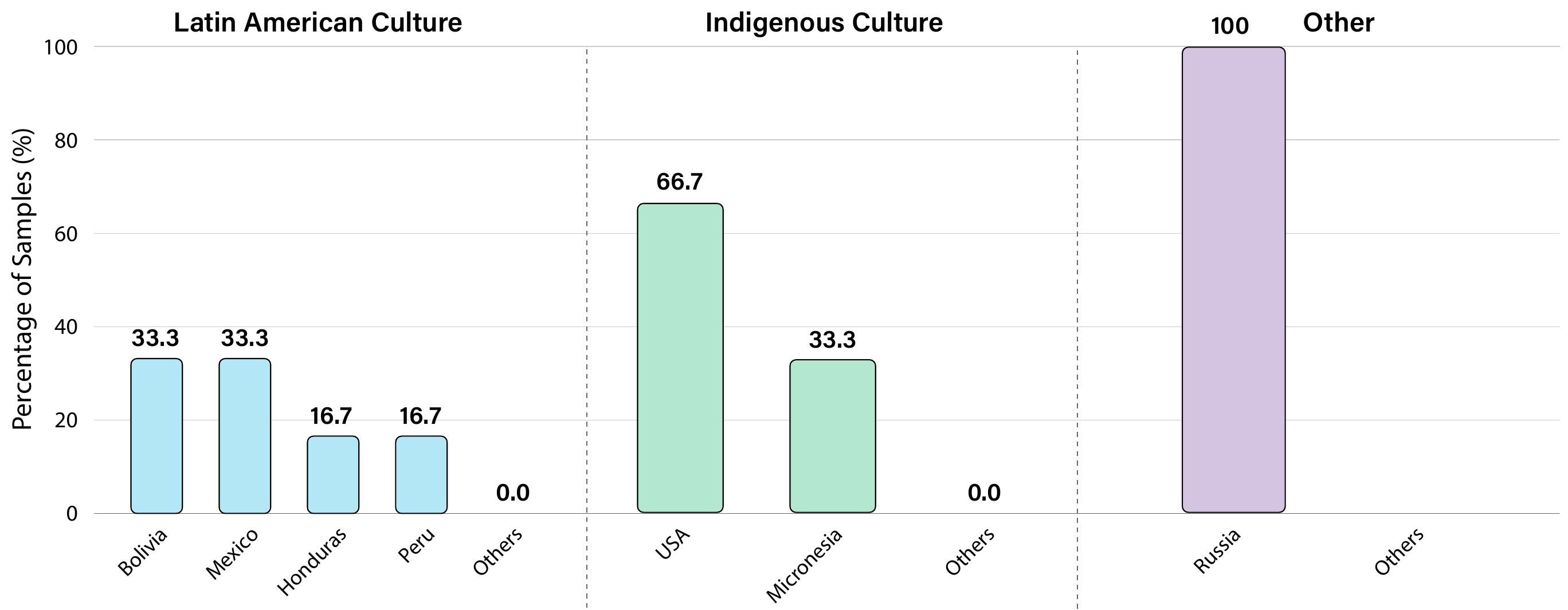}}
    \caption{Relationship between culture and country tags, focusing on Latin American and Indigeneous cultures.}
    \label{fig:mmlu_culture_region_4}
\end{figure}

Figure~\ref{fig:mmlu_culture_region_4} shows relationship between culture and country. For the Latin American culture, the distribution is balanced, with Bolivia and Mexico comprising 33.3\% each of the tags, followed by Hondurus and Peru sharing 16.7\% of the tags each. 
For Indigenous culture, the tags are shared between two countries with USA at top with 66.7\% followed by Micronesia at 33.3\%. The \emph{Other} culture category was added for representing cultures that did not fall under other pre-existing categories. We find that all samples Other category fall under Russia. 

\subsection{Region Country Relations}
\label{app:region-country}

\begin{figure}[ht!]
    \centering
    \centering
    {\includegraphics[width=0.95\linewidth]{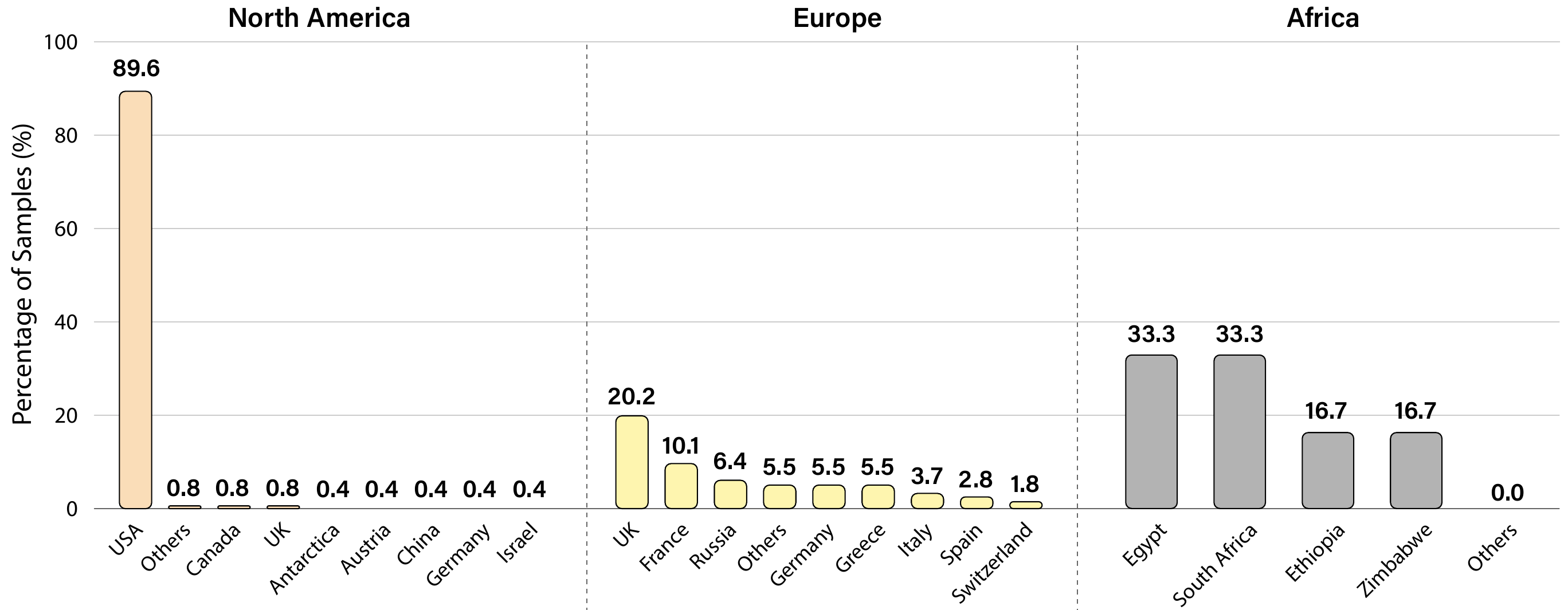}}
    \caption{Relationship between region and country tags, focusing on North America, Europe and Africa regions.}
    \label{fig:mmlu_culture_region_2}
\end{figure}

\begin{figure}[ht!]
    \centering
    \centering
    {\includegraphics[width=0.95\linewidth]{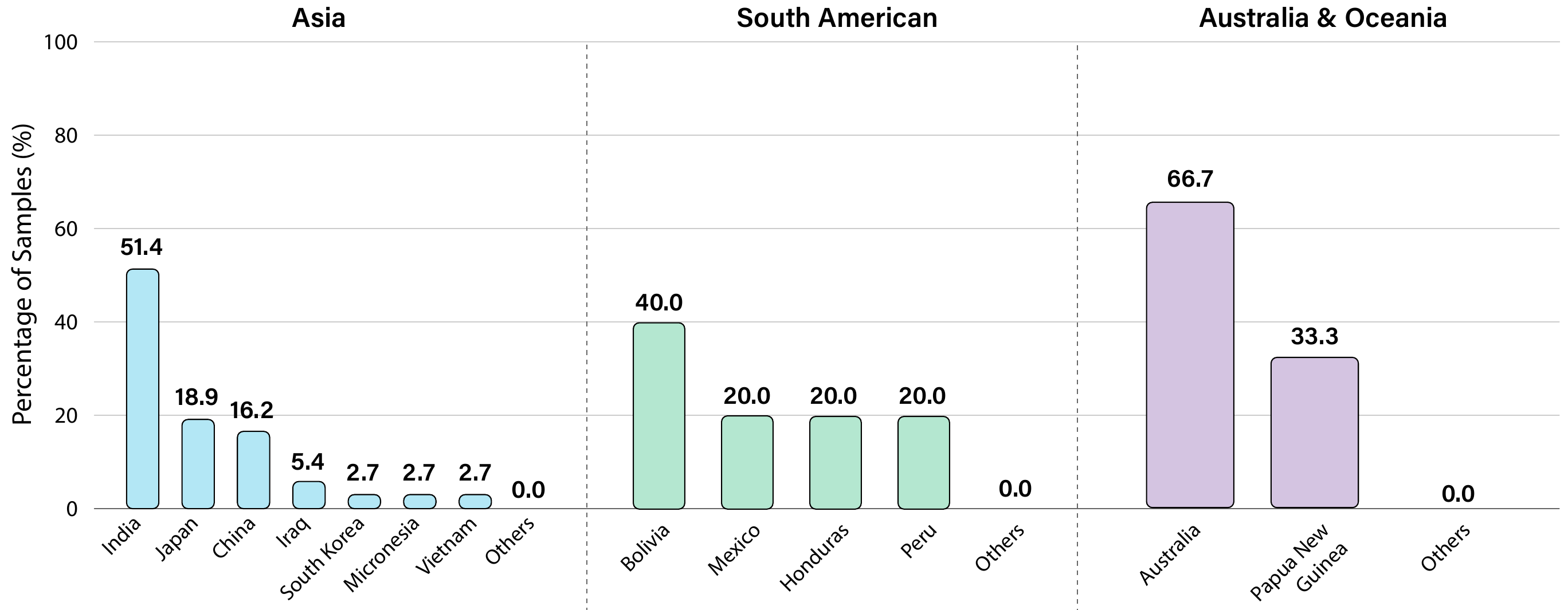}}
    \caption{Relationship between region and country tags, focusing on Asia, South America and Australia.}
    \label{fig:mmlu_culture_region_3}
\end{figure}

Figure~\ref{fig:mmlu_culture_region_2} and \ref{fig:mmlu_culture_region_3} present country-specific information for each region: \emph{North America}, \emph{Europe}, and \emph{Africa}. The United States accounts for the largest proportion of regional tags, representing 89.6\% of the tags for the North America region, followed by Canada and the United Kingdom, each with only 0.8\% of the tags. For the Europe region, the distribution is more balanced, with the United Kingdom comprising 20.1\% of the tags, followed by France at 10.1\%. In the Africa region, the distribution is even more balanced, with Egypt and South Africa sharing the top position at 33.3\% of the tags each.

\section{Annotation Process}
\label{app:annotation_process}

\textbf{Communication.} For both annotation tasks, annotators were briefed by one of the authors in a virtual introduction session and were able to ask questions and raise issues throughout the annotation task in a Discord channel. For both tasks, they were also encouraged to share frequent error patterns or artifacts that they observed throughout the tasks with the authors and capture difficult decisions and their rationales in comments for individual ratings. 
Similarly, they discussed ambiguous cases and questions. This helped calibrate annotations across annotators and languages. 

\textbf{Schedule.} Each of the annotation tasks was conducted as 2--3 week long sprints in collaboration with contributors from the community. There was no fixed time schedule for the annotations, and annotators contributed varying hours, depending on their availability and speed. 

For the cultural sensitivity evaluation task, 100\% of the selected samples were labeled whereas for the translation quality evaluation task, 37\% of the provided samples were fully reviewed 12.3\% of the samples were edited in total. 

\textbf{Interface.} The annotation interface for both tasks was built using Argilla.\footnote{\url{https://argilla.io/}} Argilla is an open-source tool that can be used for data labeling. Using Argilla's Python SDK, it was quick and easy to set up an annotation interface that could be deployed on Hugging Face Spaces. We also set up SSO so annotators could log in and easily access the UI using their Hugging Face accounts.

\begin{figure}[ht!]
    \centering
    \begin{subfigure}[b]{.9\textwidth}
    \centering
    {\includegraphics[width=\linewidth]{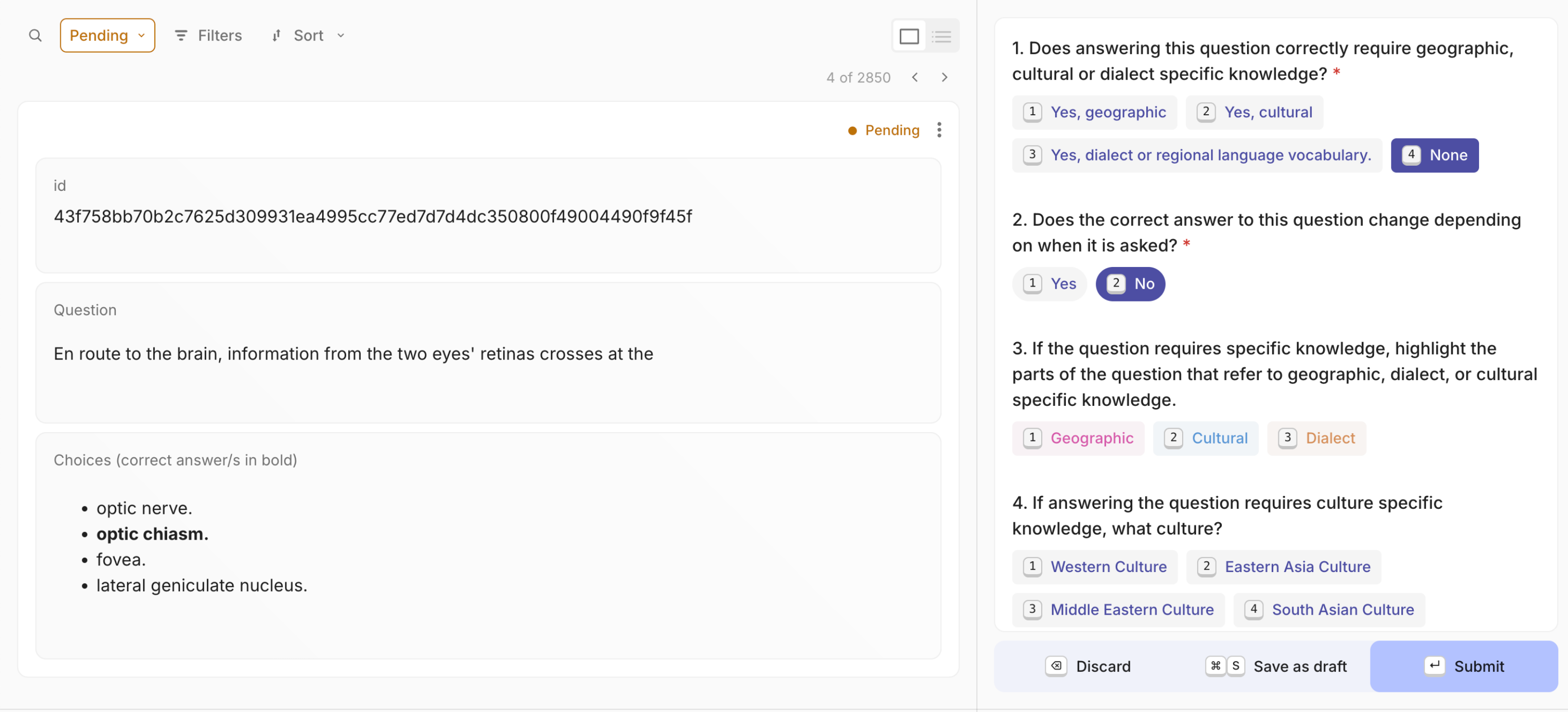}}
    \end{subfigure}
    \begin{subfigure}[b]{.9\textwidth}
    \centering
    {\includegraphics[width=\linewidth]{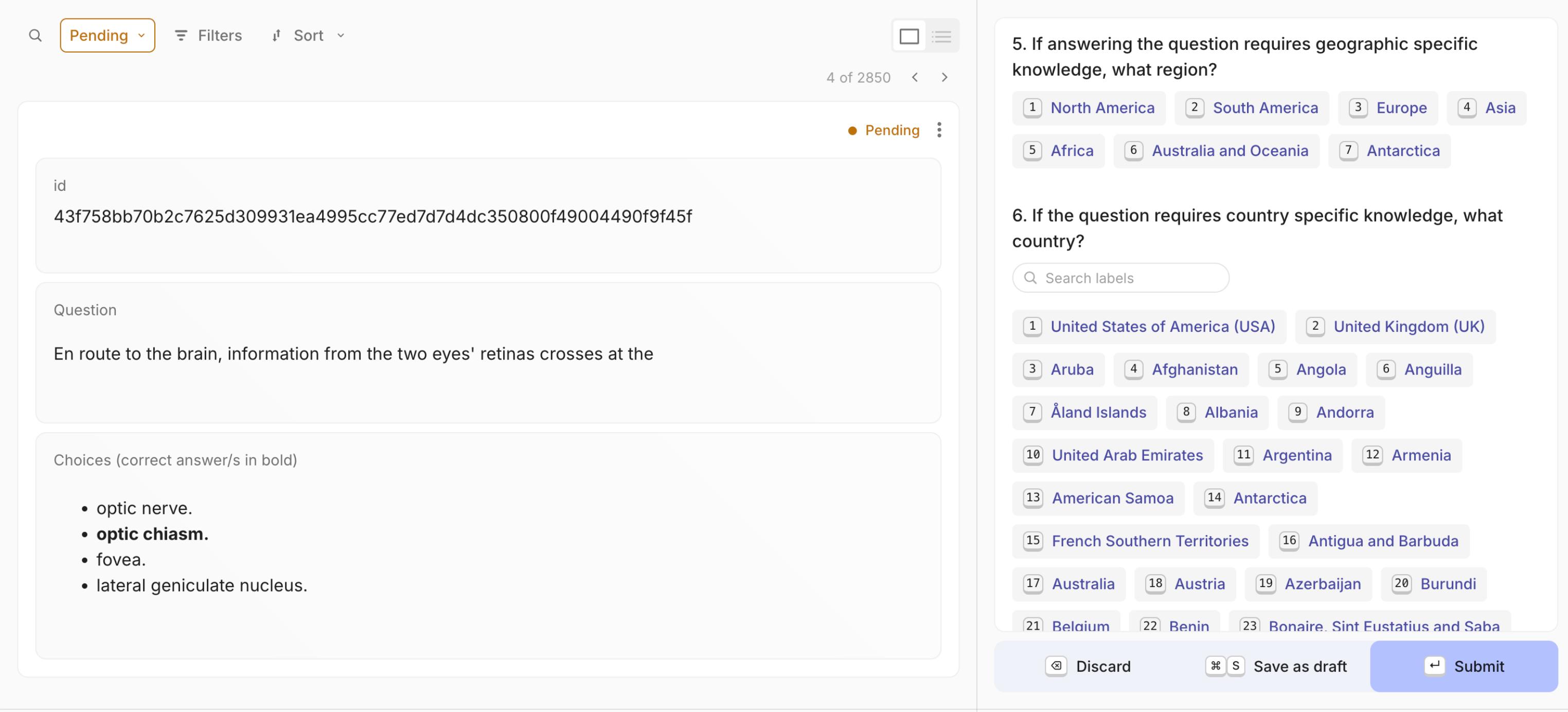}}
    \end{subfigure}
    \caption{Cultural Sensitivity evaluation annotation interface.}
    \label{fig:argilla_annotation_ui_phase1}
\end{figure}

For cultural sensitivity evaluation, annotators were shown questions one by one from each of the 57 MMLU subjects and were asked to analyze and label the questions for presence of cultural, geographic, dialect or regional knowledge as explained in~\ref{sec:annotation} and shown in Figure~\ref{fig:argilla_annotation_ui_phase1}.

\begin{figure}[ht!]
    \centering\includegraphics[width=1.0\textwidth]{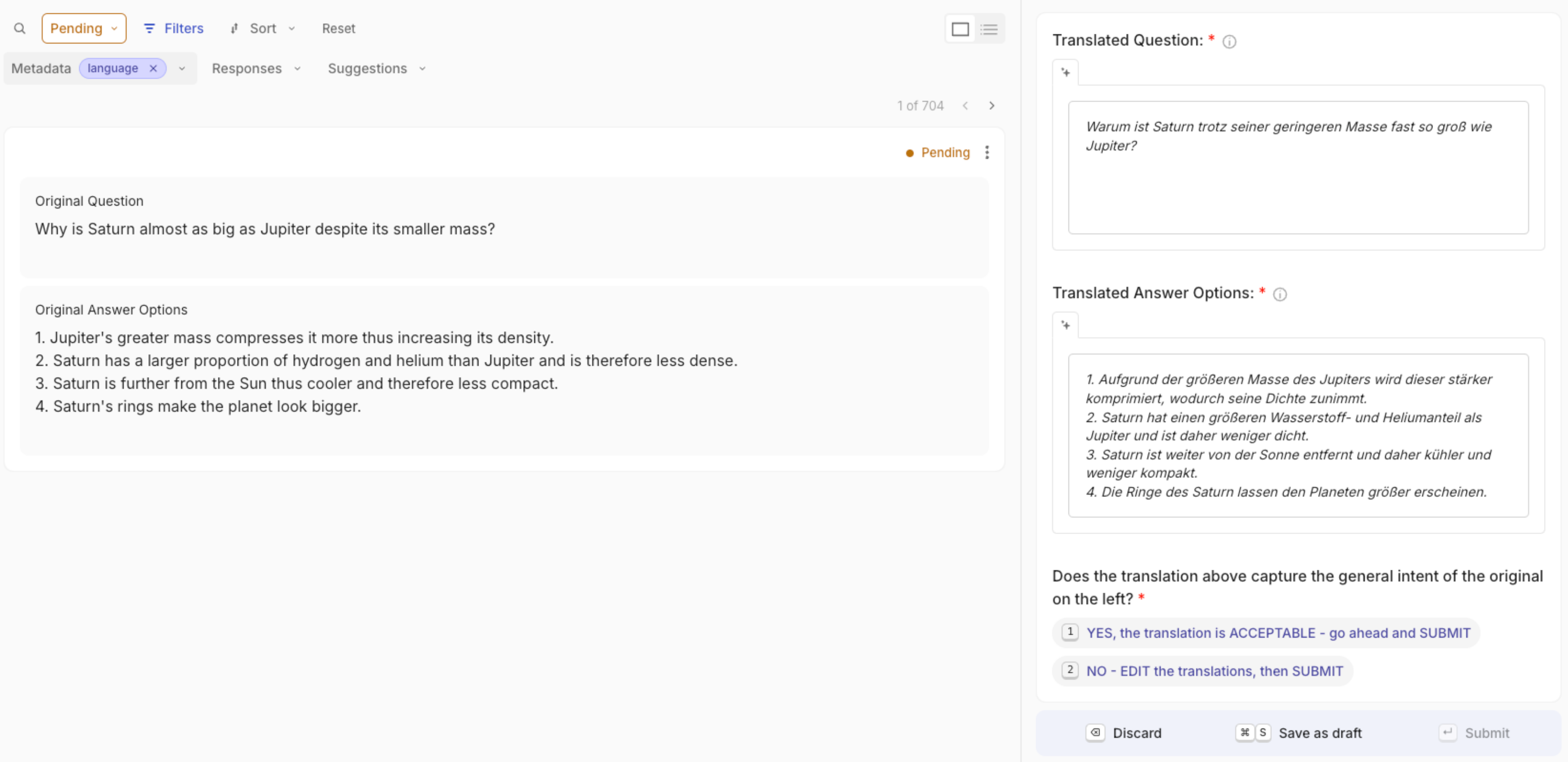}
    \caption{Translation evaluation annotation interface.}
    \label{fig:argilla_tranlation_ui_phase2}
\end{figure}

As shown in Figure~\ref{fig:argilla_tranlation_ui_phase2}, for translation quality evaluation, annotators were shown the translated question and corresponding options in their chosen language on the UI. Annotators were also shown the original question and answer options in English for reference. If the translation was good in quality and correctly represented the original English text then the annotators could mark it as acceptable in quality and proceed to next question otherwise they could edit the provided translation to improve its quality.

\subsection{Compensated Annotator Pool for Gold Standard Languages}
\label{app:annotators_gold_standard}

\textbf{Annotator Selection.} The primary demographic make-up of the participants in the evaluations was recruited based on their proficiency in the language groups. The proficiency was self-reported, and the primary requirement was native or professional proficiency in the specific languages needed for the project. 

\textbf{Socio-Demographics.} 
The annotator pool is comprised of people from diverse backgrounds, and this spans across socioeconomic backgrounds, careers, levels of education, and self-reported gender and sexual identities. We do not ask any annotators to share or report any of these statistical pieces of information in a formal way; any insights into this are gathered organically and through self-reporting by the annotators. 

\textbf{Quality Considerations.} We do not believe that any socio-demographic characteristics have led to any impact on the data that has been annotated. Through every part of the project, we have reiterated the importance of this work and the fact that it is helping support a global-scale research project. We are confident in the trust we have built with the annotators in this project, and they care greatly about the overall outcome and, therefore, have been diligent in completing the task with a high degree of accuracy. 
Where possible, we have done our best to have annotators work on this project and be representatives of the communities that the project aims to support.

\begin{figure}[h!]
    \centering\includegraphics[width=0.99\textwidth]{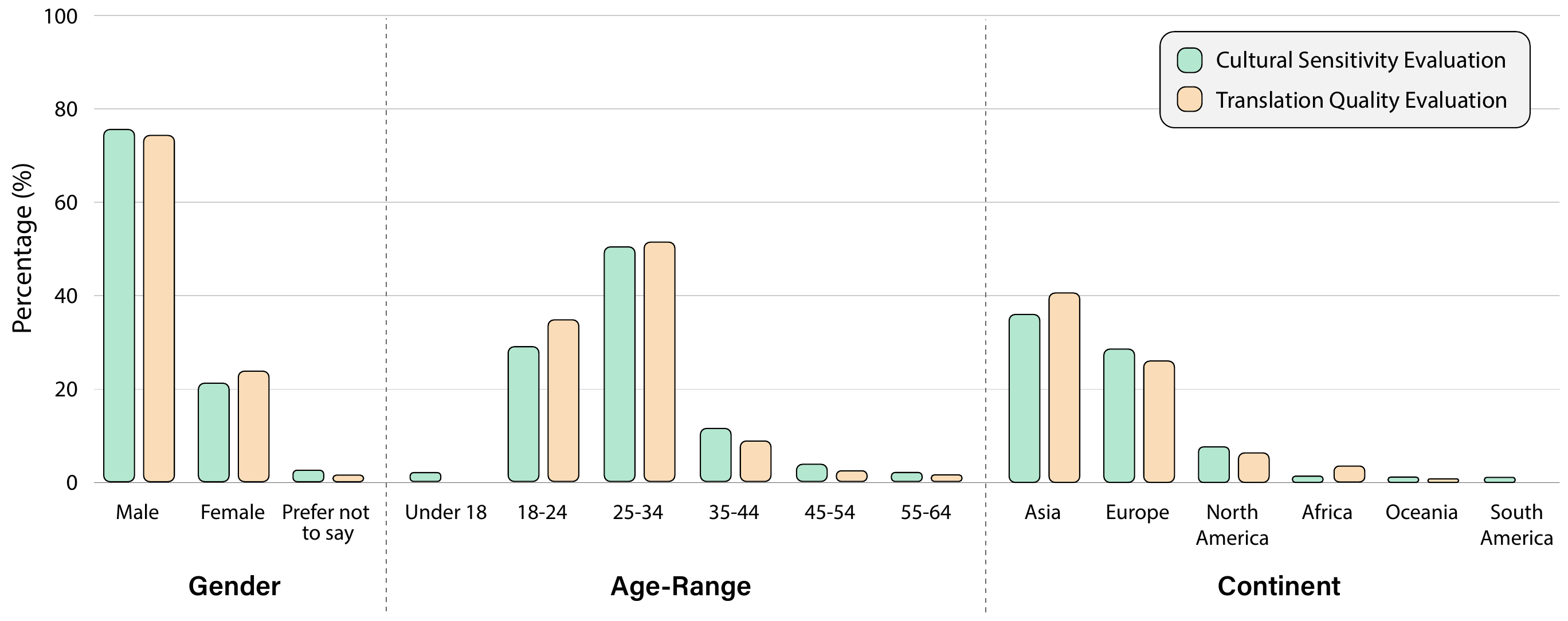}
    \caption{Demographics of annotators who registered using our annotation interface for cultural sensitivity as well as translation quality evaluation.
    }
    \label{fig:mmlu_annotator_demographics}
\end{figure}

\subsection{Agreement between Annotators}
\label{app:annotator_agreement}

\begin{figure}[ht!]
    \centering\includegraphics[width=0.9\textwidth]{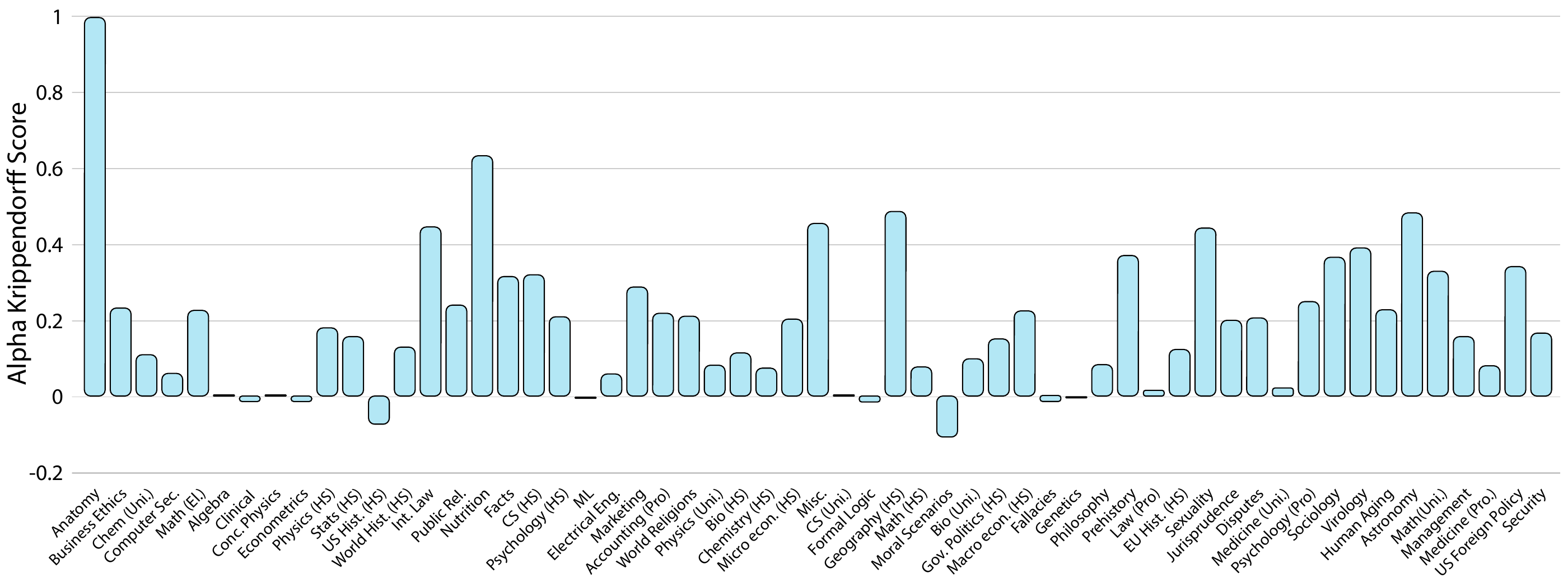}
    \caption{Krippendorff's Alpha Scores for checking annotator agreement regarding the presence of cultural or regional knowledge of samples. }
    \label{fig:alpha_krippendorff_score1}
\end{figure}

\begin{figure}[ht!]
    \centering\includegraphics[width=0.9\textwidth]{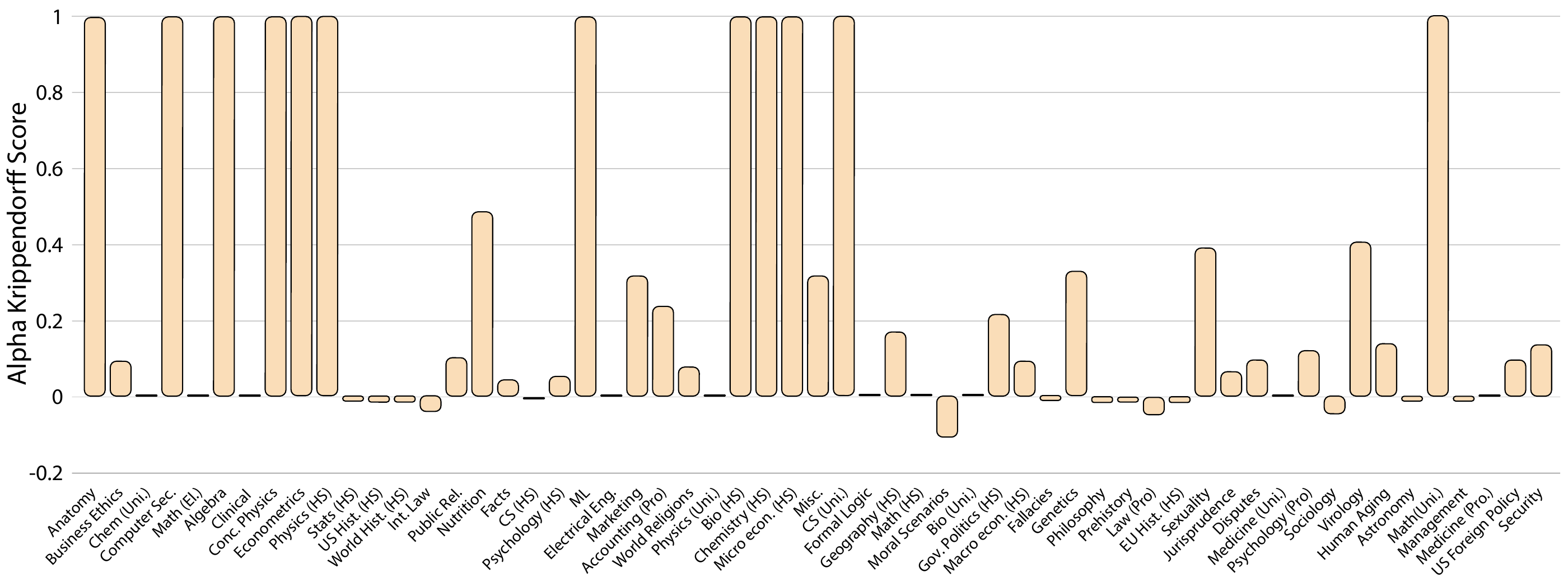}
    \caption{Krippendorff's Alpha Scores for checking annotator agreement regarding the presence of the time-sensitive nature of samples.}
    \label{fig:alpha_krippendorff_score2}
\end{figure}

For the first phase of annotations to identify culturally sensitive samples, we ensured that each sample was annotated by at least 3 annotators. We used the ratings for each sample from different annotators and aggregated it per subject to analyze the agreement among annotators. We report the corresponding Krippendorff's Alpha scores depicting annotator agreement in Figure~\ref{fig:alpha_krippendorff_score1} and ~\ref{fig:alpha_krippendorff_score2}. Krippendorff's Alpha values range between -1 and 1 where 1 denotes that all annotators agree unanimously and -1 denotes that the annotators are making opposite ratings. We observe reasonable disagreement among samples for \emph{moral scenarios} for both cultural sensitivity as well as time-sensitivity annotations. 12 subjects have complete unanimous agreement regarding time-sensitivity annotations between annotators. 

\newpage

\section{Translation Analysis}
\label{app:translation}

\subsection{Translation Quality}
\label{app:translation_quality}
Figure~\ref{fig:mmlu_gtranslate_gpt3.5_comparison} shows the translation quality comparison for Google Translate which is used to translate \gmmlu{} and GPT-3.5-turbo which is used for translating multilingual MMLU released by \citep{lai2023okapi}. We see that Google Translate is significantly better across different MMLU subject categories. For this analysis, we considered samples from MMMLU dataset\footnote{\url{https://openai.com/index/openai-o1-system-card/}} as the human reference and only considered languages which overlapped between the two machine translated sets and human translated MMMLU.

\subsection{Translation Edits}
\label{app:translation_edits}

\begin{figure}[ht!]
    \centering\includegraphics[width=0.65\textwidth]{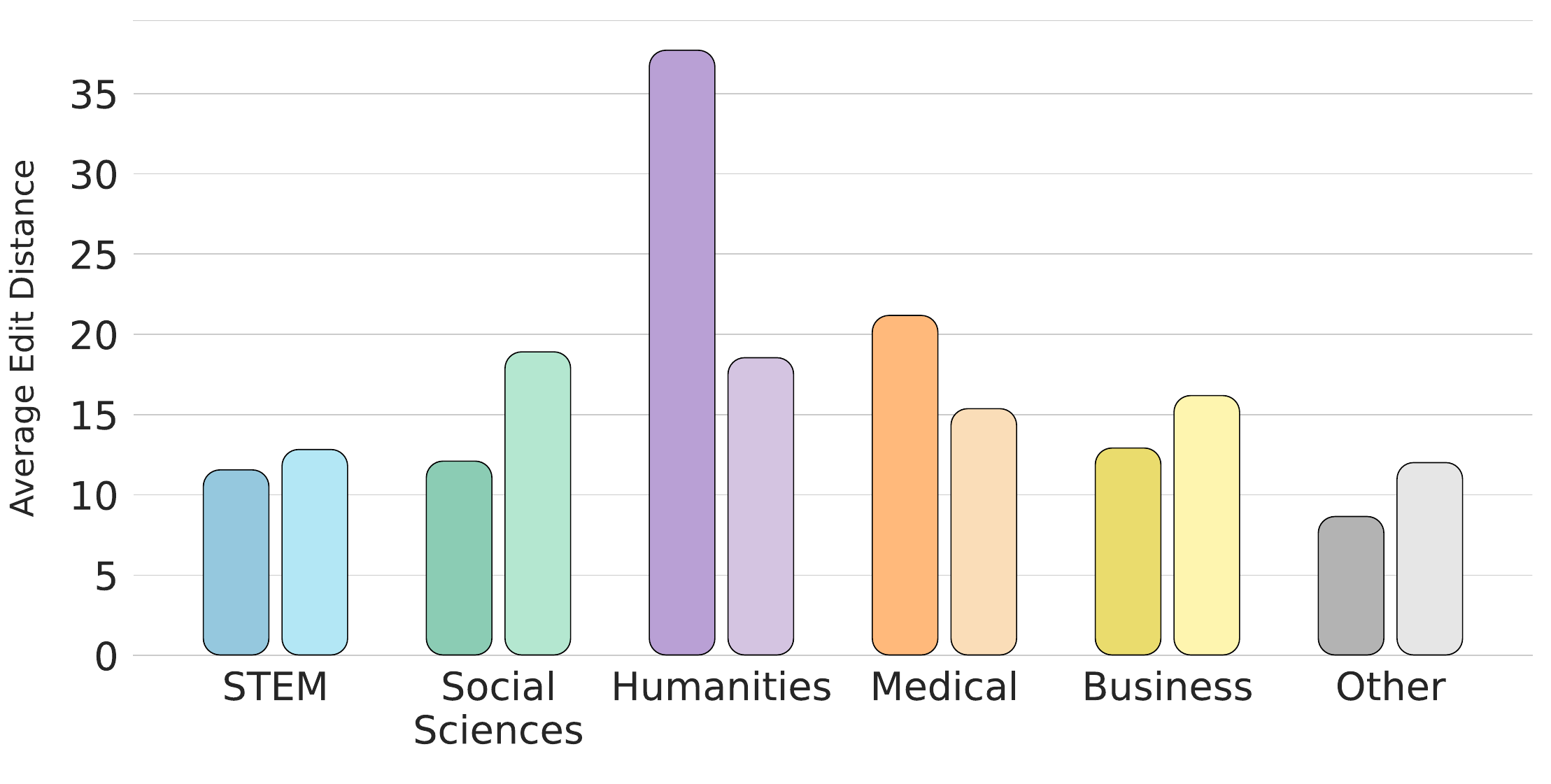}
    \caption{Average edit distance across different subject categories in MMLU. Each sample comprises a question-and-answer pair, with the left column showing edit distances for questions and the right column for answers.}
    \label{fig:mmlu_translation_edit_distance}
\end{figure}

Figure~\ref{fig:mmlu_translation_edit_distance} illustrates the \emph{edit distance}, averaged over all samples within each subject category, for edits made by professional and community annotators. The edit distance, calculated using the ``Levenshtein Distance''~\citep{levenshtein1966binary}, measures the differences between two strings. In this analysis, the machine translations were compared to their edited versions to compute the scores.

The results reveal that the \emph{Humanities} category exhibits the largest edit distances, with higher values observed for questions compared to answers.

\begin{figure}[ht!]
    \centering
    \begin{subfigure}[b]{0.65\textwidth}
    \centering
    {\includegraphics[width=\linewidth]{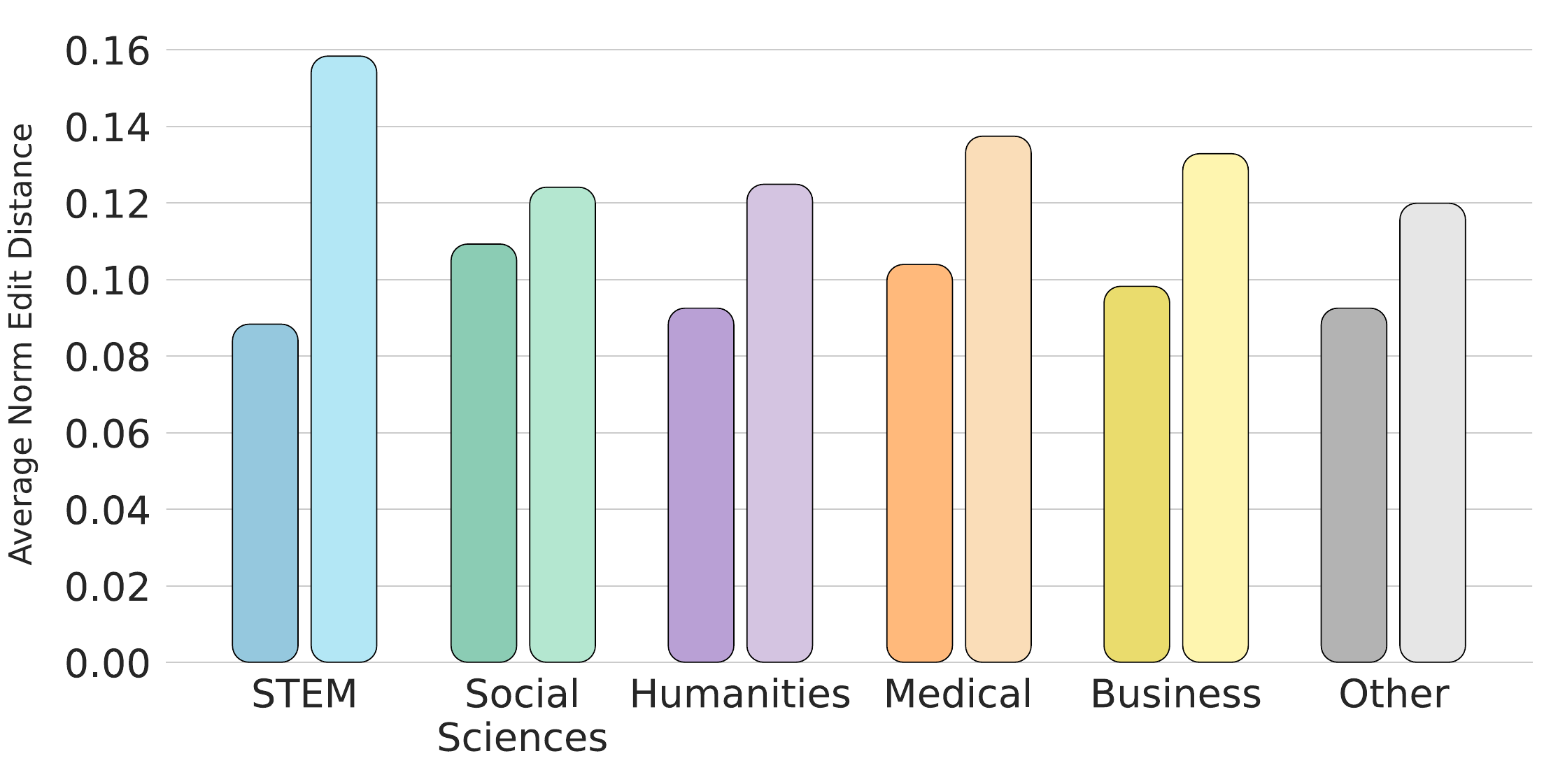}}
    \end{subfigure}
    \begin{subfigure}[b]{0.65\textwidth}
    \centering
    {\includegraphics[width=\linewidth]{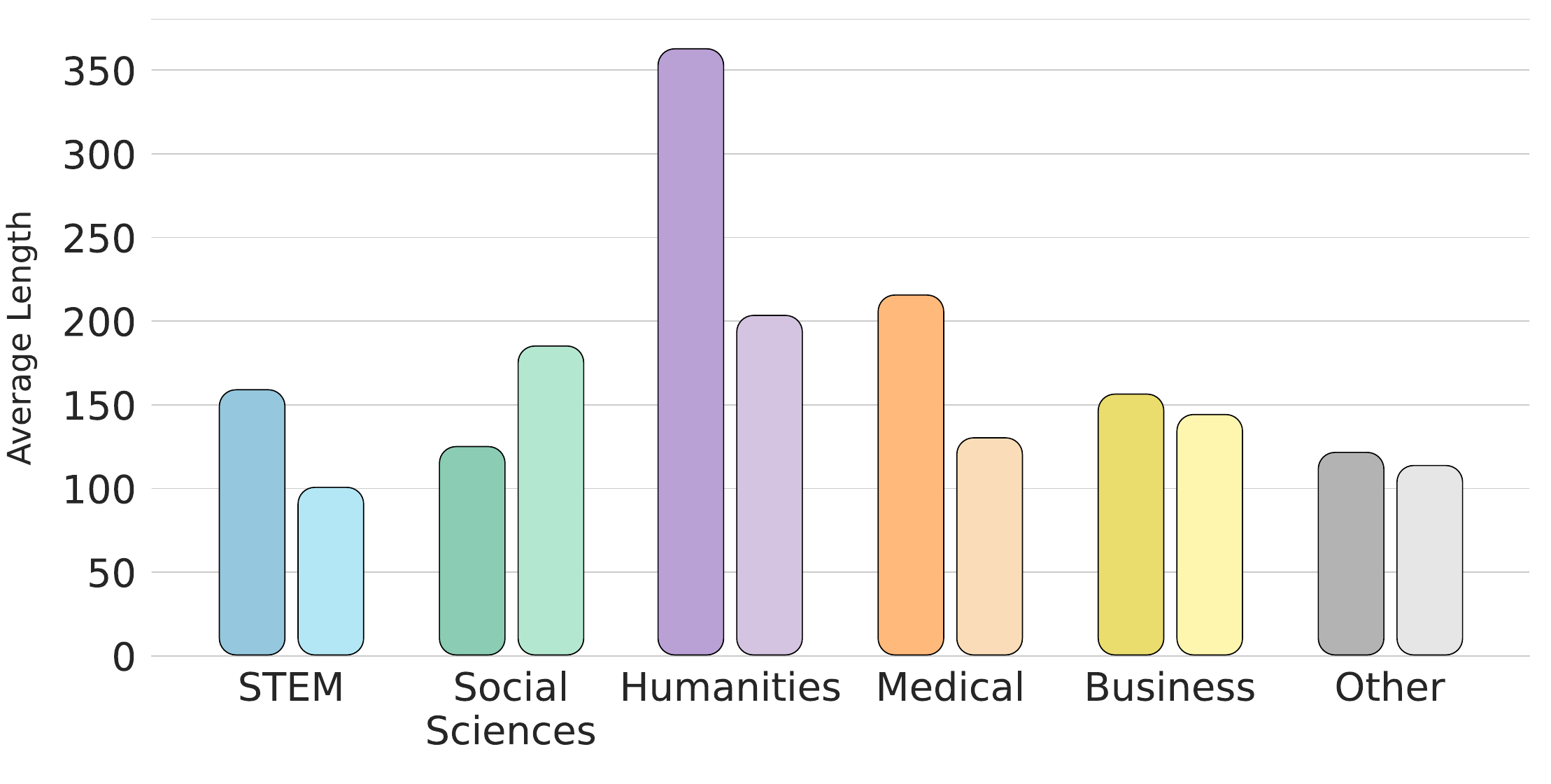}}
    \end{subfigure}
    \caption{(Top) Average normalized edit distance and (Bottom) average question and answer lengths across different subject categories. The left column represents questions, while the right column represents answers.}
    \label{fig:mmlu_translation_normalized_edit_distance}
\end{figure}

Given that longer text may inherently require more edits, we hypothesized that the observed large edit distances could be influenced by the length of the questions and answers. To account for this, we analyzed the length of each question-answer pair and computed the \emph{Normalized Edit Distance} (NED), where the edit distance is divided by the text length, shown in Figure~\ref{fig:mmlu_translation_normalized_edit_distance}.

The analysis reveals that questions in the \emph{Humanities} category have the greatest average length, whereas answers in the \emph{STEM} category exhibit the highest NED. These findings suggest that while raw edit distances are influenced by text length, normalized measures provide additional insights into the complexity of edits across categories.

\FloatBarrier

\clearpage

\section{MMLU Annotated Examples}
\label{app:ma_data_examples}

\begin{center}
\scriptsize
\begin{longtable}
{|l|p{0.2\textwidth}|p{0.3\textwidth}|p{0.4\textwidth}|}
\hline
\rowcolor[HTML]{CFE2F3} \textbf{Dataset} & \textbf{Subject} & \textbf{Question} & \textbf{Choices} \\
\hline
\statuelibertyemoji & \textbf{US Hist. (HS)} & 
This question refers to the following information:

``Some men look at constitutions with sanctimonious reverence, and deem them like the ark of the covenant, too sacred to be touched. They ascribe to the men of the preceding age a wisdom more than human, and suppose what they did to be beyond amendment \dots. But I know also, that laws and institutions must go hand in hand with the progress of the human mind. As that becomes more developed, more enlightened, as new discoveries are made, new truths disclosed, and manners and opinions change with the change of circumstances, institutions must advance also, and keep pace with the times.''

—Thomas Jefferson, 1816

Which of the following best describes a contributing factor in the crafting of the United States Constitution? &
\begin{enumerate}[label=(\Alph*)]
    \item Individual state constitutions written at the time of the Revolution tended to cede too much power to the federal government, leading to a call for reform on the part of Anti-Federalists.
    \item The weaknesses of the Articles of Confederation led James Madison to question their efficacy and prompted a formation of the Constitutional Congress in 1787.
    \item Difficulties over trade and foreign relations led to a repeal of overly restrictive tariffs required by the Articles of Confederation.
    \item Washington's embarrassing failure at the Whiskey Rebellion led to Federalist demands for a new framework for federal power.
\end{enumerate} \\
\hline
\statuelibertyemoji & \textbf{Accounting (Pro)} & 
Under the Sales Article of the UCC, which of the following circumstances best describes how the implied warranty of fitness for a particular purpose arises in a sale of goods transaction? &
\begin{enumerate}[label=(\Alph*)]
    \item The buyer is purchasing the goods for a particular purpose and is relying on the seller's skill or judgment to select suitable goods.
    \item The buyer is purchasing the goods for a particular purpose and the seller is a merchant in such goods.
    \item The seller knows the particular purpose for which the buyer will use the goods and knows the buyer is relying on the seller's skill or judgment to select suitable goods.
    \item The seller knows the particular purpose for which the buyer will use the goods and the seller is a merchant in such goods.
\end{enumerate} \\
\hline
\statuelibertyemoji & \textbf{Jurisprudence} & 
Which of the following criticisms of Llewellyn's distinction between the grand and formal styles of legal reasoning is the most compelling? &
\begin{enumerate}[label=(\Alph*)]
    \item There is no distinction between the two forms of legal reasoning.
    \item Judges are appointed to interpret the law, not to make it.
    \item It is misleading to pigeon-hole judges in this way.
    \item Judicial reasoning is always formal.
\end{enumerate} \\
\hline
\statuelibertyemoji & \textbf{Prehistory} & 
What is the name of the lithic technology seen in the Arctic and consisting of wedge-shaped cores, micro-blades, bifacial knives, and burins? &
\begin{enumerate}[label=(\Alph*)]
    \item Clovis Complex
    \item Denali Complex
    \item Folsom Complex
    \item Nenana Complex
\end{enumerate} \\
\hline
\statuelibertyemoji & \textbf{US Foreign Policy} & 
What was the key difference between US expansion pre- and post- 1865? &
\begin{enumerate}[label=(\Alph*)]
    \item US expansion was based on territory rather than markets post-1865
    \item US expansion was based on markets rather than territory post-1865
    \item US expansion was limited to Latin America post-1865
    \item US expansion ended after 1865
\end{enumerate} \\
\hline
\scalesemoji & \textbf{Econometrics} & 
Which of the following statements will be true if the number of replications used in a Monte Carlo study is small?
i) The statistic of interest may be estimated imprecisely
ii) The results may be affected by unrepresentative combinations of random draws
iii) The standard errors on the estimated quantities may be unacceptably large
iv) Variance reduction techniques can be used to reduce the standard errors
 &
\begin{enumerate}[label=(\Alph*)]
    \item (ii) and (iv) only
    \item (i) and (iii) only
    \item (i), (ii), and (iv) only
    \item (i), (ii), (iii), and (iv)
\end{enumerate} \\
\hline
\scalesemoji & \textbf{Stats (HS)} & 
An assembly line machine is supposed to turn out ball bearings with a diameter of 1.25 centimeters. Each morning the first 30 bearings produced are pulled and measured. If their mean diameter is under 1.23 centimeters or over 1.27 centimeters, the machinery is stopped and an engineer is called to make adjustments before production is resumed. The quality control procedure may be viewed as a hypothesis test with the null hypothesis $H_0: \mu=1.25$ and the alternative hypothesis $H_a: \mu\neq1.25$. The engineer is asked to make adjustments when the null hypothesis is rejected. In test terminology, what would a Type II error result in?
 &
\begin{enumerate}[label=(\Alph*)]
    \item A warranted halt in production to adjust the machinery
    \item An unnecessary stoppage of the production process
    \item Continued production of wrong size ball bearings
    \item Continued production of proper size ball bearings
\end{enumerate} \\
\hline
\scalesemoji & \textbf{Formal Logic} & 
Construct a complete truth table for the following argument. Then, using the truth table, determine whether the argument is valid or invalid. If the argument is invalid, choose an option which presents a counterexample. (There may be other counterexamples as well.)
\( M \lor N \)
\( \neg M \land \frac{O}{N} \)
 &
\begin{enumerate}[label=(\Alph*)]
    \item Valid
    \item Invalid. Counterexample when M and O are true and N is false
    \item Invalid. Counterexample when M is true and O and N are false
    \item Invalid. Counterexample when O is true and M and N are false
\end{enumerate} \\
\hline
\scalesemoji & \textbf{Geography (HS)} & 
Which of the following is MOST likely to experience population pressure?
 &
\begin{enumerate}[label=(\Alph*)]
    \item An industrial society with abundant natural resources and large imports of food
    \item A society with a highly mechanized agricultural sector
    \item A non-ecumene
    \item A slash-and-burn agricultural society
\end{enumerate} \\
\hline
\scalesemoji & \textbf{Nutrition} & 
Why might some biochemical (eg plasma or serum) indices of micronutrient status give misleading results in people with infections or inflammatory states?
 &
\begin{enumerate}[label=(\Alph*)]
    \item Because people who are sick often alter their diets, and may eat less food.
    \item Because the accuracy of some laboratory assays may be compromised in samples from people who are sick.
    \item Because some metabolic pathways are altered in sick people, which changes their micronutrient requirements.
    \item Because an acute phase reaction results in changes in inter-tissue distributions of certain micro-nutrients.
\end{enumerate} \\
\hline
\end{longtable}
\end{center}

\section{Examples of Cultural, Geographical and Dialect Knowledge}
\label{app:cul_sensitivity_examples}

This section lists some examples of cultural, geographical (or regional) and dialect knowledge that was shared with the annotators to guide them during the annotation process.

\begin{center}
\scriptsize
\begin{longtable}
{|l|p{0.4\textwidth}|p{0.5\textwidth}|p{0.4\textwidth}|}
\hline
\rowcolor[HTML]{CFE2F3} \textbf{Knowledge} & \textbf{Applicable Examples} & \textbf{Non-Applicable Examples} \\
\hline
Cultural & 
\begin{enumerate}[label=(\Alph*)]
    \item Understanding religious customs: For instance, the significance of colored powder during Holi in Hindu culture. 
    \item Awareness of traditional arts: For instance, the unique styles and techniques of Indigenous Australian art, often featuring dot painting and storytelling.
    \item References to liberal/conservative attitudes: We can’t assume the notion of liberal is specific to a certain culture or region but it inevitably involves social values and culture.
    \item References to philosophy and philosophical concepts, including philosophy of law: Some familiar philosophical concepts fall within critical cultural contexts. Hume's conception of practical reason is a familiar philosophical concept in western culture. Logical fallacies also fall under this category.
\end{enumerate} &
\begin{enumerate}[label=(\Alph*)]
    \item Universal scientific principles: Knowledge of gravity or evolution is not exclusive to any particular culture.
    \item Principles from the social sciences: The principle of social exchange, that posits that social behavior is the result of an exchange process, is used worldwide.
    \item Standardized international sports: The rules and practices of soccer (football) are consistent worldwide.
    \item Math questions which do not rely on local references: For example, the formula for the radius of a circle.
\end{enumerate} \\
\hline
Geographical & 
\begin{enumerate}[label=(\Alph*)]
    \item Natural Landmark Identification: Recognizing and knowing the significance of regional natural wonders like the Grand Canyon in the Southwestern United States or the Great Barrier Reef in Australia. 
    \item Environmental Awareness: Understanding the impact and importance of regional weather patterns, such as the monsoons in South Asian regions or the hurricanes in the Caribbean.
    \item Historical Event Memory: Knowledge of region-specific historical occurrences, such as the Gold Rush in California during the 1850s, which transformed the region's economy and demographics.
    \item Awareness of a region-specific natural phenomenon: The Northern Lights, visible in the night skies of Alaska and northern regions.
    \item Systems of measurement that are specific to a geographic area: Imperial units are used to measure distance (eg. miles), volume (eg. gallons) and weight (eg. pounds)
    \item Laws and regulations: A programmer uses code published online under a Creative Commons Attribution (CCBY) license in a commercial product. This license is specific to the regional geographic area it was created in.
    \item Behaviors and preferences of groups in specified areas: These can be noted as both ``cultural'' and ``geographic'', as in the exam ``Which of the following statements does NOT accurately describe voting behavior in the United States?'' voting practices are cultural, and the US is specified as a geographic area.
\end{enumerate} &
\begin{enumerate}[label=(\Alph*)]
    \item Global Climate Patterns: Understanding El Niño and La Niña weather phenomena, which occur worldwide and are not specific to any single region.
    \item Universal Celestial Bodies: The Sun and the Moon are visible worldwide and do not possess regional specificity.
    \item Standardized Geography Terms: Understanding the definition of a peninsula or archipelago is applicable to geographic features globally, not tied to regional knowledge.
\end{enumerate} \\
\hline
\hline
Dialect & 
\begin{enumerate}[label=(\Alph*)]
    \item Regional slang: Using the word ``wicked'' to mean ``very good'' in parts of New England, USA. Using the phrase “boot of the car” to mean “trunk” in the UK. 
    \item Unique idiomatic expressions: The phrase ``Bob's your uncle'' in British English, meaning ``there you have it'' or ``that's all there is to it.''
    \item Knowledge of social greetings: The customary handshake and verbal greeting of ``Konnichiwa'' when meeting someone in Japanese culture.
    \item Words or phrases from other languages that are brought into English: as in the sentence ``he has that je ne sais quoi'' in which je ne sais quoi is borrowed from French
\end{enumerate} &
\begin{enumerate}[label=(\Alph*)]
    \item Standardized technical jargon: Medical or legal terminology used internationally within professional fields.
    \item Formal literary language: The writings of Shakespeare or Dickens utilize sophisticated language but are not tied to specific dialects.
    \item Global brand names: Companies like Nike or Adidas use consistent branding worldwide, avoiding regional vocabulary.
\end{enumerate} \\
\hline
\end{longtable}
\end{center}

\section{MMLU Subject Name Mapping}
\label{app:mmlu_subject_mapping}

\begin{center}
\scriptsize
\begin{longtable}{lllllcc}
\toprule
Original Name & Short Name \\
\midrule 
\hline
abstract\_algebra & Algebra \\
anatomy & Anatomy \\
astronomy & Astronomy \\
business\_ethics & Business Ethics \\
clinical\_knowledge & Clinical \\
college\_biology & Bio (Uni.) \\
college\_chemistry & Chem (Uni.) \\
college\_computer\_science & CS (Uni.) \\
college\_mathematics & Math (Uni.) \\
college\_medicine & Medicine (Uni.) \\
college\_physics & Physics (Uni.) \\
computer\_security & Computer Sec \\
conceptual\_physics & Conc. Physics \\
econometrics & Econometrics \\
electrical\_engineering & Electrical Eng. \\
elementary\_mathematics & Math (El.) \\
formal\_logic & Formal Logic \\
global\_facts & Facts \\
high\_school\_biology & Bio (HS) \\
high\_school\_chemistry & Chemistry (HS) \\
high\_school\_computer\_science & CS (HS) \\
high\_school\_european\_history & EU Hist. (HS) \\
high\_school\_geography & Geography (HS) \\
high\_school\_government\_and\_politics & Gov. Politics (HS) \\
high\_school\_macroeconomics & Macro econ. (HS) \\
high\_school\_mathematics & Math (HS) \\
high\_school\_microeconomics & Micro econ. (HS) \\
high\_school\_physics & Physics (HS) \\
high\_school\_psychology & Psychology (HS) \\
high\_school\_statistics & Stats (HS) \\
high\_school\_us\_history & US Hist. (HS) \\
high\_school\_world\_history & World Hist. (HS) \\
human\_aging & Human Aging \\
human\_sexuality & Sexuality \\
international\_law & Int. Law \\
jurisprudence & Jurisprudence \\
logical\_fallacies & Fallacies \\
machine\_learning & ML \\
management & Management \\
marketing & Marketing \\
medical\_genetics & Genetics \\
miscellaneous & Misc. \\
moral\_disputes & Disputes \\
moral\_scenarios & Moral Scenarios \\
nutrition & Nutrition \\
philosophy & Philosophy \\
prehistory & Prehistory \\
professional\_accounting & Accounting (Pro) \\
professional\_law & Law (Pro) \\
professional\_medicine & Medicine (Pro) \\
professional\_psychology & Psychology (Pro) \\
public\_relations & Public Rel. \\
security\_studies & Security \\
sociology & Sociology \\
us\_foreign\_policy & US Foreign Policy \\
virology & Virology \\
world\_religions & World Religions \\
\bottomrule
\caption{This table shows the short names assigned to MMLU subjects proposed by \citep{hendrycks2020measuring} in Figures~\ref{fig:mmlu_cul_reg_dia_ref},~\ref{fig:filtered_mmlu_samples},~\ref{fig:alpha_krippendorff_score1},~\ref{fig:alpha_krippendorff_score2}.
} 
\label{tab:mmlu_subject_name_mapping}
\end{longtable}
\end{center}

\end{document}